\documentclass[lettersize,journal]{IEEEtran}
\usepackage{amsmath,amsfonts}
\usepackage{algorithmic}
\usepackage{algorithm}
\usepackage{array}
\usepackage[caption=false,font=normalsize,labelfont=sf,textfont=sf]{subfig}
\usepackage{textcomp}
\usepackage{stfloats}
\usepackage{url}
\usepackage{verbatim}
\usepackage{graphicx}
\usepackage{cite}
\usepackage{amsmath}
\usepackage{amssymb}
\usepackage{mathtools}
\usepackage{amsthm}
\usepackage{arydshln}

\theoremstyle{plain}
\newtheorem{theorem}{Theorem}[section]

\newtheorem{corollary}[theorem]{Corollary}
\theoremstyle{definition}

\theoremstyle{remark}

\usepackage{multirow}
\usepackage{makecell}

\usepackage[dvipsnames]{xcolor}
\usepackage{colortbl} 
\colorlet{colorFst}{Green!25}       
\colorlet{colorSnd}{SpringGreen!45} 
\colorlet{colorTrd}{Yellow!30}      
\colorlet{colorLow}{darkgray!30}    
\newcommand{\fs}{\cellcolor{colorFst}}   
\newcommand{\nd}{\cellcolor{colorSnd}}      
\newcommand{\rd}{\cellcolor{colorTrd}}      

\begin{document}

\title{INR-Bench: A Unified Benchmark for Implicit Neural Representations in Multi-Domain Regression and Reconstruction}

\author{
    Linfei Li, Fengyi Zhang, Zhong Wang, Lin Zhang, and Ying Shen%
    \thanks{
        Linfei Li, Lin Zhang, and Ying Shen are with the School of Computer Science and Technology, Tongji University, Shanghai, China.\\
        Fengyi Zhang is a Ph.D. student with the School of Electrical Engineering and Computer Science, The University of Queensland, Brisbane, Australia.\\
        Zhong Wang is a Postdoctoral Researcher with the Department of Automation, Shanghai Jiao Tong University, Shanghai, China.\\
        Corresponding author: Lin Zhang (email: cslinzhang@tongji.edu.cn)
    }%
}

\maketitle

\begin{abstract}
Implicit Neural Representations (INRs) have gained success in various signal processing tasks due to their advantages of continuity and infinite resolution. However, the factors influencing their effectiveness and limitations remain underexplored. To better understand these factors, we leverage insights from Neural Tangent Kernel (NTK) theory to analyze how model architectures (classic MLP and emerging KAN), positional encoding, and nonlinear primitives affect the response to signals of varying frequencies. Building on this analysis, we introduce INR-Bench, the first comprehensive benchmark specifically designed for multimodal INR tasks. It includes 56 variants of Coordinate-MLP models (featuring 4 types of positional encoding and 14 activation functions) and 22 Coordinate-KAN models with distinct basis functions, evaluated across 9 implicit multimodal tasks. These tasks cover both forward and inverse problems, offering a robust platform to highlight the strengths and limitations of different neural models, thereby establishing a solid foundation for future research. The code and dataset are available at \url{https://github.com/lif314/INR-Bench}.
\end{abstract}


\section{Introduction}
\IEEEPARstart{U}{nlike} traditional discrete signal representation methods (e.g., audio signals as amplitude samples, images as pixel grids, and 3D shapes as points, voxels, or meshes), implicit neural representations take the signal domain (coordinates) as input and employ a neural network to regress the corresponding content, thereby storing the signal within the network's weights. This representation not only exhibits continuity but also decouples the signal from spatial resolution, offering infinite resolution. As a result, the storage required to parameterize the signal is independent of spatial resolution, enabling sampling at arbitrary resolutions.  Due to these advantages, implicit neural representations have been successfully applied to a variety of signal processing tasks, including audio super-resolution \cite{kim2022lisa}, image fitting/super-resolution/denoising \cite{sitzmann2020siren, kazerouni2023incode, saragadam2023wire}, CT/MRI reconstruction \cite{tancik2020ffn, shen2022nerp}, 3D shape regression \cite{park2019deepsdf, wang2023neus, mescheder2019occupancynet} and neural radiance fields \cite{mildenhall2020nerf, barron2022mipnerf360}.

Coordinate-MLPs are the prevailing model for implicit neural representations. However, in their basic form, when using common activation functions (e.g., ReLU \cite{Nair2010relu}), they struggle to effectively encode high-frequency signals. To mitigate this issue, one solution is to use Fourier mappings as the positional encoding \cite{mildenhall2020nerf, tancik2020ffn}, while another strategy is to employ strongly linear activation functions, such as Gaussian \cite{chng2022garf, ramasinghe2022activations}, Gabor \cite{saragadam2023wire}, or Sine \cite{sitzmann2020siren, kazerouni2023incode}. Furthermore, the Kolmogorov–Arnold Network (KAN) \cite{liu2024kan} is proposed as an alternative network to MLPs and holds potential for application in implicit tasks. Despite the significant progress these methods have achieved in implicit neural representations, the fundamental principles underlying their effectiveness and limitations remain insufficiently explored.

To fill this gap, we turn to Neural Tangent Kernel (NTK) theory \cite{jacot2018neuralntk}, where NTK is defined as the inner product of the Jacobian matrix of the network's output with respect to its parameters. According to this framework, the spectral distribution of the NTK reflects the frequency response of the model during training, with larger eigenvalues indicating more pronounced low-frequency responses and faster convergence. Building on this, we conduct a comprehensive analysis of the impact of different model architectures (MLP and KAN), positional encoding and nonlinear primitives on the NTK spectral distribution, thereby elucidating the implicit representation capabilities of various models. Our main insights are summarized as follows:

(1) KAN exhibits a smaller spectral bias in the low-frequency domain compared to MLP, making it more suitable for low-frequency learning tasks (e.g., image denoising). However, as a heuristic deep Gaussian process, KAN faces challenges related to high computational complexity and training difficulties. In contrast, although MLP has a little larger spectral bias, its simple architecture and fast training still provide significant performance advantages in implicit neural representation tasks.

(2) Fourier feature positional encoding maps input coordinates from Euclidean space to a hypersphere, which theoretically adjusts the width of the NTK spectrum, thereby enhancing the model's convergence and generalization. However, existing positional encodings require an extensive search for frequency hyperparameters to adapt to the NTK spectral distribution in different implicit tasks. To address this issue, we propose FKAN, which utilizes Fourier series as the basis functions for KAN. FKAN is a fully learnable positional encoding that can adaptively adjust the width of the NTK spectrum, thereby enabling Coordinate-MLPs to learn different frequency components in the signal. 

(3) By analyzing nonlinear primitives' impact  on the spectral distribution of the NTK, including their gradients, curvature, Fourier transform, L-Lipschitz continuity, and L-Smoothness,  we derive the following main conclusions: (a) ReLU exhibits zero curvature and its frequency response decays rapidly, which hinders its ability to effectively learn high-frequency signals; (b) The periodic gradient and curvature of Sine enable efficient learning of high-frequency components in low-dimensional signals. However, its fixed Lipschitz constant limits its ability to capture varying frequency signals in complex scenes, lacking multi-scale learning capability; (c) The frequency response of the Gaussian/Quadratic activation function decays exponentially, resulting in a pronounced response to low frequencies, making it particularly well-suited for tasks that require stable learning.

To further verify the aforementioned insights, we propose INR-Bench, the first comprehensive benchmark for implicit neural representation tasks. This benchmark includes performance evaluations of 56 Coordinate-MLPs and 22 Coordinate-KANs across 9 implicit tasks. For Coordinate-MLPs, we investigate 4 types of positional encoding, including our proposed FKAN, along with 23 commonly used activation functions (e.g., ReLU, Sine and Gaussian). This allows for a thorough exploration of the impact of positional encoding and activation functions on representational capacity. The Coordinate-KANs, which include 22 KAN models from four categories: wavelets, polynomials, Fourier, and radial basis functions. Finally, the 9 multimodal tasks consist of 3 forward problems (audio, image and 3D shape regression) and 6 inverse problems (image inpainting/super-resolution/denoising, CT reconstruction and neural radiance fields), which effectively assess the models' abilities to learn multi-scale frequencies and perform inverse reasoning.

In summary, our main contributions are as follows:
\begin{itemize}
    \item Based on NTK theory, we comprehensively analyze the impact of model architecture (MLP and KAN), positional encoding, and nonlinear primitives on the spectral distribution of the implicit neural model's NTK. The analysis reveals the effectiveness and limitations of the implicit representations across different models.

    \item To address positional encoding's sensitivity to frequency hyperparameters, we propose a fully learnable positional encoding (FKAN) using Fourier series as KAN's basis functions, enhancing convergence and generalization of Coordinate-MLPs by adapting the NTK's spectral width.
    
    \item We introduce INR-Bench, the first benchmark for implicit neural representations, evaluating 56 Coordinate-MLPs and 22 Coordinate-KANs across 9 tasks. It can effectively evaluate the frequency learning, inverse reasoning and generalization of different models.
\end{itemize}

\section{Related Work}
\label{sec:formatting}
\textbf{Implicit Neural Models.} Following the breakthroughs in neural radiance fields \cite{mildenhall2020nerf}, Coordinate-MLPs have been applied to various signal tasks, including audio regression \cite{sitzmann2020siren, kazerouni2023incode}, image regression/super-resolution/denoising \cite{tancik2020ffn, saragadam2023wire, lindell2022bacon, ramasinghe2022activations}, 3D shape regression \cite{wang2023neus, yu2022monosdf, yariv2023bakedsdf}, and CT/MRI reconstruction \cite{tancik2020ffn}. 
Recently, Liu \textit{et al.} \cite{liu2024kan} introduce KAN, a learnable network architecture based on Kolmogorov-Arnold theory, which uses B-splines as basis functions. While KAN excels in symbolic regression, its application in implicit representation tasks still lacks theoretical analysis and experimental validation.

\textbf{Positional Encoding.} NeRF \cite{mildenhall2020nerf} improves the ability of ReLU-MLPs to capture high-frequencies by mapping input coordinates into a Fourier space. Additionally, FFN \cite{tancik2020ffn} introduces random Gaussian noise to control the NTK bandwidth, thereby enhancing the generalization ability of Coordinate-MLPs. Although FFN allows MLPs to represent high-frequency components, selecting the appropriate frequency scale often requires an extensive hyperparameter search. 

\textbf{Nonlinear Primitives.} While activation functions like ReLU \cite{Nair2010relu} are simple and widely used, they are limited in their ability to capture high-frequency components of signals, often requiring additional positional encoding \cite{mildenhall2020nerf, barron2022mipnerf360, yu2021pixelnerf, Chen2022tensorf}. Garf \cite{chng2022garf} introduces Gaussian activation function for pose-independent radiance field reconstruction, though it tends to overfit both noise and signal. To enhance robustness to noise, Wire \cite{saragadam2023wire} proposes Gabor activation function. Further, Siren \cite{sitzmann2020siren} introduces frequency-dependent initialization schemes, enabling the Sine activation function to effectively capture high-frequencies. Building on Siren, Incode \cite{kazerouni2023incode} makes the parameters of the sine activation function learnable, thereby reducing its sensitivity to frequency parameters. For KAN, due to its structural flexibility, researchers propose KAN architectures with various basis functions, including wavelet \cite{bozorgasl2024wav}, polynomial \cite{seydi2024Polynomialkan, aghaei2024Jacobikan}, and radial \cite{li2024radialkan} basis functions.

\section{INR-Bench}
\textbf{Preliminary: Spectral Bias.} Given a dataset \((\mathbf{X}, \mathbf{y}) = \left\{(\mathbf{x}_i, y_i), \mathbf{x}_i \in \Omega\right\}_{i=1}^n\), where \(\mathbf{x}_i\) denotes the input coordinate and \(y_i = f(\mathbf{x}_i)\) is the corresponding output label, the implicit neural representation employs a neural network \(\Phi\) to approximate the function values \(\{f(\mathbf{x}_i)\}_{i=1}^n\), thereby encoding the data within the network's weights to facilitate the representation of continuous signals. Formally, the implicit neural representation can be defined as a partial differential equation with respect to \(\Phi\)  \cite{sitzmann2020siren}, 
\begin{equation}
    \text{INR}(\mathbf{x}, \Phi, \nabla_{\mathbf{x}} \Phi, \nabla_{\mathbf{x}}^2 \Phi, \ldots) = 0,  \Phi: \mathbf{x} \mapsto \Phi(\mathbf{x}),
    \label{eq:inr_formulation}
\end{equation}
where \(\nabla_{\mathbf{x}} \Phi\) and \(\nabla_{\mathbf{x}}^2 \Phi\) represent the gradient and Hessian matrix of \(\Phi\) with respect to \(\mathbf{x}\), respectively, and so on.
Suppose an implicit representation task requires satisfying \(M\) constraint conditions \(\{\mathcal{C}_m(\mathbf{y}, \Phi(\mathbf{x}), \nabla \Phi(\mathbf{x}), \ldots)\}_{m=1}^M\). Then, \(\Phi(\mathbf{x})\) can then be obtained through gradient descent, with the objective function defined as,
\begin{equation}
    \resizebox{0.43\textwidth}{!}{
    $
    \mathcal{L}=\int_{\Omega} \sum_{m=1}^M \mathbf{1}_{\Omega_m}(\mathbf{x})\left\|\mathcal{C}_m(\mathbf{y}, \Phi(\mathbf{x}), \nabla \Phi(\mathbf{x}), \ldots)\right\| d \mathbf{x},$
    \label{eq:inr_loss}
    }
\end{equation}  where \(\mathbf{1}_{\Omega_m}(\mathbf{x})\) is an indicator function that equals 1 if \(\mathbf{x} \in \Omega_m\) and 0 otherwise.

If Eq. \eqref{eq:inr_loss} is an L2 norm (which holds for most implicit tasks), the network \(\Phi\) is trained with a learning rate \(\eta\), and the network's weights are initialized such that its output is close to zero. Under the NTK asymptotic conditions \cite{lee2019wide}, after \(t\) training iterations, the network's output for any test data \(\mathbf{X}_{\text{test}}\) can be approximated as \cite{tancik2020ffn},
\begin{equation}
    \Phi^{(t)}(\mathbf{X}_{\text{test}}) \approx \mathbf{K}_{\text{test}} \mathbf{K}^{-1}\left(\mathbf{I}-e^{-\eta \mathbf{K} t}\right) \mathbf{y},
\end{equation} where \(\mathbf{K}\) denotes the NTK matrix for training coordinates \(\mathbf{X}_{\text{train}}\), and \(\mathbf{K}_{\text{test}}\) is the NTK matrix for all coordinates in both the test and the training set.

When considering the training error \(\mathbf{e}_{\text{train}}=\Phi^{(t)}(\mathbf{X}_\text{train}) - \mathbf{y}\) of the network's predictions \(\Phi^{(t)}(\mathbf{X}_\text{train})\) on the training dataset at the \(t\)-th iteration, we notice that since the NTK matrix \(\mathbf{K}\) must be positive semi-definite, it can be spectrally decomposed as \(\mathbf{K} = \mathbf{Q} \boldsymbol{\Lambda} \mathbf{Q}^{\mathrm{T}}\), where \(\mathbf{Q}\) is an orthogonal matrix, and \(\boldsymbol{\Lambda}\) is a diagonal matrix with the eigenvalues \(\lambda_i \geq 0\) of \(\mathbf{K}\) on the diagonal. Thus, since \(e^{-\eta \mathbf{K} t} = \mathbf{Q} e^{-\eta \boldsymbol{\Lambda} t} \mathbf{Q}^{\mathrm{T}}\) holds, we have,
\begin{equation*}
        \mathbf{Q}^{\mathrm{T}}\mathbf{e}_{\text{train}} \approx \mathbf{Q}^{\mathrm{T}}\left(\left(\mathbf{I}-e^{-\eta \mathbf{K} t}\right) \mathbf{y}-\mathbf{y}\right)
        =-e^{-\eta \boldsymbol{\Lambda} t} \mathbf{Q}^{\mathrm{T}} \mathbf{y}.
\end{equation*} 
Then, the following corollary can be derived,
\begin{corollary}
\label{ntk_inr_corollary}
\textbf{Spectral Bias of INR Models. }
During training, each component of the training error \(| \mathbf{e}_{\text{train}}|_i\) decays exponentially, with the decay rate determined by the NTK eigenvalues \(\lambda_i\) and learning rate \(\eta\),
\[
|\mathbf{e}_{\text{train}}|_i \sim e^{-\eta \lambda_i t}.
\]
\end{corollary}
This indicates that larger eigenvalues in the NTK spectrum (corresponding to low-frequency features) converge more rapidly, while smaller eigenvalues exhibit slower learning, suggesting that the network exhibits spectral bias. Additionally, the width of the NTK spectrum (the range of eigenvalues) should be appropriately balanced, as excessively large or small spectral widths can lead to uneven convergence or underfitting. Building on these insights, we will next analyze the impact of network architecture, positional encoding, and nonlinear primitives on the NTK spectral distribution (eigenvalue distribution), thereby elucidating the properties of different INR models.

\subsection{Coordinate Models}
\label{sec:cn_model}
\textbf{Coordinate-MLPs.} Over the past three decades, researchers have developed various MLP-based neural networks to approximate \(\Phi(\mathbf{x})\), building upon Universal Approximation theorem \cite{cybenko1989approximation}. These methods are collectively referred to as Coordinate-MLPs, which are defined as,
\begin{equation}
\label{eq:mlp}
 \resizebox{0.43\textwidth}{!}{$
\begin{split}
    \Phi(\mathbf{x})_{\text{MLP}}=&\mathbf{W}_n\left(\boldsymbol{\Phi}_{n-1} \circ \boldsymbol{\Phi}_{n-2} \circ \ldots \circ\boldsymbol{\Phi}_0\right)(\gamma(\mathbf{x}))+\mathbf{b}_n, \\
    &\mathbf{x}_l \mapsto \boldsymbol{\Phi}_l\left(\mathbf{x}_l\right)=\sigma \left(\mathbf{W}_l \mathbf{x}_l+\mathbf{b}_l\right),
\end{split}$}
\end{equation} 
where $\boldsymbol{\Phi}_l: \mathbb{R}^{M_l} \mapsto \mathbb{R}^{N_l}$ represents the $l$-th layer of the network. It comprises an affine transformation defined by the weight matrix $\mathbf{W}_l \in \mathbb{R}^{N_l \times M_l}$ and biases $\mathbf{b}_l \in \mathbb{R}^{N_l}$ applied to the layer input $\mathbf{x}_l \in \mathbb{R}^{M_l}$. $\sigma(\cdot)$ denotes a non-learnable activation function, and $\gamma(\cdot)$ represents an optional positional encoding layer.

\textbf{Coordinate-KANs}. 
Unlike MLPs, which employ fixed activation functions at the nodes (``neurons"), KANs \cite{liu2024kan}, inspired by the Kolmogorov-Arnold representation theorem \cite{kolmogorov1957representation}, utilize learnable activation functions at the edges (``weights"). To validate the ability of KANs in implicit neural representations, we adopt nonlinear components similar to those used in MLPs to represent various signals. We refer to these models as Coordinate-KANs, which can be defined as,
\begin{equation}
\label{eq:kan}
\begin{split}
        \Phi(\mathbf{x})_{\text{KAN}}=& \left(\boldsymbol{\Phi}_{n-1} \circ \boldsymbol{\Phi}_{n-2} \circ \ldots \circ \boldsymbol{\Phi}_0\right)(\mathbf{x}), \\
    &\mathbf{x}_l \mapsto \boldsymbol{\Phi}_l\left(\mathbf{x}_l\right),
\end{split}
\end{equation} where $\boldsymbol{\Phi}_l$ denotes the $l$-th KAN layer of the network, which essentially represents the transformation matrix between the layer's input $\mathbf{x}_l$ and output $\mathbf{x}_{l+1}$. Assuming a KAN has a shape of $\left[n_0, n_1, \cdots, n_L\right]$, where $n_i$ denotes the number of neurons in the $i$-th layer of the computational graph. For the $i$-th neuron on the $l$-th layer $(l, i)$, the activation value is $x_{l,i}$. Between the $l$-th and $(l+1)$-th layers, there are $n_ln_{l+1}$ activation functions: the activation function connecting neuron $(l,i)$ with neuron $(l+1, j)$ can be denoted by $ \phi_{l, j, i}$.
The pre-activation of $\phi_{l, j, i}$ is simply $x_{l, i}$; the post-activation of $\phi_{l, j, i}$ is denoted by $\tilde{x}_{l, j, i}=\phi_{l, j, i}\left(x_{l, i}\right)$. The activation value of the $(l+1, j)$ neuron is simply the sum of all incoming post-activations,
$$
x_{l+1, j}=\sum_{i=1}^{n_l} \tilde{x}_{l, j, i}=\sum_{i=1}^{n_l} \phi_{l, j, i}\left(x_{l, i}\right), j=1, \cdots, n_{l+1} .
$$
Rewriting it under the matrix form of the $l$-th KAN layer $\mathbf{\Phi}_l$  will give,
$$
\small
\mathbf{x}_{l+1}=\underbrace{\left(\begin{array}{cccc}
\phi_{l, 1,1}(\cdot) & \phi_{l, 1,2}(\cdot) & \cdots & \phi_{l, 1, n_l}(\cdot) \\
\phi_{l, 2,1}(\cdot) & \phi_{l, 2,2}(\cdot) & \cdots & \phi_{l, 2, n_l}(\cdot) \\
\vdots & \vdots & & \vdots \\
\phi_{l, n_{l+1}, 1}(\cdot) & \phi_{l, n_{l+1}, 2}(\cdot) & \cdots & \phi_{l, n_{l+1}, n_l}(\cdot)
\end{array}\right)}_{\boldsymbol{\Phi}_l} \mathbf{x}_l.
$$ Due to the flexibility of this structure, different KAN networks can be constructed using various basis functions \(\phi\).

\textbf{Discussion:} Due to the superior interpolation capability of B-splines, KANs utilizing B-splines exhibit smaller spectral bias in low frequencies compared to ReLU-based MLPs (see Appendix \ref{sec:Spectral_Bias_of_KAN}). This suggests that KANs are more suitable for fitting low-frequency signals (e.g., symbolic regression) rather than implicitly representing high-frequency signals. Moreover, the inner functions (\(\phi\)) and the outer functions (summation) in KANs can be viewed as approximations to the kernel and regression functions in deep Gaussian processes with the covariance matrix of the inner functions omitted (see Appendix \ref{sec:KAN_is_a_Heuristic_Deep_Gaussian_Process}). Therefore, as a heuristic approach based on deep Gaussian processes, KANs are well-suited for modeling strong non-linearity and uncertainty; however, they also encounter inherent challenges associated with the high computational complexity and training difficulties of deep Gaussian processes.

\subsection{Positional Encoding}
\label{sec:pe}
Unlike high-dimensional machine learning tasks, implicit neural representations involve a low-dimensional regression task, where the inputs are dense low-dimensional coordinates (e.g., pixel coordinates) distributed in Euclidean space. When coordinates are directly fed as inputs to a Coordinate-MLP, NTK is a function of the dot product between its inputs and their norms \cite{basri2020frequency, bietti2019inductive, bordelon2020spectrum}. This results in the NTK exhibiting rotational invariance, but lacking translational invariance (i.e., it depends only on the relative differences between input points). However, for implicit tasks, it is crucial to model the data in a manner that is independent of the object or scene's location. Thus, translational invariance (or stationarity) becomes an essential property for such tasks.

As shown in Table \ref{tab:pe_kernel}, both \texttt{NeRF} \cite{mildenhall2020nerf} and \texttt{RFF} \cite{tancik2020ffn} employ Fourier feature mapping as the position encoding, which map the input coordinates to a hypersphere, thereby enabling the resulting NTK to exhibit translational invariance. Furthermore, \texttt{RFF} introduces random Gaussian noise (\(\mathbf{b}_i\sim \mathcal{N}(0, \sigma^2)\)), allowing for the adjustment of the Gaussian variance parameter (\(\sigma\)) to control the bandwidth of the NTK, thereby enhancing training speed and generalization ability. As stated in Corollary \ref{ntk_inr_corollary}, a ``wider'' kernel with a slower spectral falloff achieves faster training convergence for high frequency component. Nonetheless, from a signal processing perspective, using an excessively broad spectral kernel for signal reconstruction can lead to high-frequency aliasing artifacts. Therefore, \texttt{RFF} requires hyperparameter search across different tasks to determine an appropriate Gaussian variance.

\begin{table}
    \centering
    \caption{Positional encoding kernel. See Appendix \ref{sec:Details_of_Positional_Encoding} for details.}
\resizebox{0.48\textwidth}{!}{%
\begin{tabular}{ccc}
    \hline
    P.E. ($\gamma$) & Formula  & Kernel  $k_\gamma(\mathbf{x}_1, \mathbf{x}_2)$ \\
    \hline
    
    \texttt{Identity} & $\mathbf{x}$ & $\mathbf{x}_1^T\mathbf{x}_2$ \\
    \hline
    
    \texttt{NeRF} & $\left[\sin (2^{L-1} \pi \mathbf{x}), \cos (2^{L-1} \pi \mathbf{x}) \right]$ & $ \sum_{i=0}^{L-1} \cos ( 2^i \pi (\mathbf{x}_1 - \mathbf{x}_2))$\\
    \hline
    
    \texttt{RFF} & $\left[a_{L-1} \cos \left(2 \pi \mathbf{b}_{L-1}^{\mathrm{T}} \mathbf{x}\right), a_{L-1} \sin \left(2 \pi \mathbf{b}_{L-1}^{\mathrm{T}} \mathbf{x}\right)\right]$ & $\sum_{i=0}^{L-1} a_i^2 \cos \left(2 \pi \mathbf{b}_i^{\mathrm{T}}\left(\mathbf{x}_1-\mathbf{x}_2\right)\right)$ \\
    \hline

     \texttt{FKAN (Ours)} &  Eq. \eqref{eq:pe_fkan} & Eq. \eqref{eq:fkan_ntk} \\

    \hline
\end{tabular}
}
 \label{tab:pe_kernel}
 \vspace{-10pt}
\end{table}

\begin{table*}[!htbp]
    \centering
    \caption{NTK analysis of nonlinear primitives. More details are provided in Appendix \ref{sec:Details_of_Activation_Functions}. }
\resizebox{\textwidth}{!}{%
\begin{tabular}{cccccccc}
    \hline Nonlinearity ( $\sigma$/$\phi$ ) & Equation  & $\sigma^{\prime}$/$\phi^{\prime}$ & $\sigma^{\prime \prime}$/$\phi^{\prime \prime}$ & Fourier Trans. $\hat{\sigma}(k)$ & L-Lipschitz & L-Smooth \\
    \hline
    ReLU  & $\max (0, x)$ & $1$ & 0 & $-\frac{1}{k^2}+ i \pi \delta'(k)$ & $1$ & $\times$ \\
    \hline
    
    Sine& $
    \sin (\omega x)
    $  & $\omega \cos (\omega x)$ & $-\omega^2 \sin (\omega x)$ & $ -i \pi (\delta(\omega - k) - \delta(k + \omega))$ & $\omega$ & $\omega^2$ \\
    \hline

     Gaussian & $e^{-\frac{x^2}{2\sigma^2}}$ & $-\frac{x}{\sigma^2} e^{-\frac{x^2}{2\sigma^2}}$ & $\left(\frac{x^2}{\sigma^4} - \frac{1}{\sigma^2}\right) e^{-\frac{x^2}{2\sigma^2}}$ & $\sqrt{2\pi \sigma^2} \cdot e^{-\frac{1}{2} \sigma^2 k^2}$ &  $\frac{1}{\sigma}e^{-\frac{1}{2}}$ & $\frac{2}{\sigma^2} e^{-\frac{3}{2}}$ \\
    \hline
\end{tabular}
}
    \label{tab:nonlinear}
\vspace{-10pt}
\end{table*}

To address this issue, we observe that although KANs tend to learn low-frequency components and exhibit high computational complexity, KAN models that utilize Fourier series as basis functions can serve as a better positional encoding for Coordinate-MLPs. Such models not only maintain translation invariance but also exhibit the capability to adaptively learn components across different frequencies.

Specifically, as described in Sec. \ref{sec:cn_model}, when employing Fourier kernels as nonlinear basis functions and disregarding the bias term, a single layer of the Fourier Coordinate-KAN network can be expressed as,
\begin{equation}
    \label{eq:pe_kan_model}
    \Phi(\mathbf{x})=\sum_{i=1}^D \sum_{\omega=1}^{\Omega}\left(a_{i \omega}  \cdot \cos \left(\omega x_i\right) +  b_{i \omega} \cdot  \sin \left(\omega x_i\right)\right),
\end{equation} where \(D\) is the dimensionality of the input coordinate \(\mathbf{x}=[x_1, \cdots, x_D]^T\), and the Fourier coefficients \(a_{i \omega}\) and \(b_{i \omega}\) are trainable parameters. The hyperparameter \(\Omega\) represents the maximum frequency threshold, which controls the range of frequencies used in the Fourier series expansion.
Furthermore, Eq. \eqref{eq:pe_kan_model} can be rewritten as a Fourier feature mapping,
\begin{equation}
    \label{eq:pe_fkan}
    \small
    \begin{split}
        \gamma(\mathbf{x}) = & \left[ a_{11} \cos(\omega_1 \mathbf{x}_1), b_{11} \sin(\omega_1 \mathbf{x}_1), \dots, \right. \\
         & \left. a_{D \Omega} \cos(\omega_\Omega \mathbf{x}_D), b_{D \Omega} \sin(\omega_\Omega \mathbf{x}_D) \right],
    \end{split}
\end{equation} which leads to the kernel function induced by this mapping,
\begin{equation}
    \label{eq:fkan_ntk}
    \begin{split}
         k_{\gamma}(\mathbf{x}_1, \mathbf{x}_2) 
         &= \gamma(\mathbf{x}_1)^\top \gamma(\mathbf{x}_2) \\
        =
       &\sum_{i=1}^{D} \sum_{\omega=1}^{\Omega} \left( a_{i\omega}^2 + b_{i\omega}^2 \right) \cos\left( \omega (\mathbf{x}_{1,i} - \mathbf{x}_{2,i}) \right),
    \end{split}
\end{equation} where \(\mathbf{x}_{1,i} \) denotes the \( i \)-th component of the input \( \mathbf{x}_1 \). Clearly, this kernel is stationary and can be interpreted as a Fourier approximation of a kernel function: $\omega$ denotes the Fourier basis frequencies, while $a_{iw}$ and $b_{iw}$ are the corresponding Fourier series coefficients.

In our benchmark, as shown in Table \ref{tab:pe_kernel}, we primarily examine the impact of 3 types of positional encoding on the NTK (with \texttt{Identity} denoting no positional encoding). All three positional encodings map the input coordinates from Euclidean space to a hypersphere, thereby making the NTK translation-invariant, which enhances both model convergence and generalization. Unlike \texttt{NeRF}, both \texttt{RFF} and our proposed \texttt{FKAN} have the capability to adjust the bandwidth of the NTK, but only \texttt{FKAN} can adaptively adjust the bandwidth to learn different frequencies, thereby reducing the model's sensitivity to hyperparameters.

\subsection{Nonlinear Primitives}
\label{sec:nonlinear}
In Coordinate-MLPs, the representational capacity of the network primarily arises from the nonlinear activation functions \(\sigma(\cdot)\), whereas in Coordinate-KANs, this capacity is derived from the nonlinear basis functions \(\phi(\cdot)\). We collectively refer to these as nonlinear primitives. According to NTK theory \cite{jacot2018neuralntk}, given the network inputs \( \mathbf{x}_i \) and \( \mathbf{x}_j \), the corresponding elements of the NTK \( \mathbf{K}_{ij} \) in Coordinate-MLPs (without considering bias terms) and Coordinate-KANs are as follows (see notation in Sec. \ref{sec:cn_model}),
\begin{equation*}
    \begin{split} 
   \mathbf{K}_{ij}^{\text{MLP}} &= \sum_{l=0}^{n-1} \left( \prod_{k=l+1}^{n-1} \mathbf{W}_k^T \sigma'(\mathbf{W}_k \mathbf{x}_k) \right) \mathbf{W}_l^T \sigma'(\mathbf{W}_l \mathbf{x}_l), \\
   \mathbf{K}_{ij}^{\text{KAN}} &= \sum_{l=0}^{L-1} \sum_{k=1}^{n_l} \sum_{m=1}^{n_{l+1}} \left( \phi'_{l, m, k} (x_{l,k}) \cdot \phi'_{l, m, k} (x_{l, k}) \right), \\
    \end{split}
\end{equation*}
which reveal that the nonlinear primitives determine the gradient propagation properties between layers, thereby influencing the spectral distribution of the NTK. However, directly computing the NTK is exceedingly challenging \cite{han2022fast}. As a result, we indirectly explore the  spectral distribution by analyzing the characteristics of the nonlinear primitives associated with different activation functions.

Specifically, as shown in Table \ref{tab:nonlinear}, we analyze the following properties of the nonlinear primitives:

(1) Gradient and Curvature: The gradient and curvature directly influence the eigenvalue distribution of the NTK, particularly the spectral width of the eigenvalues. A larger gradient magnitude typically results in an increase in certain eigenvalues within the NTK, thereby enabling the network to learn and adjust more rapidly. In contrast, a larger curvature often induces stronger nonlinearity in the network's local kernel, which can lead to a broader spectral width of the NTK eigenvalues. For instance, the ReLU activation function, with a gradient of 1 in the positive domain, results in a NTK eigenvalue distribution with a larger amplitude, which effectively captures low-frequency features. However, due to its zero curvature, the NTK spectrum lacks smoothness and tends to concentrate in certain regions, making it challenging to capture high-frequency features.

(2) Fourier Transform: Fourier transform provides insight into the frequency-domain properties of nonlinear primitives, thereby reflecting the frequency components of NTK spectrum distribution. Given a nonlinear primitive \( \sigma(x) \), its Fourier transform is defined as $\hat{\sigma}(k) = \int_{-\infty}^{\infty} \sigma(x) e^{-i k x} dx$, where \( k \) represents the frequency components in the frequency domain. As shown in Table \ref{tab:nonlinear}, the frequency response of ReLU decays inversely with the square of the frequency, while Gaussian exhibits exponential decay. Consequently, Gaussian activation show a stronger response to low-frequencies, facilitating rapid convergence, but struggle with high-frequencies. In contrast, the frequency response of Sine is impulsive, meaning it can stably learn frequency signals within a specific range. However, this impulsive nature prevents Sine from responding to a broader range of frequencies, limiting its ability to perform multi-scale learning. This is why Sine is difficult to model neural radiance fields (see Table \ref{tab:benchmark_leaderboard} for experimental evidence.).

(3) L-Lipschitz and L-Smooth conditions: For any two inputs \(x_1\) and \(x_2\), a nonlinear primitive \(\sigma\) is L-Lipschitz continuous if $
|\sigma(x_1) - \sigma(x_2)| \leq L |x_1 - x_2|$, and L-smooth if $\| \nabla \sigma(x_1) - \nabla \sigma(x_2) \| \leq L \| x_1 - x_2 \|$, where \(L\) is the condition constant. The Lipschitz and smoothness properties collectively influence the spectral characteristics of the NTK. A larger Lipschitz constant generally increases the NTK eigenvalues, thereby accelerating network convergence, while a larger smoothness constant tends to mitigate excessive fluctuations in the eigenvalues, ensuring stable training \cite{nguyen2021tight}.  As shown in Table \ref{tab:nonlinear}, Sine exhibits stable Lipschitz and smoothness constants, which not only allow for a rapid response to low frequencies, but also ensure a relatively uniform learning rate, offering a significant advantage in learning low-dimensional high-frequency signals. In contrast, both the Lipschitz and smoothness constants of Gaussian are related to the derivative of the variance parameter $\sigma$, indicating potential instability when learning high-frequency components.

\section{Experiments}
\label{sec:experiments}

\subsection{INR Models for Benchmarking}
As presented in Table \ref{tab:benchmark_leaderboard}, Coordinate-MLPs consist of 56 implicit neural models, which are formed by the combination of 14 activation functions and 4 types of positional encoding. Moreover, Coordinate-KANs include 22 distinct models derived from Fourier, polynomial, wavelet, and radial basis functions. It is noteworthy that due to the high nonlinearity of the learnable KAN basis functions, we observe a significant increase in the number of model parameters upon the inclusion of positional encoding, leading to convergence difficulties. Therefore, additional positional encoding layers are not considered for the Coordinate-KANs. Further details on the Coordinate-KAN models can be found in Appendix \ref{sec:Basis_Functions_of_Coordinate-KANs} of the supplementary materials.

\subsection{INR Tasks for Benchmarking}
As shown in Table \ref{tab:benchmark_leaderboard}, to evaluate the models' ability to learn high and low frequencies, multi-scale learning, and inverse reasoning, we designed two types of problems—forward and inverse—comprising 9 tasks related to implicit representations of audio, images, 3D shapes, CT scans and neural radiance fields. Further details about the tasks (e.g, dataset) are provided in Appendix \ref{sec:Details_of_INR_Tasks}.
\begin{table*}[t!]
    \centering
    \caption{Benchmark leaderboard. Best result of each activation (row) with different PEs is in \textbf{bold}. Best results in each task (column) are highlighted as \colorbox{colorFst}{first}, \colorbox{colorSnd}{second} and \colorbox{colorTrd}{third}. Results above/below the double line: Coordinate-MLPs/Coordinate-KANs. Note that ``*'' denotes results in radiance field task based on NGP \cite{muller2022ngp}, not vanilla NeRF \cite{mildenhall2020nerf}.}
    
\resizebox{0.99\textwidth}{!}{%
    \begin{tabular}{ccc|ccccccccccc}
        \cline{1-14}
        \multirow{2}{*}{\makecell{Nonlinearity}} & \multirow{2}{*}{\makecell{ Formula  \\ $\sigma(\cdot)$ / $\phi(\cdot)$}} &  \multirow{2}{*}{ \makecell{P.E. \\ $\gamma(\cdot)$}}& \makecell{ Audio\\ Reg.} &  \makecell{ Image\\ Reg.} &  \makecell{ SDF\\ Reg.} &\makecell{Image\\Inp.} &  \makecell{Image\\ Super-Res.}  &  \makecell{ Image\\ Deno.} &  \makecell{ CT\\Recon.} & \multicolumn{3}{c}{ \makecell{Poisson Recon. \\ \texttt{ (PSNR$^\uparrow$})}} & \makecell{Radiance \\ Field}  \\

       & & & \texttt{ (SNR$^\uparrow$)} &\texttt{ (PSNR$^\uparrow$)} &\texttt{ (IoU$^\uparrow$) } & \texttt{ (PSNR$^\uparrow$)} &\texttt{ (PSNR$^\uparrow$)} &\texttt{ (PSNR$^\uparrow$)} &\texttt{ (PSNR$^\uparrow$)}  &  Img. &  Grad. &  Lap. & \texttt{ (PSNR$^\uparrow$)}  \\
    
        \cline{1-14}
        \multirow{4}{*}{\makecell{ReLU}} &\multirow{4}{*}{\makecell{$\max(0, x)$ }} & \texttt{Id.} & 0.02 & 22.99 & 0.947 & 14.43 & 18.03 & 24.61 & 18.78 & 5.58 & 13.03 & 12.37 &  26.37/32.31*  \\
        & & \texttt{NeRF} & 21.93 & 36.28 & 0.981 & 14.12 & 13.83 & 25.50 & \bf 20.06 & 12.18 & 11.78 & 9.12 &  \bf 27.04 \\
        & & \texttt{RFF} &  \bf 29.38 & \bf 33.65 & 0.990 &  19.71 & 27.34 & 24.39 & 14.66 &  \bf 12.58 & 13.41 & 9.26 & 26.27  \\
        & & \texttt{FKAN} & 17.20 & 35.61 & \bf 0.991 & \bf \nd 21.15 & \bf 27.64 & \bf 26.14 & 12.02 & 8.23 & \bf 24.72 & \bf 13.93 &  26.59 \\
        \cdashline{1-14}

        \multirow{4}{*}{\makecell{PReLU}} &\multirow{4}{*}{\makecell{$\begin{cases}x, \text{ if } x \geq 0 \\ ax, \text{ otherwise} \end{cases}$ }} & \texttt{Id.} & 0.00  & 22.01  & 0.905 & 14.05 & 17.57 & 21.76 & 16.54 & 4.62 & 13.33 & 12.37 & 9.27/\colorbox{colorSnd}{32.46*}   \\
        & & \texttt{NeRF} & 21.93 &  31.58 & 0.968 & 15.09& 14.01 & 26.00& \bf 18.75 & 12.21 & 11.77 & 8.81 &  \bf 27.04  \\
        & &  \texttt{RFF} & \bf 24.27 & 31.56 & 0.949 & 20.07 & \bf 26.90 & 24.82 & 13.54 & \bf 13.53 & 14.27 & 9.16 & 24.64 \\
        & & \texttt{FKAN} & 23.13 & \bf 32.92 & \bf 0.990 & \bf 20.91 & 16.60 & \bf 26.46 & 11.72 & 8.95 & \bf 24.30 & \bf 13.80 & 25.81  \\
        \cdashline{1-14}

        \multirow{4}{*}{\makecell{Sine}} &\multirow{4}{*}{\makecell{$\sin(\omega x)$ }} & \texttt{Id.} &  18.61 & 23.47  & \bf 0.992 & 15.68 & \bf 15.52 & \bf 26.83 & \bf \nd 27.62 & \bf 13.59 & \bf 23.58 & \bf 14.91 & \textbf{7.92}/14.54*  \\
        & &\texttt{NeRF} & \bf \nd 42.17 & 28.27 & 0.974 & 13.30 & 12.51 & 21.62 & 18.42 & 12.23 & 9.36 & 9.52 & 7.25  \\
        & & \texttt{RFF} & 29.58 & 26.85 & NA &  \bf 19.85 & 11.90 & 21.48 & 16.45 & 11.57 & 21.68 & 12.43 & 7.90 \\
        & & \texttt{FKAN} & \rd 35.86 & \bf 43.34 & 0.990 & 18.90 & 11.92 & 24.16 & 15.76 & 12.17 & 9.66 & 9.72 & 7.66 \\
        \cdashline{1-14}

        \multirow{4}{*}{\makecell{ScaledSine}} &\multirow{4}{*}{\makecell{$a\sin(\omega bx +c)+d$ }} & \texttt{Id.} & 20.94 & \nd 44.44 & \nd 0.995 & \bf \fs 22.44 & \bf 18.56 & \bf 26.31 & \bf \rd 26.57 & \bf 12.60 & \bf \nd 36.78 & \bf  \nd 15.28 & 7.75/30.17*  \\
        & & \texttt{NeRF} & 26.42 & \bf \fs 44.74 & \bf \fs 0.996 & 13.31 & 13.62 & 21.45  & 16.21 & 12.19 & \rd 36.06 & 11.31 &  \bf 8.18 \\
        & & \texttt{RFF} & \bf \fs 45.29 & \rd 43.80 & 0.609 & 20.38 & 11.91 & 21.20 & 13.04 &  11.50 & 27.70 & 7.37 & 7.78 \\
        & & \texttt{FKAN} & 35.79 & 42.14 & 0.991 & 13.30 & 11.92 & 23.74 & 18.17 & 11.95 & 9.59 & 9.51 & 8.01  \\
        \cdashline{1-14}

         \multirow{4}{*}{\makecell{Gaussian}} &\multirow{4}{*}{\makecell{$ae^{-bx^2}$ }} & \texttt{Id.} & 5.25 & \bf 36.72 & 0.988 & 16.69 & \bf \fs 28.86 & 19.73 & 9.68 & 12.13 & 24.69 & 9.40 & \textbf{25.71}/32.27* \\
        & & \texttt{NeRF} & 16.32 & 28.10 & NA & 12.87 & 11.13 & 26.28 & 9.54 & 12.17 & 9.95 & 9.01 & 23.66 \\
        & &  \texttt{RFF} & 9.48 & 23.34 & NA & 13.05 & 11.18 & 22.27 & 24.20 &  12.16 & 11.22 & 10.50 & 12.21 \\
        & & \texttt{FKAN} &  \bf  31.44 & 34.70 & \bf \rd 0.993 & \bf 21.03 & \nd 28.06 &\bf \fs 30.29 & \bf \fs 28.02 &  \bf \rd 14.18 & \bf 30.22 & \bf 13.23 & 10.03 \\
        \cdashline{1-14}

         \multirow{4}{*}{\makecell{Laplacian}} &\multirow{4}{*}{\makecell{$ae^{-b|x|}$ }} & \texttt{Id.} & 10.92 & \bf 38.30 & \bf 0.981 & \bf 19.09 &  \bf 27.80 & \bf 23.46 & 9.70 & 12.18 & 10.66 & 10.44 & 25.39/28.55*  \\
        & & \texttt{NeRF} & 0.69 & 26.61 & NA & 13.31 & 11.87 & 19.54 & \bf 11.61 & \bf 12.23 & 10.16 & 10.17 & \bf 25.92  \\
        & & \texttt{RFF} & 0.24 & 21.38 & NA & 13.31 & 11.87 & 18.23 & 9.86 &  12.01 & 10.62 & 10.18 &  12.67 \\
         & & \texttt{FKAN} & \bf 26.56 & 12.21 & NA & 13.34 & 11.90 & 15.43 & 8.51 & 12.00 & \bf 11.54 & \bf 10.68 & 10.00  \\
        \cdashline{1-14}

        \multirow{4}{*}{\makecell{SuperGaussian}} & \multirow{4}{*}{\makecell{$e^{-(\frac{|x|}{\sigma})^{2n}} $}} & \texttt{Id.} & 5.15 & \bf 36.19 & \bf 0.988 & \bf 16.55 & \bf 22.41 & \bf 20.87 & \bf 10.37 & 12.16 & \bf 27.64 & 9.31 & \textbf{25.71}/30.15* \\
        & & \texttt{NeRF}  & 17.11 & 28.30 & NA &  13.34 & 11.90 & 20.19 &  7.68 & 11.85 & 10.05 & 9.27 & 23.66  \\
        & & \texttt{RFF} & 9.31 & 27.43 & NA &  13.34 &  11.90 & 20.11 &6.65 & 12.20 & 11.29 & 10.56 & 12.21 \\
         & & \texttt{FKAN} & \bf 25.45 & 12.21 & NA & 13.34 & 11.91 & 15.45 & 7.31 & \bf 12.18 & 12.13 & \bf 11.10 & 10.03  \\
        \cdashline{1-14}

         \multirow{4}{*}{\makecell{Gabor}} &\multirow{4}{*}{\makecell{$e^{-bx^2}\cos(ax)$ }} & \texttt{Id.} & 0.00 & 23.26 & 0.938 & 17.27 & 17.54 & 22.36 & \bf 18.04 & \bf 11.78 & 18.25 & \bf 13.47 & NA/31.90*  \\
        & & \texttt{NeRF} & 19.49 & 34.20 & 0.985 & 18.51 & 13.27 &\bf  25.15 & 18.01 & 10.24 & 20.27 & 12.46 & \bf \colorbox{colorSnd}{29.29} \\
        & &  \texttt{RFF} & 29.05 &  29.85 &  0.969 & 20.11 & 12.11 & 22.79 & 13.69 &  4.62 & \bf 25.70 & 9.01 & NA \\
         & & \texttt{FKAN} & \bf 24.86 & \bf 34.21 &\bf  0.991 & \bf 20.54 & \bf 27.73 & 24.27 & 13.63 & 11.49 & 17.81 & 10.02 & 26.58  \\
        \cdashline{1-14}

         \multirow{4}{*}{\makecell{Sinc}} &\multirow{4}{*}{\makecell{$\frac{\sin(\pi x)}{\pi x}$ }} &\texttt{Id.} & 0.00 & 24.18 & 0.956 &  18.76 & 15.70 & \bf 25.67 & \bf 19.65 & \bf 13.21 & \bf 11.96 & \bf 9.87 & 6.12/9.68*  \\
        & & \texttt{NeRF} & \bf 31.11 & \bf 34.14 & 0.989 & 10.77 & 9.56 & 20.26 & 13.72 & 12.24  & 9.59 & 9.19 & \bf 7.73   \\
        & & \texttt{RFF} & 25.09 & 28.68 & 0.700 &  13.10 & 12.83 & 20.55 & 9.69 & 12.20 & 9.52 & 8.99 & 7.25 \\
         & & \texttt{FKAN} & 25.31 & 32.97 & \bf 0.990 & \bf 18.78 & \bf 27.54 & 23.84 & 8.84 & 12.17 & 9.57 & 9.39 & 6.22  \\
        \cdashline{1-14}

        \multirow{4}{*}{\makecell{ExpSin}} &\multirow{4}{*}{\makecell{$e^{\sin(ax)}$ }} & \texttt{Id.} & 0.00 & 22.36 & 0.912 & 13.79 & 11.84 & 22.49 & \bf 18.03 & 5.58 & 16.27 & \bf \bf 12.59 & 24.47/32.34* \\
        & & \texttt{NeRF} & 16.99 & 30.93 & 0.973 & 13.13 & 8.93 & \bf 26.10 & 13.14 & 7.80 & \bf 10.55 & 9.51 & 28.31   \\
        & & \texttt{RFF} &\bf  24.30 & 30.65 & 0.957 & 18.74 & 6.49 & 23.10 & 6.64 &  \bf 10.55 & 11.32 & 9.22 & 25.70  \\
         & & \texttt{FKAN} & 13.19 & \bf 31.16 &\bf  0.990 & \bf 20.59 & \bf 13.01 & 25.63 & 6.47 & 4.64 & \bf 29.74 & 9.92 & \bf 28.32 \\
        \cdashline{1-14}

         \multirow{4}{*}{\makecell{Sigmoid}} &\multirow{4}{*}{\makecell{$\frac{1}{1 + e^{-x}}$ }} & \texttt{Id.} & 0.00 & 18.88  & 0.770 & 13.72 & 14.51 & 17.77 & 12.79 & 7.86 & 13.13 & 11.37 & 23.49/ \colorbox{colorFst}{32.50*}  \\
        & & \texttt{NeRF} & 0.95 & 23.03 & 0.898 & 15.69 & 12.78 & 24.48 & 15.64 &  4.62 & 12.42 & 9.12 & 27.19 \\
        & &  \texttt{RFF} & 6.40 &  23.23 & 0.922 & 17.88 & 13.76 & 20.65 &11.63 &  5.58 & 13.41 & 9.83 & 22.84  \\
        & & \texttt{FKAN} & \bf 18.93 & \bf 33.34 & \bf 0.992 & \bf \rd 21.20 & \bf \rd 27.88 & \bf 25.63 & \bf 26.07 & \bf 12.83 & \bf 31.53 & \bf \fs 15.44 & \bf 27.88  \\
        \cdashline{1-14}

        \multirow{4}{*}{\makecell{Tanh}} &\multirow{4}{*}{\makecell{$\frac{e^{x} - e^{-x}}{e^{x} + e^{-x}}$ }} & \texttt{Id.} & 0.00 & 22.32  & 0.908 & 14.67 & 13.21 & 21.86 & 15.68 & 14.22 & 16.37 & \bf 13.05 & 25.45/\colorbox{colorTrd}{32.45*}  \\
        & & \texttt{NeRF} & 14.57 & 32.61 & 0.976 & 16.84 & 10.81 & 25.71 & \bf 21.07&  12.26 & 14.40 & 9.44 & \bf \colorbox{colorFst}{29.78}\\
        & &  \texttt{RFF} & \bf 21.95 & 31.51 & 0.972 &  20.59 & \bf 26.39 & 25.30 & 19.70 &  \bf \nd 15.64 & \bf \fs 38.39 & 10.31 & 25.86  \\
        & & \texttt{FKAN} & 18.71 & \bf 32.80 & \bf 0.991 & \bf 21.05 & 14.62 &\bf  26.51 & 13.70 & 10.09 & 23.81 & 12.47 & 28.36  \\
         \cdashline{1-14}

         \multirow{4}{*}{\makecell{Quadratic}} &\multirow{4}{*}{\makecell{ $\frac{1}{1+(ax)^2}$}} & \texttt{Id.} & 13.78 & 26.66 & 0.967 & 14.44 & 16.05 & 21.05 & 14.70 & \bf \fs 17.20 & \bf 27.02 & \bf \rd 15.06 &   26.30/31.66*  \\
        & & \texttt{NeRF} & 0.25 & 28.93 & 0.969 & 17.46 & 13.59 & \bf \rd 28.07 & 17.67 & 12.23 & 9.62 & 9.06 &  \bf \colorbox{colorTrd}{28.80}   \\
        & &  \texttt{RFF} & 0.06 &  27.50 & 0.139 & \bf 20.11 & 11.21 & 24.71 & 14.06 &  12.01 & 22.80 & 9.45 & 25.33 \\
         & & \texttt{FKAN} & \bf 23.81 & \bf 35.73 &\bf  0.989 & 20.04 & \bf 27.24 & 26.38 & \bf 19.36 & 12.23 & 10.74 & 27.98 & 27.86   \\
         \cdashline{1-14}

        \multirow{4}{*}{\makecell{MultiQuadratic}} &\multirow{4}{*}{\makecell{ $\frac{1}{\sqrt{1+(ax)^2}}$ }} & \texttt{Id.} & 3.16 & 23.04 & 0.929 & 13.70 & 7.42 & 18.16 & 13.85 & 4.62 & 15.09 & \bf 12.31 &  25.28/31.64*  \\
        & & \texttt{NeRF} & 12.49 & 30.61 & 0.971 & 17.43 & 12.38 & 25.81 &  17.26 & \bf 12.05 & 11.81 & 8.98 &  \bf 28.50 \\
        & &  \texttt{RFF} & 10.18 & 26.76 & 0.917 & 19.10 & 18.74 & 20.71 & 16.42 &  9.88 & \bf 19.97 & 8.89 &  24.79 \\
        & & \texttt{FKAN} &\bf  26.83 & \bf 33.79 &\bf  0.991 & \bf 20.30 & \bf 27.46 & \bf \nd 28.23 & \bf 24.40 & 11.39 & 9.97 & 9.53 & 27.59 \\
        \hline
        \hline
        
        B-Spline & Eq.\eqref{eq:bspline} & \texttt{Id.} & NA & 21.99 & 0.981 & 17.00 & \rd 19.16 & 30.59 & 20.69 & 10.57 & 17.26 & 13.08 & 32.33*   \\

        Chebyshev & Eq. \eqref{eq:chebyshev1} & \texttt{Id.} & NA & 24.52 & \fs \bf 0.993 & 19.62 & 12.42 & 30.20 & \rd 21.89 & \nd 11.02 & \rd 26.60 & \fs \bf 14.51 & 28.56*       \\
        
        Chebyshev2 & Eq. \eqref{eq:chebyshev2} & \texttt{Id.} & NA & \rd 27.73 & NA & 20.62 & 11.51 & 30.05 & \fs \bf 24.70 & 9.94 & \fs \bf 45.58 & \nd 14.46 & 28.53*    \\

        Gegenbauer & Eq. \eqref{eq:gegenbauer} & \texttt{Id.} & NA & 12.20 & NA & 13.26 & NA & 15.43 & 11.49 & NA & NA & NA & 28.39*   \\

         Hermite & Eq. \eqref{eq:hermite} & \texttt{Id.} & NA & \nd 28.35 & 0.104 & 13.25 & NA  & 15.42 & 11.49 & NA & NA & NA & 27.58*   \\
         
        Jacobi & Eq. \eqref{eq:jacobi} & \texttt{Id.} & NA & 24.21 & NA &  \fs \bf 21.93 &  10.97 &  \rd 30.76 & \nd 23.60 & \fs \bf 15.21 & \nd 26.91 & 14.73 & 27.88*   \\
        
        Laguerre & Eq. \eqref{eq:laguerre} & \texttt{Id.} &NA & 21.98 & 0.976 & 16.07 &  11.66 & \fs \bf 32.15 & 18.83 & 6.08 & 18.43 & \rd 13.77 & 27.39*    \\
        
        Legendre & Eq. \eqref{eq:legendre} & \texttt{Id.} &NA & 22.75 & \nd 0.992 & \nd 21.07 & 4.99  & \nd 31.20 & 21.64 & 8.50 & 17.68 & 13.35 & 26.64*   \\

         Tayler & Eq. \eqref{eq:taylor} & \texttt{Id.} &NA & 20.10 & 0.104 & NA  & 11.89  & NA & NA & NA & NA & NA & 12.03* \\

        Bessel & Eq. \eqref{eq:bessel} & \texttt{Id.} & NA & 24.15 & 0.104 & NA & NA & NA & NA & NA & NA & NA & 25.79*  \\

        Fibonacci & Eq. \eqref{eq:fibonacci} & \texttt{Id.} & NA & 21.50 & 0.962 & 13.87 & 10.49 & 21.49 & 16.56 & 4.61 & 13.36 & 12.03 & 28.30*     \\

        Lucas & Eq. \eqref{eq:lucas} & \texttt{Id.} &NA & 21.96 & 0.932 &  11.66 & 9.82 & 14.39 & 11.49 & NA & NA & NA & 27.95*  \\

        Fourier & Eq. \eqref{eq:fourier} & \texttt{Id.} & \fs \bf 31.60 & \fs \bf 33.56 & 0.980 & 17.75 & \fs \bf 27.42 & 23.22 & 6.37 & \rd 10.86 & 10.29 & 9.44 & 31.72*    \\

        Sine & Eq. \eqref{eq:sine} & \texttt{Id.} &12.34 & 21.20 & 0.967 & 13.44 & 10.31 & 26.64 & 12.72 & 4.61 & 9.76 & 6.61 & 27.34*   \\
         
        MexicanHat & Eq. \eqref{eq:mexicanhat} & \texttt{Id.} &NA & 24.07 & 0.980 & 15.10  & 11.37 & 28.85 & NA & NA & NA & NA & 31.23*    \\
        Meyer & Eq. \eqref{eq:meyer} & \texttt{Id.} &NA &  24.52 & \rd 0.991 & 20.30 & 9.90 & \rd 30.76 & NA & NA & NA & NA & 11.91*  \\
        Morlet & Eq. \eqref{eq:morlet} & \texttt{Id.} &NA & 12.21 & \fs \bf 0.993 & \rd 20.74 & \nd 26.23 & 29.60 & NA & NA & NA & NA & 13.06*   \\
        DoG & Eq. \eqref{eq:dog} & \texttt{Id.} & NA & 22.96 & 0.976 &  14.50 & 15.81 & 27.35 & NA & NA & NA & NA & \fs \bf 32.59*  \\
        Shannon & Eq. \eqref{eq:shannon} & \texttt{Id.} &NA & 12.21 & 0.952 & 13.74  & 14.85 & 25.55 & NA & NA & NA & NA & 9.15*\\
        
        BSRBF & Eq. \eqref{eq:bsrbf} & \texttt{Id.} & NA & 12.48 & NA & 13.35 & 10.58 & 15.73 & 11.43 & 10.18 & 11.85 & 10.75 & 31.56*   \\
       
        GRBF & Eq. \eqref{eq:grbf} & \texttt{Id.} & NA & 22.31 & 0.961 & 14.18 & 15.22 & 26.36 & 13.88 & 9.16 & 14.52 & 11.84 & \rd 32.39*  \\

        RBF & Eq. \eqref{eq:rbf} & \texttt{Id.} &NA & 21.61 & 0.962 & 14.03 & 14.98 & 26.05 & 13.76 &  5.63 & 14.31 & 11.47 & \nd 32.57*   \\
        \cline{1-14}
    \end{tabular}
}
    \label{tab:benchmark_leaderboard}
    \vspace{-10pt}
\end{table*}

\subsubsection{Signal Representation}
The implicit representation of signals involves using a neural network function $\Phi$ to directly fit raw signals such as images, audio, and 3D occupancy volumes. In these tasks, the supervised loss in Eq. \eqref{eq:inr_loss} can be directly expressed as the discrepancy between the ground truth signal $f(\mathbf{x})$ and the predicted signal $\Phi(\mathbf{x})$: $\mathcal{L} = \int_{\Omega}\|\Phi(\mathbf{x}) - f(\mathbf{x})\|d \mathbf{x}$. By learning a continuous implicit representation from discrete signals, this method can be applied to tasks like signal super-resolution, denoising, and compression.

\textbf{Audio Regression.} The audio regression network \(\Phi_{\Theta}: t \rightarrow a\) takes timestamp coordinates \(t\) as input and outputs the corresponding amplitude \(a\). Given the high-frequency and locally periodic nature of audio signals, this task serves as an effective benchmark for evaluating the model's ability to learn high-frequency components.

\textbf{Image Regression.}  The image regression network $\Phi_{\Theta}: (u,v) \rightarrow (r,g,b)$ takes pixel coordinates $(u,v)$ as input and outputs the corresponding color values $(r,g,b)$. Images contain a substantial amount of low-frequency information and a smaller proportion of high-frequency components, making them an effective benchmark for assessing the model's ability to integrate both high-frequencies and low-frequencies.

\textbf{3D Shape Regression.} In this task, we utilize the Signed Distance Functions (SDF) \cite{Park2019DeepSDFLC} to model the surface of a 3D object. The SDF network \(\Phi_{\Theta}: (x, y, z) \rightarrow \mathbb{R}\) takes a 3D coordinate \((x, y, z)\) as input and outputs a scalar value representing the signed distance to the nearest surface. This task serves as an effective evaluation of the model's interpolation and boundary representation abilities.

\subsubsection{Inverse Problems}
Inverse problems typically involve inferring the internal parameters, states, or structure of a system from indirect observational data. Given that the solution to inverse problems is often affected by noise, incomplete data, or measurement errors, these tasks provide an effective means of evaluating a model's ability to perform inverse mapping, multi-scale reasoning and generalization.

\textbf{Image Inpainting.} Image inpainting involves learning an implicit representation from a sparse set of pixels (e.g., 20\%), enabling the inference of the majority of unobserved pixel information (e.g., 80\%). This task can effectively assesse the model's generalization ability.

\textbf{Image Super-Resolution.} This task aims to generate high-resolution images (e.g., with a 4X Res.) from low-resolution inputs, providing an effective evaluation of the model's interpolation and multi-scale learning capabilities.

\textbf{Image Denoising.} This task aims to recover the original clean image from a noisy input. During training, implicit network $\Phi$ acts as a low-pass filter, attenuating high-frequency noise while preserving low-frequency information. Therefore, this task effectively evaluates the model's ability to learn low-frequency features.

\textbf{Poisson Reconstruction \cite{sitzmann2020siren}.}
This task reconstructs images by solving the Poisson equation with the derivatives of the network \(\Phi(x)\). Specifically, the model is indirectly supervised by real gradients $\nabla_{\mathbf{x}} f(\mathbf{x})$ or Laplacians $\Delta f(\mathbf{x})$ to optimize the reconstruction. The supervision loss in Eq. \eqref{eq:inr_loss} can be expressed as, \(\mathcal{L}_{\text{grad.}} = \int_{\Omega}\left\|\nabla_{\mathbf{x}} \Phi(\mathbf{x}) - \nabla_{\mathbf{x}} f(\mathbf{x})\right\| d \mathbf{x}\) or \(\mathcal{L}_{\text{lapl.}} = \int_{\Omega}\|\Delta \Phi(\mathbf{x}) - \Delta f(\mathbf{x})\| d \mathbf{x}\). This task needs reconstructing a dense image from sparse, high-frequency gradients and Laplacians, thereby providing an effective assessment of the model's inverse reasoning and convergence.

\textbf{CT Reconstruction.} The Computed Tomographic reconstruction network takes 2D pixel coordinates as input and predicts the corresponding 3D volume density. The network is trained by minimizing the discrepancy between the integral projections of the predicted density field and those of ground-truth. Given that this task requires inferring the complete density field from limited projection data, the network must possess the ability to handle high-dimensional data and deduce global structures from partial information.

\textbf{Neural Radiance Fields.}
Neural Radiance Fields \cite{mildenhall2020nerf} is an inverse rendering problem that reconstructs a dense 3D scene from a sparse set of 2D images. Specifically, the radiance field network \(\Phi_{\Theta}: (x, y, z, \mathbf{d}) \rightarrow (r, g, b, \sigma)\) takes the 3D spatial coordinates \((x, y, z)\) and the viewing direction \(\mathbf{d}\) as input, predicting the color \((r, g, b)\) and density \(\sigma\) at each point. It is noteworthy that while the purely implicit, vanilla NeRF \cite{mildenhall2020nerf} effectively evaluates the multi-scale learning and complex representation capabilities of various models, our experiments reveal that the low-frequency characteristics of Coordinate-KANs lead to poor convergence. As a result, we utilize the NGP \cite{muller2022ngp} model to investigate the potential of KANs compared to MLPs within NGP radiance fields. Unlike vanilla NeRF, NGP stores the scene in a hash table without requiring additional positional encoding, with the network serving as a feature extractor that maps high-dimensional features to color and density.

\subsection{Benchmark Leaderboard and Insights}
\label{sec:benchmark_leaderboard_insights}
Based on the benchmark in Table \ref{tab:benchmark_leaderboard}, we highlight several key insights (see Appendix \ref{sec:Insights_One_by_One} for insights one by one).

\textbf{Insights of Coordinate Models.} 
As analyzed in Sec. \ref{sec:cn_model}, KAN exhibits a smaller spectral bias in the low-frequency domain compared to MLP, making it effective for low-frequency image denoising task. For instance, Laguerre-KAN outperforms the runnerup FKAN+Gaussian (MLP) by 1.86 dB. Furthermore, as a heuristic Gaussian process, KAN is well-suited for uncertainty modeling and inverse reasoning, showing strong performance in Poisson reconstruction task, where Chebyshev2-KAN outperforms ScaledSine-MLP by  8.8 dB. However, the Gaussian process property of KAN also lead to high computational complexity and challenges in training, which can hinder convergence in certain tasks. In addition, as a high-dimensional feature extractor in NGP radiance field task, KAN has slower training times but comparable performance to MLP. Overall, due to its positional encoding and fast training, MLP remains the most effective model for implicit  tasks.

\textbf{Insights of Positional Encoding.}
In forward problem-solving, positional encoding can enhance the model's ability to learn high-frequencies by adjusting the spectral width of the NTK. For instance, \texttt{NeRF} significantly improves the performance of Sine-type activations in tasks such as audio regression ($\approx$ 14dB $\uparrow$), image regression ($\approx$ 3dB $\uparrow$), and SDF regression ($\approx$ 0.002 IoU $\uparrow$). However, in inverse problem-solving, \texttt{NeRF} and \texttt{RFF} don't lead to significant improvements due to the manually set frequencies, which lack generalizability. Our proposed \texttt{FKAN}, as a learnable positional encoding, outperforms both \texttt{NeRF} and \texttt{RFF} in generalization. Specifically, \texttt{FKAN} notably improves the inverse reasoning and high-frequency learning capabilities of Gaussian-type and Quadratic-type activations. However, in complex radiance field task, the intricate frequencies are challenging to model using the frequency parameters of positional encoding. As a result, the fixed frequency mapping in \texttt{NeRF} is more conducive to stable training.

\textbf{Insights of Nonlinear Primitives.} 
In Coordinate-MLPs, the Sine-type activations exhibit fixed frequency parameters for both Lipschitz and Smoothness constants, enabling the model to uniformly learn various frequency components of audio and image signals. Therefore, Sine significantly outperform suboptimal methods in tasks such as audio regression ($\approx$ 14dB $\uparrow$), image regression ($\approx$ 6dB $\uparrow$), SDF regression ($\approx$ 0.003 IoU $\uparrow$) and image inpainting ($\approx$ 1.3dB $\uparrow$). However, the Fourier transform of Sine reveals that their frequency spectrum tends to concentrate at a specific frequency, lacking multi-scale learning capabilities. As a result, for neural radiance fields, which involve complex frequency distributions, Sine-MLPs struggle to effectively capture the diverse frequency components within the field, leading to a performance gap of approximately 17 dB compared to ReLU-MLPs. Gaussian-type activations exhibit an exponential decay in frequency response, making them more suitable for learning low-frequency components. Therefore, when combined with \texttt{FKAN}, they demonstrate strong inverse reasoning and generalization abilities, significantly outperforming other methods in tasks such as image super-resolution ($\approx$ 2dB $\uparrow$), image denoising ($\approx$ 3dB $\uparrow$), and CT reconstruction ($\approx$ 1dB $\uparrow$). For Coordinate-KANs, they exhibit limited cross-task generalization due to the variability caused by different basis functions, requiring task-specific adaptations. For example, while Chebyshev2-KAN achieves optimal performance in Poisson reconstruction, it struggles to converge in SDF regression.

\section{Conclusion}
We propose INR-Bench, the first benchmark for implicit neural representations, exploring the effects of model architecture, positional encoding, and nonlinear primitives on the NTK spectrum. Our results show that KAN excels in low-frequency spectral bias but faces challenges with high computational cost and training difficulty. Positional encoding improves high-frequency learning, while nonlinear primitives  influence frequency response. INR-Bench, evaluated on 9 tasks, highlights performance differences between 56 Coordinate-MLP and 22 Coordinate-KAN models, offering novel insights for future research endeavors.

\bibliographystyle{IEEEtran}
\bibliography{IEEEabrv,myrefs}

\clearpage
\appendices
\section*{Appendix}
In this appendix, we provide further details as follows:
\begin{itemize}
\item Sec. \ref{sec:Coordinate-KANs}: provides additional details on Coordinate-KANs.
    \begin{itemize}
        \item Sec. \ref{sec:Basis_Functions_of_Coordinate-KANs}: presents the explanation of 22 basis functions of Coordinate-KANs.
        \item Sec. \ref{sec:Spectral_Bias_of_KAN}: provides a formal proof of the spectral bias of KAN and MLP.
        \item Sec. \ref{sec:KAN_is_a_Heuristic_Deep_Gaussian_Process}: presents how KAN can be interpreted as a heuristic deep Gaussian process.
    \end{itemize}
\item Sec. \ref{sec:appendix_Coordinate-MLPs}: presents additional details regarding Coordinate-MLPs.
    \begin{itemize}
        \item Sec. \ref{sec:Details_of_Positional_Encoding}: presents positional encodings and their kernels, along with an analysis of their impact on the NTK.
        \item Sec. \ref{sec:Details_of_Activation_Functions}: presents the gradients, curvatures, Fourier transforms, and smoothness conditions of 14 activations.
    \end{itemize}

\item Sec. \ref{sec:Insights_One_by_One}: presents various insights derived from both analytical and experimental findings.
    \begin{itemize}
        \item Sec. \ref{sec:appendix_Coordinate_Models}: presents insights into the impact of model architectures on Implicit Neural Representations.
        \item Sec. \ref{sec:appebdix_Positional_Encoding}: provides insights into the impact of positional encoding on Implicit Neural Representations.
        \item Sec. \ref{sec:appendix_Nonlinear_Primitives}: offers insights into the impact of different nonlinear primitives on Implicit Neural Representations.
    \end{itemize}

\item Sec. \ref{sec:Details_of_INR_Tasks}: provides details for 9 INR tasks.

\item Sec. \ref{sec:Further_Experiments}: presents additional experimental details and results.
    \begin{itemize}
        \item Sec. \ref{sec:Experimental_Setup}: Experimental setup and evaluation metrics.
        \item Sec. \ref{sec:Parameter_Sensitivity_Analysis}: Experiments on the parameter sensitivity of positional encoding and nonlinear primitives.
        \item Sec. \ref{sec:Impact_of_Positional_Encoding_on_Coordinate-KANs}: Experiments on the impact of positional encoding on Coordinate-KANs.
        \item Sec. \ref{sec:Impact_of_Normalization_on_Coordinate-MLPs}: Experiments on the impact of normalization on Coordinate-MLPs.
        \item Sec. \ref{sec:Qualitative_Experiments}: Qualitative experiments.
    \end{itemize}
\end{itemize}

\section{Coordinate-KANs}
\label{sec:Coordinate-KANs}

\subsection{22 Basis Functions of Coordinate-KANs}
\label{sec:Basis_Functions_of_Coordinate-KANs}
Due to the ease of use of the KAN structure (Sec. \ref{sec:cn_model}), different basis functions can be employed to suit various tasks. In this section, we provide a detailed description of four types of basis functions—polynomial, Fourier, wavelet, and radial basis functions—comprising a total of 22 variations. These functions serve as supplementary information to Table \ref{tab:benchmark_leaderboard} in the main text.

\subsubsection{Polynomial Basis Functions}
\textit{Characteristics}: Polynomial basis functions are characterized by their strong analytical properties and computational efficiency, enabling effective function approximation in low-dimensional spaces. The degree of the polynomial can be adjusted flexibly, providing a means to control the model's complexity.

\textit{Applicable Tasks}: These functions are well-suited for tasks that involve smooth approximation of data, particularly when the data exhibit simple or low-dimensional relationships, such as in regression analysis and interpolation problems.

\textbf{B-Splines:} The B-splines are defined as,
   \begin{equation*}
   \begin{split}
        B_{i,p}(t) &= \frac{t - t_i}{t_{i+p} - t_i} B_{i,p-1}(t) + \frac{t_{i+p+1} - t}{t_{i+p+1} - t_{i+1}} B_{i+1,p-1}(t), \\
        B_{i,0}(t) &= 
   \begin{cases}
   1 & \text{if } t_i \leq t < t_{i+1}, \\
   0 & \text{otherwise,}
   \end{cases}
   \end{split}
   \end{equation*}
where \(p\) represents the order of the basis function. Increasing the order typically enhances the smoothness of the function, but also leads to an increase in computational complexity. The function \(B_{i,p}(t)\) denotes the B-spline basis function of order \(p\), defined over the interval \([t_i, t_{i+p+1}]\). \(t_i\) is the \(i\)-th node in the node vector \(\{t_0, t_1, \dots, t_m\}\), which is generally a non-decreasing sequence. And the parameter \(t\) lies within the domain and typically falls within the interval \([t_i, t_{i+p+1}]\).  For instance, in the original LiuKAN \cite{liu2024kan}, its basis functions can be expressed as,
\begin{equation}
    \begin{split}
        \phi(x) &= w_bsilu(x) + w_s spline(x) \\
        &= w_b \frac{x}{1+e^{-x}} + w_s \sum_{i}^Pc_iB_i(x) \\
    \end{split}
       \label{eq:bspline}
\end{equation} 
where \(P\) denotes the order of the B-spline, while \(w_b\), \(w_s\), and \(c_i\) are learnable parameters.

\textbf{Chebyshev (Type 1):}
The Chebyshev polynomials of the first kind are defined as:
 \begin{equation*}
   T_n(x) = \cos(n \cos^{-1}(x)), \quad -1 \leq x \leq 1,
   \end{equation*}
where \( n \) denotes the order of the polynomial. In the context of KAN, the basis functions can be expressed as a weighted sum of Chebyshev polynomials of different orders, as given by:
\begin{equation}
    \phi(x) = \sum_{n=0}^Pw_n T_n(x)
    \label{eq:chebyshev1}
\end{equation}
where \( w_n \) represents the learnable parameters associated with each Chebyshev polynomial \( T_n(x) \).

\textbf{Chebyshev (Type 2):}
The recurrence relations for the Chebyshev polynomials of the second kind, \( U_n(x) \), are given by:
\begin{equation*}
    \begin{split}
        U_0(x) &= 1, \\
        U_1(x) &= 2x, \\
        U_n(x) &= 2x U_{n-1}(x) - U_{n-2}(x) \quad \text{for} \quad n \geq 2.
    \end{split}
\end{equation*}
Subsequently, the basis function used in KAN can be expressed as a weighted sum of these Chebyshev polynomials:
\begin{equation}
\phi(x) = \sum_{n=0}^Pw_n U_n(x)
\label{eq:chebyshev2}
\end{equation} 
where \( w_n \) represents the learnable parameters associated with each Chebyshev polynomial \( U_n(x) \).

\textbf{Gegenbauer: }
The Gegenbauer polynomials \( C_n^{(\alpha)}(x) \) form a class of orthogonal polynomials defined by the following recurrence relation,
\begin{equation*}
    \begin{split}
        C_0^{(\alpha)}(x) &= 1, \\
        C_1^{(\alpha)}(x) &= 2 \alpha x, \\
        C_{n+1}^{(\alpha)}(x) &= \frac{2 (n + \alpha) x C_n^{(\alpha)}(x) - (n + 2\alpha - 1) C_{n-1}^{(\alpha)}(x)}{n + 1}. \\
    \end{split}
\end{equation*}
In the context of KAN, the basis functions are represented as a weighted sum of the Gegenbauer polynomials. Specifically, the basis function \( \phi(x) \) is expressed as:
  \begin{equation}
  \phi(x) = \sum_{n=0}^{P} w_n C_n^{(\alpha)}(x)
   \label{eq:gegenbauer}
   \end{equation}
where \( C_n^{(\alpha)}(x) \) represents the \( n \)-th order Gegenbauer polynomial with parameter \( \alpha \), \( w_n \) are the learnable coefficients associated with each Gegenbauer polynomial \( C_n^{(\alpha)}(x) \), and \( P \) denotes the maximum polynomial degree.

\textbf{Hermite: }
The Hermite polynomials \( H_n(x) \) are defined by the recurrence relation,
\[
H_0(x) = 1, \quad H_1(x) = 2x
\]
For \( n \geq 2 \), the recurrence relation is given by,
\[
H_n(x) = 2x H_{n-1}(x) - 2(n-1) H_{n-2}(x).
\]
The basis function in KAN can be computed as a weighted sum of Hermite polynomials,
\begin{equation}
   \phi(x) = \sum_{n=0}^{P} w_n H_n(x)
   \label{eq:hermite}
\end{equation}
where \( H_n(x) \) is the \( n \)-th order Hermite polynomial, \( w_n \) is the learnable coefficient for the \( n \)-th Hermite polynomial, and \( P \) is the maximum degree of the Hermite polynomials used.

\textbf{Jacobi:}
Jacobi polynomials \( P_n^{(\alpha, \beta)}(x) \) are defined by the recurrence relation,
\begin{equation*}
\begin{split}
    P_0^{(\alpha, \beta)}(x) &= 1, \\
     P_1^{(\alpha, \beta)}(x) &= \frac{1}{2} \left( (\alpha - \beta) + (\alpha + \beta + 2)x \right). \\
\end{split}
\end{equation*} For \( n \geq 2 \), the recurrence relation is,
\begin{equation*}
    \begin{split}
        P_{n+1}^{(\alpha, \beta)}(x) &= \left( \frac{(2n + \alpha + \beta)(2n + \alpha + \beta - 1)}{2n(n + \alpha + \beta)} \right) x P_n^{(\alpha, \beta)}(x) \\ 
        & - \left( \frac{(2n + \alpha + \beta - 1)(\alpha^2 - \beta^2)}{2n(n + \alpha + \beta)(2n + \alpha + \beta - 2)} \right) P_{n-1}^{(\alpha, \beta)}(x),
    \end{split}
\end{equation*}
where \( \alpha \) and \( \beta \) are parameters, and \( x \) is the input normalized to the range \([-1, 1]\). The basis function in KAN can be computed as a weighted sum of the Jacobi polynomials,
   \begin{equation}
  \phi(x) = \sum_{n=0}^{P} w_n P_n^{(\alpha, \beta)}(x),
   \label{eq:jacobi}
   \end{equation}
where \( w_n \) are the learnable coefficients associated with each Jacobi polynomial \( P_n^{(\alpha, \beta)}(x) \), and \( P \) denotes the maximum degree of the Jacobi polynomial.

\textbf{Laguerre:}
The Laguerre polynomials \( L_n^\alpha(x) \) are calculated using the recurrence relation,
\begin{equation*}
    \begin{split}
        L_0^\alpha(x) &= 1, \\
        L_1^\alpha(x) &= 1 + \alpha - x, 
    \end{split}
\end{equation*} where $\alpha$ is the parameter that defines the generalized Laguerre polynomials. For \( n \geq 2 \), each polynomial is computed using the recurrence relation,
\[
L_{n+1}^\alpha(x) = \frac{(2n + 1 + \alpha - x) L_n^\alpha(x) - (n + \alpha) L_{n-1}^\alpha(x)}{n+1}.
\]
Then, the basis function in KAN is computed as a weighted sum of the Laguerre polynomials,
\begin{equation}
    \phi(x)= \sum_{n=0}^{P} w_n L_n^\alpha(x),
     \label{eq:laguerre}
\end{equation} where each term \( L_n^\alpha(x) \) is computed using the above recurrence relation, and \( w_n \) are the learnable weights.

\textbf{Legendre:}
Legendre polynomials \( P_n(x) \) are defined using the following recurrence relation,
\begin{equation*}
    \begin{split}
        P_0(x) &= 1, \\
        P_1(x) &= x,
    \end{split}
\end{equation*} for \( n \geq 2 \), the recurrence relation is,
\[
P_{n+1}(x) = \frac{(2n + 1) x P_n(x) - n P_{n-1}(x)}{n + 1},
\]
where \( x \) is the input, and the polynomials are normalized within the range \([-1, 1]\). Then, the basis function in KAN can be computed as a weighted sum of the Legendre polynomials,
\begin{equation}
   \phi(x) =\sum_{n=0}^{P} w_n P_n(x),
   \label{eq:legendre}
   \end{equation}
where \( w_n \) are the learnable coefficients associated with each polynomial \( P_n(x) \), and \( P \) is the maximum polynomial degree.

\textbf{Taylor:} The Taylor series expansion for a function \( f(x) \) around 0 (Maclaurin series) is given by,
\[
f(x) = \sum_{i=0}^{n} \frac{f^{(i)}(0)}{i!} x^i,
\]
where \( f^{(i)}(0) \) is the \( i \)-th derivative of the function evaluated at 0, \( x \) is the input, and \( i \) is the order of the derivative. In the context of KAN, the coefficients of the Taylor series \( \frac{f^{(i)}(0)}{i!} \) are learned parameters, and the sum is computed up to a specified order. Thus, the basis function in KAN can be expressed as,
\begin{equation}
\phi(x) =  \sum_{n=0}^{P} \left( x^i \cdot w_n \right),
\label{eq:taylor}
\end{equation} where $w_n$ are the learnable Taylor coefficients for the \( i \)-th term, and $P$ is the maximum degree of the Taylor expansion.

\textbf{Bessel:} Bessel polynomials \( y_n(x) \) are defined by a recurrence relation and are typically used in the context of solving differential equations, particularly in problems related to cylindrical symmetry. The recurrence relation for Bessel polynomials \( y_n(x) \) is,
\begin{equation*}
    \begin{split}
        y_0(x) &= 1, \\
        y_1(x) &= x + 1, \\
    \end{split}
\end{equation*}for \( n \geq 2 \), the recurrence relation is,
\[
y_n(x) = (2n - 1) \cdot x \cdot y_{n-1}(x) + y_{n-2}(x).
\] Then, the basis function in KAN is expressed as:
\begin{equation}
\phi(x) = \sum_{n=0}^{P} w_n \cdot y_n(x),
\label{eq:bessel}
\end{equation} where \( w_n \) are the learnable coefficients associated with each Bessel polynomial, and \( P \) is the maximum polynomial degree.

\textbf{Fibonacci:} Fibonacci polynomials \( F_n(x) \) are defined recursively as follows,
\begin{equation*}
    F_0(x) = 0, \quad F_1(x) = 1,
\end{equation*} for \( n \geq 2 \), the Fibonacci polynomials are given by the recurrence relation,
\[
F_n(x) = x \cdot F_{n-1}(x) + F_{n-2}(x).
\] Then, the basis function in KAN can be expressed as,
 \begin{equation}
\phi(x) = \sum_{n=0}^{P} w_n \cdot F_n(x),
\label{eq:fibonacci}
\end{equation}where \( w_n \) are the learnable coefficients associated with each Fibonacci polynomial, and \( P \) is the maximum polynomial degree.

\textbf{Lucas:} Lucas polynomials \( L_n(x) \) are defined recursively as follows,
\[
L_0(x) = 2, \quad L_1(x) = x,
\] 
for \( n \geq 2 \), the Lucas polynomials follow the recurrence relation,
\[
L_n(x) = x \cdot L_{n-1}(x) + L_{n-2}(x).
\]
The basis function in KAN is computed as a weighted sum of Lucas polynomials up to a specified degree \( P \),
\begin{equation}
\phi(x) = \sum_{n=0}^{P} w_n \cdot L_n(x),
\label{eq:lucas}
\end{equation} where \( w_n \) are the learnable coefficients, and \( P \) is the maximum polynomial degree.

\subsubsection*{A.1.2 Fourier Basis Functions}
\textit{Characteristics}: Fourier basis functions exhibit periodicity and are highly effective in representing periodic signals and frequency domain information. They possess significant analytical capability in the frequency domain, making them well-suited for handling both periodic and time-varying signals.

\textit{Applicable Tasks}: Fourier basis functions are particularly useful in tasks such as signal processing, frequency analysis, and time-series data modeling, especially when the data exhibits prominent periodicity or frequency domain characteristics.

\textbf{Fourier: } The Fourier Series is a method for decomposing a periodic function into a sum of sine and cosine functions. Specifically, a periodic function \( f(x) \) can be expressed as,
\[
f(x) = a_0 + \sum_{n=1}^{\infty} \left( a_n \cos(nx) + b_n \sin(nx) \right),
\] where \( a_0 \) is the constant term, \( a_n \) and \( b_n \) are the Fourier coefficients, representing the frequency domain information of the signal, \( \cos(nx) \) and \( \sin(nx) \) are the sine and cosine basis functions, and \( n \) represents the frequency components.

The Fourier coefficients \( a_n \) and \( b_n \) are calculated using the following formulas,
\begin{equation*}
    \begin{split}
        a_n &= \frac{2}{T} \int_0^T f(x) \cos(nx) dx,\\
        b_n &= \frac{2}{T} \int_0^T f(x) \sin(nx) dx,
    \end{split}
\end{equation*} where \( T \) is the period of the function.

The application of the Fourier series to KAN results in the use of Fourier basis functions,
\begin{equation}
    \phi(x) = \sum_{k=1}^{P} \left( c_k \cdot \cos(k \cdot x) + s_k \cdot \sin(k \cdot x) \right),
    \label{eq:fourier}
\end{equation} where \( c_k \) and \( s_k \) are the learnable Fourier coefficients, representing the linear weights for each Fourier component.

\textbf{Sine: } The Fourier series formulation with sine components leads to the following form,
\begin{equation}
\phi(x) = \sum_{k=1}^{P} \left( a_k \cdot \sin\left( \omega_k \cdot x + b_k \right) \right),
\label{eq:sine}
\end{equation} where \( a_k \) are the learnable amplitudes associated with each frequency component, \( \omega_k \) are the frequencies, which are either learned or predefined, and \( b_k \) are the phase shifts, which are computed based on the grid phase and the input phase.

\subsubsection*{A.1.3 Wavelet Basis Functions}
\textit{Characteristics}: Wavelet basis functions exhibit localized time-frequency characteristics, enabling efficient transitions between the time and frequency domains. They excel in multi-scale analysis, making them particularly effective for handling signals with irregularities or abrupt changes.

\textit{Applicable Tasks}: Wavelet basis functions are well-suited for tasks such as signal denoising, image compression, and feature extraction, especially when dealing with non-stationary signals requiring multi-scale analysis.

\textbf{Mexican Hat: } The basis function implemented in MexicanHat KAN is derived from the Mexican Hat wavelet, which is mathematically expressed as,
\[
\psi(x) = \frac{2}{\sqrt{3} \pi^{1/4}} \cdot \left( x^2 - 1 \right) \cdot e^{-\frac{x^2}{2}},
\] where \( x \) is the input variable, and the scaled and translated input \( x_{\text{scaled}} \) is defined as,
\[
x_{\text{scaled}} = \frac{x - \text{translation}}{\text{scale}},
\] where \( \text{translation} \) and \( \text{scale} \) are learnable parameters. The scaled Mexican Hat wavelet basis function becomes,
\[
\psi(x_{\text{scaled}}) = \frac{2}{\sqrt{3} \pi^{1/4}} \cdot \left( x_{\text{scaled}}^2 - 1 \right) \cdot e^{-\frac{x_{\text{scaled}}^2}{2}}.
\]

The final weighted basis function used in the layer is:
\begin{equation}
    \phi(x) = w \cdot \psi(x_{\text{scaled}}),
    \label{eq:mexicanhat}
\end{equation}
where \( w \) are the learnable wavelet weights. The output of the layer aggregates contributions from all these weighted basis functions.

\textbf{Meyer Wavelet:}
Let the input be denoted as \( x \). First, the input is scaled and shifted to obtain the scaled input \( x_{\text{scaled}} \),
\[
x_{\text{scaled}} = \frac{x - \text{translation}}{\text{scale}}
\]
Next, the absolute value of the scaled input is calculated as \( v = |x_{\text{scaled}}| \). Based on the value of \( v \), the auxiliary function \( \text{meyer\_aux}(v) \) is used to compute the Meyer wavelet. The auxiliary function is defined as,
\[
\text{meyer\_aux}(v) = 
\begin{cases} 
1 & \text{if } v \leq \frac{1}{2} \\
0 & \text{if } v \geq 1 \\
\cos\left(\frac{\pi}{2} \cdot \nu(2v - 1)\right) & \text{if } \frac{1}{2} < v < 1
\end{cases},
\] where the function \( \nu(t) \) is a quartic polynomial defined as,
\[
\nu(t) = t^4 \left( 35 - 84t + 70t^2 - 20t^3 \right)
\]
With these steps, the Meyer wavelet function \( \psi(x) \) is computed as,
\[
\psi(x) = \sin(\pi v) \cdot \text{meyer\_aux}(v)
\]

Then, the wavelet function \( \psi(x) \) is weighted by the learned wavelet weights \( w \), resulting in the final basis function,
 \begin{equation}
  \phi(x) = w \cdot \psi(x)
   \label{eq:meyer}
   \end{equation} where \( w \) is the learned wavelet weight, and \( \psi(x) \) is the Meyer wavelet function.

\textbf{Morlet Wavelet:}
The basis function in the Morlet KAN is the Morlet wavelet, which is defined by the product of a Gaussian envelope and a cosine function. Specifically, the wavelet function \( \psi(x) \) is,
\[
\psi(x) = e^{-\frac{x_{\text{scaled}}^2}{2}} \cdot \cos(\omega_0 \cdot x_{\text{scaled}}),
\] where \( x_{\text{scaled}} \) is the scaled input, obtained by applying learned translation and scaling parameters,
\[
x_{\text{scaled}} = \frac{x - \text{translation}}{\text{scale}}.
\]

Then, the wavelet is then weighted by the learned wavelet weights \( w \), resulting in the final basis function:
\begin{equation}
   \phi(x) = w \cdot \psi(x),
   \label{eq:morlet}
   \end{equation}
where \( w \) represents the learned wavelet weight.

\textbf{DoG (Difference of Gaussians):}
The basis function in the DoG KAN is derived from the Derivative of Gaussian (DoG) wavelet. Mathematically, it is expressed as,
\[
\psi(x) = -x_{\text{scaled}} \cdot e^{-\frac{x_{\text{scaled}}^2}{2}},
\] where \( x_{\text{scaled}} \) is the scaled input computed as,
\[
x_{\text{scaled}} = \frac{x - \text{translation}}{\text{scale}}.
\]

The final wavelet basis function is weighted by learned parameters \( w \), resulting in,
 \begin{equation}
\phi(x) = w \cdot \psi(x) = w \cdot \left(-x_{\text{scaled}} \cdot e^{-\frac{x_{\text{scaled}}^2}{2}}\right),
\label{eq:dog}
\end{equation}
where \( w \) represents the learned wavelet weights.

\textbf{Shannon Wavelet:}
The basis function in Shannon KAN is derived from the Shannon wavelet, which is represented as the product of a sinc function and a windowing function (Hamming window) to restrict its support. Mathematically, the Shannon wavelet is expressed as,
\[
\psi(x) = \text{sinc}\left(\frac{x_{\text{scaled}}}{\pi}\right) \cdot w_h(x),
\]
where \( \text{sinc}(x) = \frac{\sin(\pi x)}{\pi x} \) is the sinc function, and \( w_h(x) \) represents the Hamming window function applied to \( x_{\text{scaled}} \). The scaled input \( x_{\text{scaled}} \) is computed as,
\[
x_{\text{scaled}} = \frac{x - \text{translation}}{\text{scale}},
\]
where \( \text{scale} \) and \( \text{translation} \) are learned parameters.

The final basis function weighted by learnable parameters \( w \) is,
\begin{equation}
  \phi(x) = w \cdot \psi(x),
   \label{eq:shannon}
   \end{equation}
where \( w \) are the learned wavelet weights.

\subsubsection{Radial Basis Functions}
\textit{Characteristics:} Radial basis functions (RBFs) exhibit a pronounced localized response, enabling the formation of smooth interpolation or approximation within the input space. Their activation values are computed based on the distance between input vectors, making them particularly effective for non-linear mapping in high-dimensional spaces.  

\textit{Applicable Tasks:} RBFs are well-suited for tasks such as pattern recognition, classification, and function approximation. They demonstrate significant advantages, particularly in handling nonlinear and complex relationships.

\textbf{BSRBF (Basic Spline Radial Basis Function):}
The basis functions in this model combine two components: Radial Basis Functions (RBF) and B-spline basis functions. The RBF is represented by a Gaussian function,
\[\phi_{\text{RBF}}(x) = \exp\left(-\left(\frac{x - c}{\sigma}\right)^2\right),\]
where \(c\) corresponds to the center points, and \(\sigma\) is the scaling factor. The computation of the B-spline basis functions follows a recursive approach analogous to the formulation described in Eq. \eqref{eq:bspline}. Thus, the combined basis function is a sum of the RBF and B-spline components, expressed as,
\begin{equation}
   \phi(x) = w \cdot (\phi_{\text{B-spline}}(x) + \phi_{\text{RBF}}(x)),
   \label{eq:bsrbf}
\end{equation} where \(w\) is the learnable weight.

\textbf{GRBF (Gaussian Radial Basis Function):}
In the GRBF KANL, the basis functions are defined as:  
\[
\phi_{i}(x) = \exp\left(-\left(x - c_{i}\right)^2\right),
\] where \( c_i \) represents the fixed grid points in the interval \([ \text{grid\_min}, \text{grid\_max} ]\) with a total of \(\text{grid\_size}\) points. The RBF basis functions are combined with learnable weights \( w_{o,i,n} \) during the forward pass using the following computation:
 \begin{equation}
   \text{RBF}(x) = \sum_{n} \phi_{n}(x) \cdot w_{o,i,n},
   \label{eq:grbf}
   \end{equation}
where \( \phi_n(x) \) is the RBF basis centered at grid point \( c_n \), and \( w_{o,i,n} \) are the parameters learned during training for each output \( o \), input \( i \), and grid point \( n \).

\textbf{RBF (Radial Basis Function):}
The Radial Basis Function  basis functions are defined as,

\begin{equation}
\phi_i(x) = \exp\left(-\left(\frac{x - c_i}{\Delta}\right)^2\right),
\label{eq:rbf}
\end{equation} where \( c_i \) are the grid centers distributed between the range \([ \text{grid\_min}, \text{grid\_max} ]\), and \( \Delta \) is the scaling factor,
\[
\Delta = \frac{\text{grid\_max} - \text{grid\_min}}{\text{grid\_size} - 1}.
\]

\subsection{Spectral Bias of KAN}
\label{sec:Spectral_Bias_of_KAN}
This section provides a supplementary explanation to the argument presented in Sec. \ref{sec:cn_model}, specifically regarding the claim that ``Bspline-KAN exhibits smaller spectral bias in the low-frequency domain compared to ReLU-MLP". In this section, we will analyze the spectral bias characteristics of KAN and MLP, with a primary focus on Bspline-KAN and ReLU-MLP.

\begin{theorem}
\label{thm:relu_sb}
\textbf{Spectral Bias of ReLU-MLPs \cite{hong2022activation}.} Let \( u: [0, 1] \to \mathbb{R} \) be the one-dimensional function to be approximated, and assume the neural network is of the form,
\[
f_{\text{NN}}(x, \vec{a}) = \sum_{i=1}^n a_i \sigma \left( x - \frac{i}{n} \right),
\]
where \( \sigma \) represents the ReLU activation function, and only the weights \( a_i \) are learned parameters. The fixed parameters include \( \omega_i = 1 \) and \( b_i = \frac{i}{n} \). The loss function is defined as,
\[
L(\vec{a}) = \frac{1}{2} \int_0^1 \left( u(x) - f_{\text{NN}}(x, \vec{a}) \right)^2 \, dx.
\]
According to finite element theory, the structure of the loss function can be expressed as,
\[
L(\vec{a}) = \vec{a}^T M_\sigma \vec{a} - b_{u, \sigma}^T \vec{a},
\]
where the components of the vector \( b_{u, \sigma} \) are given by,
\[
\left(b_{u, \sigma}\right)_i = \int_0^1 u(x) \sigma \left( x - \frac{i}{n} \right) \, dx,
\]
and the elements of the mass matrix \( M_\sigma \) are,
\[
(M_\sigma)_{ij} = \int_0^1 \sigma \left( x - \frac{i}{n} \right) \sigma \left( x - \frac{j}{n} \right) \, dx.
\]
Furthermore, the matrix \( M_\sigma \) is positive definite.
Let \( \lambda_1 \leq \lambda_2 \leq \dots \leq \lambda_n \) denote the eigenvalues of \( M_\sigma \). These eigenvalues satisfy the following asymptotic relation,
\[
\frac{\lambda_n}{\lambda_j} \sim \frac{n^4}{j^4}, \quad j = 1, 2, \dots, n.
\]
This theorem reveals the spectral bias associated with the ReLU activation function, indicating that as the network size \( n \) increases, the ratio of the largest eigenvalue to smaller eigenvalues grows rapidly.
\end{theorem}

Following the derivation outlined in Theorem \ref{thm:relu_sb} \cite{hong2022activation}, we analyze the spectral properties of the single-layer KAN. Specifically, we consider the LiuKAN \cite{liu2024kan} implemented using B-splines, while neglecting the SiLU nonlinearity (Eq. \ref{eq:bspline}). Therefore, when the input is \( x \in \mathbb{R}^d \), a single-layer KAN network can be represented as,
\begin{equation}
    \Phi(\mathbf{x}, \vec{c})_i = \sum_{j=1}^d \sum_{l=1}^{G+k-1} c_{i j l} B_l\left(\mathbf{x}_j\right),
\end{equation}
where \( c_{ijl} \) denotes the unique learnable parameters, \( k \) represents the degree of the B-splines, and \( G \) is the grid size. The indices \( j = 1, \dots, d \) correspond to the input dimension, while \( i = 1, \dots, d' \) denotes the output dimension (\( d' \) represents the output dimensionality).

Let \( u: [-1,1]^d \to \mathbb{R}^{d^2} \) represent the target function we aim to learn. The least squares regression loss is then given by,
\begin{equation}
    L(\vec{c}) = \int_{-1}^{1}(u(\mathbf{x})-\Phi(\mathbf{x}, \vec{c}))^2 d \mathbf{x}.
\end{equation}
Similar to Theorem \ref{thm:relu_sb}, the loss function can be expressed as a quadratic function of the parameters \( \vec{c} \), i.e.,
\begin{equation}
   L(\vec{c}) = \frac{1}{2}\vec{c}^T M \vec{c} + b^T \vec{c},
\end{equation} where the Hessian matrix \( M \) is given by,
\begin{equation}
    M_{(i, j, l),\left(i^{\prime}, j^{\prime}, l^{\prime}\right)}= \begin{cases}\int_{-1}^{1} B_l\left(\mathbf{x}_j\right) B_{l^{\prime}}\left(\mathbf{x}_{j^{\prime}}\right) d \mathbf{x} & i=i^{\prime} \\ 0 & i \neq i^{\prime} .\end{cases}
\end{equation}

It is important to observe that the matrix \( M \) is a block diagonal matrix with \( d^{\prime} \) identical blocks. These blocks, each of size \( (G+k-1) d \times (G+k-1) d \), are denoted as \( H \). Therefore, the ratio of the eigenvalues (\( 0 \leq \lambda_1(H) \leq \cdots \leq \lambda_{(G+k-1) d}(H) \)) of \( H \) has the following upper bound \cite{wang2024expressiveness},
\begin{equation}
    \frac{\lambda_{(G+k-1) d}(H)}{\lambda_d(H)} \leq dK,
\end{equation} where $K$ is a constant that depends solely on $k$.

Since \( M \) is a block diagonal matrix, its eigenvalues are simply the collection of the eigenvalues of the individual blocks \( H \). Therefore, the largest eigenvalue of \( M \) is bounded by the largest eigenvalue of \( H \), and similarly, the smallest eigenvalue of \( M \) corresponds to the smallest eigenvalue of \( H \). Given that the eigenvalue ratio for each block \( H \) is bounded by \( dK \), this upper bound can be applied to the entire matrix \( M \). Consequently, the ratio of the largest to the smallest eigenvalue of \( M \) is also constrained by the same factor \( dK \).
Therefore, for the eigenvalues of the matrix \( M \), denoted as \( 0 \leq \lambda_1(M) \leq \cdots \leq \lambda_{(G+k-1) d d^{\prime}}(M) \), we have the following relationship (\textbf{\textit{Spectral Bias of Bspline-KANs}}),
\begin{equation}
    \label{eq:kan_sb}
    \frac{\lambda_{(G+k-1) d d^{\prime}}(M)}{\lambda_{d^{\prime}(d-1)+1}(M)} \leq dK.
\end{equation} where $K$ is a constant that depends solely on $k$.

The eigenvalues of the matrix \( M \) associated with the Bspline-KAN network are subject to the constraint specified in Eq. \eqref{eq:kan_sb}, which bounds the ratio of the largest eigenvalue to the \( (d^{\prime}(d-1)+1) \)-th smallest eigenvalue by \( dK \), where \( K \) is a constant dependent solely on \( k \). This property implies that, apart from the first \( d^{\prime}(d-1) \) smaller eigenvalues, the remaining eigenvalues exhibit a relatively uniform distribution. Consequently, the condition number of the matrix \( M \) (defined as the ratio of the largest to the smallest non-zero eigenvalue) is significantly constrained. Such uniformity in eigenvalue distribution indicates that Bspline-KAN networks facilitate a balanced learning rate across all eigenvector directions during gradient descent optimization, thereby mitigating excessive bias toward specific frequency components.

In contrast, the Hessian matrix \( M_\sigma \) of ReLU-MLP networks exhibits a highly non-uniform eigenvalue distribution. As established in Theorem \ref{thm:relu_sb}, the eigenvalue ratios follow the relation \( \lambda_n / \lambda_j \sim n^4 / j^4 \), where the growth rate of eigenvalues is inversely proportional to the fourth power of their indices. This disparity causes the condition number of the Hessian matrix to increase rapidly, scaling as \( n^4 \) with the network width \( n \). The substantial growth in condition number directly impacts the behavior of gradient descent optimization, whereby low-frequency components (associated with smaller eigenvalues) are prioritized due to their significantly slower convergence rates, while high-frequency components (associated with larger eigenvalues) converge faster but are learned later in the training process. This phenomenon, termed spectral bias, is a defining characteristic of ReLU-MLP networks during optimization.

\textbf{Task Adaptability.} The above theoretical analysis and comparative characteristics highlight the distinct advantages of Bspline-KAN and ReLU-MLP across various task scenarios. Bspline-KAN, with its uniform eigenvalue distribution and constrained condition number, facilitates consistent feature learning during gradient descent, exhibiting weak spectral bias. This makes it particularly effective in image denoising tasks, especially in scenarios requiring the preservation of low-frequency information and suppression of high-frequency noise. However, its ability to capture high-frequency details is relatively limited, rendering it less effective for modeling complex edges and textures compared to ReLU-MLP. In contrast, ReLU-MLP excels in image fitting tasks, leveraging its powerful expressive capacity and ability to capture high-frequency features, particularly in tasks that require precise reconstruction of high-resolution details. Nevertheless, the strong spectral bias and rapidly growing condition number in ReLU-MLP may lead to reduced optimization efficiency during training, particularly in tasks involving low-frequency global structure modeling, where Bspline-KAN performs more effectively. Thus, Bspline-KAN is more suitable for tasks emphasizing smoothness and global consistency, while ReLU-MLP is advantageous for tasks requiring high-frequency detail and complex texture modeling.

\subsection{KAN is a Heuristic Deep Gaussian Process}
\label{sec:KAN_is_a_Heuristic_Deep_Gaussian_Process}

This section provides a supplementary explanation to the argument presented in Sec. \ref{sec:cn_model}, which posits that ``KAN can be viewed as a heuristic deep Gaussian process". We will first introduce the deep Gaussian process network, followed by a comparison with the KAN network.

\textbf{Deep Gaussian Processes.} Deep Gaussian Processes (DGP) \cite{damianou2013deep} are a class of models that incorporate deep learning principles into Gaussian processes \cite{Rasmussen2004}, enabling the modeling of more complex covariance structures through multiple layers of Gaussian processes. The core idea of DGP is to apply the methodology of deep neural networks to the kernel function modeling in traditional Gaussian processes, thereby capturing the nonlinear relationships between inputs.

The DGP network consists of multiple layers of transformations, similar to Coordinate-KANs. However, the primary distinction between DGP and KAN lies in the fact that each transformation in DGP is represented by a Gaussian process, rather than a simple activation function. As a result, each layer of the DGP network can be viewed as a mapping from one layer to the next through a Gaussian process.

Assuming a DGP network with \(L\) layers, where the input to the \(l\)-th layer is denoted as \(\mathbf{x}_l\) and the output as \(\mathbf{x}_{l+1}\), the network structure can be represented as follows,
\begin{equation}
\label{eq:dgp}
\begin{split}
        \Phi(\mathbf{x})_{\text{DGP}} = & \left(\mathcal{G}_{n-1} \circ \mathcal{G}_{n-2} \circ \ldots \circ \mathcal{G}_0\right)(\mathbf{x}), \\
    &\mathbf{x}_l \mapsto \mathcal{G}_l(\mathbf{x}_l),
\end{split}
\end{equation}
where \(\mathcal{G}_l\) denotes the Gaussian process transformation at the \(l\)-th layer, defining the mapping from input \(\mathbf{x}_l\) to output \(\mathbf{x}_{l+1}\). The input and output of each layer are modeled through the covariance function of the Gaussian process, thereby capturing the complex nonlinear relationships.

In DGP, the transformation $\mathcal{G}_l$ at the $l$-th layer is determined by the covariance function of the Gaussian process. Assuming that the input at the \(l\)-th layer is \(\mathbf{x}_l\) and the output is \(\mathbf{x}_{l+1}\), we use a covariance function \(k_l(\mathbf{x}_l, \mathbf{x}_l')\) to measure the correlation between the input points \(\mathbf{x}_l\) and \(\mathbf{x}_l'\). The covariance matrix \(\mathbf{K}_l\) for the Gaussian process at the \(l\)-th layer is formed from these covariance functions, capturing the similarity structure among data points in the input space. Specifically, the elements of the covariance matrix \(\mathbf{K}_l\) for the \(l\)-th layer can be written as,
\begin{equation}
\label{eq:dgp_kernel}
\mathbf{K}_l = \begin{bmatrix}
k_l(\mathbf{x}_1, \mathbf{x}_1') & k_l(\mathbf{x}_1, \mathbf{x}_2') & \cdots & k_l(\mathbf{x}_1, \mathbf{x}_N') \\
k_l(\mathbf{x}_2, \mathbf{x}_1') & k_l(\mathbf{x}_2, \mathbf{x}_2') & \cdots & k_l(\mathbf{x}_2, \mathbf{x}_N') \\
\vdots & \vdots & \ddots & \vdots \\
k_l(\mathbf{x}_N, \mathbf{x}_1') & k_l(\mathbf{x}_N, \mathbf{x}_2') & \cdots & k_l(\mathbf{x}_N, \mathbf{x}_N'),
\end{bmatrix}
\end{equation}
where \(k_l(\mathbf{x}_i, \mathbf{x}_j)\) represents the covariance between the \(i\)-th and \(j\)-th input points \(\mathbf{x}_i\) and \(\mathbf{x}_j\) at the \(l\)-th layer. This covariance matrix is used to define the transformation from the input \(\mathbf{x}_l\) to the output \(\mathbf{x}_{l+1}\).

At each layer of the DGP, the relationship between the input \(\mathbf{x}_l\) and the output \(\mathbf{x}_{l+1}\) is defined through the covariance matrix. The covariance function dictates the correlation between input points, which in turn influences the output. Specifically, the transformation at the \(l\)-th layer can be expressed as,
\[
\mathbf{x}_{l+1} = \mathcal{G}(\mathbf{x}_l)= \mathbf{K}_l \mathbf{x}_l,
\]
where \(\mathbf{K}_l\) is the covariance matrix at the \(l\)-th layer, calculated using the kernel function \(k_l(\mathbf{x}_l, \mathbf{x}_l')\).

\textbf{KAN is a Heuristic Deep Gaussian Process.} As outlined in Sec. \ref{sec:cn_model}, for a KAN network with \(L\) layers, its structure is analogous to that of a DGP and can be expressed as a product of transformation matrices, namely,
\begin{equation}
\label{eq:kan_sm}
\begin{split}
        \Phi(\mathbf{x})_{\text{KAN}}=& \left(\boldsymbol{\Phi}_{n-1} \circ \boldsymbol{\Phi}_{n-2} \circ \ldots \circ \boldsymbol{\Phi}_0\right)(\mathbf{x}), \\
    &\mathbf{x}_l \mapsto \boldsymbol{\Phi}_l\left(\mathbf{x}_l\right),
\end{split}
\end{equation} where $\boldsymbol{\Phi}_l$ denotes the $l$-th KAN layer of the network, which essentially represents the transformation matrix between the layer's input $\mathbf{x}_l$ and output $\mathbf{x}_{l+1}$,
\begin{equation}
    \mathbf{x}_{l+1} = \boldsymbol{\Phi}_l\mathbf{x}_l.
\end{equation}

In contrast to DGP, where the transformation matrices are composed of covariance functions (Eq. \ref{eq:dgp_kernel}), the transformation matrices in KAN are formed by fully learnable basis functions, namely,
\begin{equation}
\label{eq:kan_kernel}
   \boldsymbol{\Phi}_l = \begin{bmatrix}
\phi_{l, 1, 1}(\cdot) & \phi_{l, 1, 2}(\cdot) & \cdots & \phi_{l, 1, n_l}(\cdot) \\
\phi_{l, 2, 1}(\cdot) & \phi_{l, 2, 2}(\cdot) & \cdots & \phi_{l, 2, n_l}(\cdot) \\
\vdots & \vdots & \ddots & \vdots \\
\phi_{l, n_{l+1}, 1}(\cdot) & \phi_{l, n_{l+1}, 2}(\cdot) & \cdots & \phi_{l, n_{l+1}, n_l}(\cdot)
\end{bmatrix}
\end{equation}
where \(\phi_{l, j, i}\) represents the basis function connecting the \(i\)-th neuron in the \(l\)-th layer to the \(j\)-th neuron in the \((l+1)\)-th layer.

Based on the above analysis, we can draw the following conclusions:
\begin{itemize}
    \item \textit{Transformation Mechanism:} In KAN, the transformation at each layer is achieved through activation functions, whereas in DGP, it is accomplished through covariance functions. Both activation functions and covariance functions effectively capture the similarity and nonlinear relationships between data points. Therefore, the activation functions in KAN can be viewed as analogous to the covariance functions in DGP.

    \item \textit{Nonlinear Modeling:} Both KAN and DGP are capable of modeling complex nonlinear relationships between input data. KAN achieves this through nonlinear transformations via activation functions, while DGP captures nonlinear similarities between data points using covariance functions.

    \item \textit{Hierarchical Structure:} Both KAN and DGP utilize a multi-layer structure to progressively abstract the complex relationships within the input data. The output of each layer depends on the input from the previous layer, allowing them to incrementally capture and model more complex input-output relationships.
\end{itemize}

Therefore, although KAN employs activation functions at each layer and DGP utilizes covariance functions, both serve similar purposes: capturing the similarity between data points and modeling nonlinear relationships. In this regard, KAN can be regarded as a heuristic deep Gaussian process, with the transformations at each layer functioning similarly to the role of covariance functions in DGP. Consequently, KAN shares several characteristics with DGPs: while it demonstrates certain advantages in modeling nonlinearity and uncertainty, it also inevitably encounters challenges such as high computational complexity and difficulties in training \cite{jakkala2021deep}.

\section{Coordinate-MLPs}
\label{sec:appendix_Coordinate-MLPs}

\subsection{Details of Positional Encoding}
\label{sec:Details_of_Positional_Encoding}

This section serves as a supplementary explanation to Sec. \ref{sec:pe}, providing a detailed discussion on the specific forms of positional encoding and their influence on the kernel functions within the NTK.

\textbf{Positional Encoding Formula.} As demonstrated in Theorem \ref{thm:relu_sb}, networks such as those utilizing ReLU activations exhibit spectral bias. These networks tend to first learn low-frequency functions, as these functions exhibit global behavior and are more stable for optimization. However, spectral bias also hinders the network's ability to learn high-frequency information. To address this issue, NeRF \cite{mildenhall2020nerf} proposes mapping input coordinates into a higher-dimensional space to enable the learning of high-frequency information within radiance fields. Specifically, given an input coordinate \(\mathbf{x} \in \mathbb{R}\), NeRF maps it into \(\mathbb{R}^{2L}\) space as follows,
\begin{equation}
\label{eq:pe_nerf}
\begin{aligned}
    &\gamma_{\text{NeRF}}({\mathbf{x}}) = \\ &\left[\sin (2^0 \pi {\mathbf{x}}), \cos (2^0 \pi {\mathbf{x}}), \ldots, \sin (2^{L-1} \pi {\mathbf{x}}), \cos (2^{L-1} \pi {\mathbf{x}})\right]
\end{aligned}
\end{equation} where \(L\) represents the dimension of the high-frequency space. We refer to this positional encoding scheme as \texttt{NeRF}. 

To modulate the learnable frequency range of the network, FFN \cite{tancik2020ffn} introduces Gaussian random noise into the  \texttt{NeRF} positional encoding, leveraging insights from NTK \cite{jacot2018neuralntk} theory, as follows,
\begin{equation}
\label{eq:pe_rff}
\begin{split}
     \gamma_{\text{RFF}}({\mathbf{x}}) &=  
            [ a_0 \cos(2 \pi \mathbf{b}_0^{\mathrm{T}} \mathbf{x}),
            a_0 \sin(2 \pi \mathbf{b}_0^{\mathrm{T}} \mathbf{x}), \ldots,\\
            & a_{L-1}  \cos(2 \pi \mathbf{b}_{L-1}^{\mathrm{T}} \mathbf{x}), 
            a_{L-1} \sin(2 \pi \mathbf{b}_{L-1}^{\mathrm{T}} \mathbf{x})],\\
            & \quad \mathbf{b}_i \sim \mathcal{N}(0, \sigma^2).
\end{split}
\end{equation}
where frequency vectors $\mathbf{b}_i$ are randomly sampled from a Gaussian distribution with a standard deviation of $\sigma$ (hyper-parameter). In the \texttt{RFF} positional encoding, the hyperparameter \(\sigma\) requires extensive parameter tuning for different tasks, posing a challenge in practical applications. 

To address this issue, we propose in this study that a KAN network employing Fourier kernels as basis functions can serve as a fully learnable positional encoding mechanism. Specifically, it can be expressed as,
\begin{equation}
    \label{eq:pe_fkan_sm}
    \small
    \begin{split}
        \gamma_{\text{FKAN}}(\mathbf{x}) = & \left[ a_{11} \cos(\omega_1 \mathbf{x}_1), b_{11} \sin(\omega_1 \mathbf{x}_1), \dots, \right. \\
         & \left. a_{D \Omega} \cos(\omega_\Omega \mathbf{x}_D), b_{D \Omega} \sin(\omega_\Omega \mathbf{x}_D) \right],
    \end{split}
\end{equation} 
where \(D\) represents the dimensionality of \(\mathbf{x}\), \(\mathbf{x}_i\) denotes the \(i\)-th component of \(\mathbf{x}\), and \(\Omega\) is a parameter defining the maximum frequency threshold, similar to \(L\) in \texttt{NeRF} and \texttt{RFF}.

\textbf{Position Encoding Kernel.}
\texttt{NeRF}, \texttt{RFF} and \texttt{FKAN} adopt Fourier feature mapping to project input coordinates \(\mathbf{x}\) from Euclidean space onto a hypersphere, leveraging the identity \(\sin^2 x + \cos^2 x = 1\). This transformation imparts translation invariance to coordinates that are originally only rotationally invariant, effectively decoupling the network from absolute positional dependencies. This characteristic is particularly crucial for implicit representation tasks, where the goal is to model objects or scenes consistently, regardless of their spatial location \cite{tancik2020ffn}.

The translation invariance can be evaluated through the kernel of the position encoding. Specifically, for a position encoding \(\gamma\), its kernel can be expressed as,
\begin{equation}
    k_{\gamma}(\mathbf{x}_1, \mathbf{x}_2) = \gamma(\mathbf{x}_1)^T\gamma(\mathbf{x}_2),
\end{equation} 
where \(\mathbf{x}_1\) and \(\mathbf{x}_2\) represent the input coordinates. Therefore, the kernels of \texttt{NeRF}, \texttt{RFF}, and \texttt{FKAN} are given by,
\begin{equation}
     \begin{split}
    k_{\text{NeRF}}(\mathbf{x}_1, \mathbf{x}_2) &= \\
          &\sum_{i=0}^{L-1} (\cos(2^i \pi \mathbf{x}_1) \cos(2^i \pi \mathbf{x}_2) \\
          & + \sin(2^i \pi \mathbf{x}_1) \sin(2^i \pi \mathbf{x}_2)), \\
    k_{\text{RFF}}(\mathbf{x}_1, \mathbf{x}_2) &= \\
        &\sum_{i=0}^{L-1} a_i^2 (\cos(2\pi \mathbf{b}_i^T \mathbf{x}_1) \cos(2\pi \mathbf{b}_i^T \mathbf{x}_2) \\ 
        &+ \sin(2\pi \mathbf{b}_i^T \mathbf{x}_1) \sin(2\pi \mathbf{b}_i^T \mathbf{x}_2)), \\
    k_{\text{FKAN}}(\mathbf{x}_1, \mathbf{x}_2) &= \\
        &\sum_{i=1}^{D} \sum_{\omega=1}^{\Omega} a_{i\omega}^2 \cos(\omega (\mathbf{x}_{1,i} - \mathbf{x}_{2,i}))\\
        & + b_{i\omega}^2 \cos(\omega (\mathbf{x}_{1,i} - \mathbf{x}_{2,i})), \\
    \end{split}
\end{equation}
By applying the trigonometric identity \(\cos \theta_1 \cos \theta_2 + \sin \theta_1 \sin \theta_2 = \cos(\theta_1 - \theta_2)\), we can further simplify the expression as follows,
\begin{equation}
    \label{eq:pe_kernels}
    \begin{split}
        k_{\text{NeRF}}(\mathbf{x}_1, \mathbf{x}_2) &= \sum_{i=0}^{L-1} \cos\left( 2^i \pi (\mathbf{x}_1 - \mathbf{x}_2) \right), \\
        k_{\text{RFF}}(\mathbf{x}_1, \mathbf{x}_2) &= \sum_{i=0}^{L-1} a_i^2 \cos\left( 2 \pi \mathbf{b}_i^T (\mathbf{x}_1 - \mathbf{x}_2) \right), \\
        k_{\text{FKAN}}(\mathbf{x}_1, \mathbf{x}_2) &= \sum_{i=1}^{D} \sum_{\omega=1}^{\Omega} \left( a_{i\omega}^2 + b_{i\omega}^2 \right) \cos\left( \omega (\mathbf{x}_{1,i} - \mathbf{x}_{2,i}) \right).
    \end{split}
\end{equation}
Clearly, these kernel functions exhibit translation invariance, enabling the network to remain invariant to changes in spatial positions. However, they differ in how they regulate frequency. \texttt{NeRF} sets fixed frequencies using \( 2^i \), while \texttt{RFF} adjusts frequencies through random Gaussian vectors \( \mathbf{b}_i \). In contrast, \texttt{FKAN} only requires the specification of a maximum frequency threshold \( \Omega \) to control the range of frequency parameters \( \omega \). Furthermore, \texttt{FKAN}'s Fourier coefficients are fully learnable, and they respond differently to each component of the input coordinates.

\textbf{Impact of Positional Encoding on NTK.} 
According to the theory related to the Neural Tangent Kernel (NTK) \cite{jacot2018neuralntk, arora2019fine, basri2020frequency}, as the width of the layers in \( \Phi \) tends to infinity and the learning rate for SGD approaches zero, the function \( \Phi(\mathbf{x}; \theta) \) converges to the solution of kernel regression using the NTK during the training process. The NTK is defined as follows,
\begin{equation}
    k_{\mathrm{NTK}}\left(\mathbf{x}_i, \mathbf{x}_j\right) = \mathbb{E}_{\theta \sim \mathcal{N}}\left\langle \frac{\partial \Phi \left(\mathbf{x}_i; \theta\right)}{\partial \theta}, \frac{\partial \Phi \left(\mathbf{x}_j; \theta\right)}{\partial \theta} \right\rangle .
\end{equation}
Additionally, when the inputs are restricted to a hypersphere, the NTK of the MLP can be expressed as a dot product kernel, taking the form \( h_{\mathrm{NTK}}\left(\mathbf{x}_i^{\mathrm{T}} \mathbf{x}_j\right) \), where \( h_{\mathrm{NTK}}: \mathbb{R} \rightarrow \mathbb{R} \) is a scalar function.

After the input coordinates undergo positional encoding, the mapped features are constrained to lie on the hypersphere. When these features are subsequently passed through the MLP network \( \Phi(\gamma(\mathbf{x}), \theta) \), the resulting combined NTK is expressed as,
\begin{equation}
\begin{split}
     h_{\mathrm{NTK}}\left(\gamma\left(\mathbf{x}_i\right)^{\mathrm{T}} \gamma\left(\mathbf{x}_j\right)\right) &= h_{\mathrm{NTK}}\left(h_\gamma\left(\mathbf{x}_i-\mathbf{x}_j\right)\right) \\
     & = h_{\mathrm{NTK}}\left(k_{\gamma}(\mathbf{x}_i, \mathbf{x}_j)\right),
\end{split}
\end{equation}
where \( k_{\gamma} \) is defined in Eq. \eqref{eq:pe_kernels}.
Therefore, training the network at these embedded input points is equivalent to performing kernel regression using the stationary combination of the NTK function \( h_{\mathrm{NTK}} \circ h_\gamma \). The MLP function approximates the convolution of the combined NTK with a weighted Dirac delta function at each training input point \(\mathbf{x}_i\) \cite{tancik2020ffn},
\begin{equation}
    \hat{\Phi} = \left(h_{\mathrm{NTK}} \circ h_\gamma\right) * \sum_{i=1}^n w_i \delta_{\mathbf{x}_i}
\end{equation}
where \(\mathbf{w} = \mathbf{K}^{-1} \mathbf{y}\) (\(\mathbf{K}\) is the kernel regression matrix and \(\mathbf{y}\) represents the true label vector). This formulation enables an analogy to signal processing, where the combined NTK functions similarly to a reconstruction filter.

\textbf{Task adaptability.} As shown in Table \ref{tab:benchmark_leaderboard}, for signals with a relatively uniform frequency distribution (such as images and audio), both \texttt{RFF} and \texttt{FAKN} can increase the bandwidth of the NTP spectrum through the Gaussian scale parameter \(\sigma\) and the maximum frequency threshold parameter \(\Omega\), effectively capturing different frequency components. In comparison to \texttt{RFF}, our proposed \texttt{FKAN} is a fully learnable positional encoding, which can adaptively learn different frequencies, thus avoiding the cumbersome process of hyperparameter tuning. However, for signals with complex frequency distributions (such as neural radiance fields), it is challenging to identify an appropriate Gaussian scale parameter \(\sigma\) for \texttt{RFF} to represent different frequency components. Similarly, for \texttt{FKAN}, it is difficult to fully capture diverse frequency components with a limited set of learnable parameters. Therefore, in such cases, the simplicity and fixed frequency characteristics of \texttt{NeRF} are more advantageous for model training and can provide a more stable representation of the signal.

\subsection{Details of Activation Functions}
\label{sec:Details_of_Activation_Functions}
This section provides supplementary explanations to Sec. \ref{sec:nonlinear}. We will analyze the properties of different activation functions from the perspectives of gradient, curvature, Fourier transform, L-Lipschitz continuity, and L-smoothness. For convenience, the activation function is denoted as \( f(x) \).

1. \textbf{ReLU}:
\[
f(x) = \max(0, x)
\]

\textit{Gradient:}
\[
f'(x) =
\begin{cases}
1, & \text{if } x > 0 \\
0, & \text{if } x < 0 \\
\text{undefined}, & \text{if } x = 0
\end{cases}
\]

\textit{Curvature:}
\[
f''(x) = 0, \quad \text{for } x \neq 0
\]

\textit{Fourier Transform:}
\[
\mathcal{F}\{f(x)\}(k)  = -\frac{1}{k^2}+ i \pi \delta'(k)
\]

\textit{L-Lipschitz:}
\[
L = \max|f'(x)| = 1
\]

\textit{L-Smoothness:} The gradient of ReLU is not Lipschitz continuous because it jumps discontinuously at \( x = 0 \). Thus, \( f(x) \) is not \( L \)-smooth.

2. \textbf{PReLU}:
\[
f(x) =
\begin{cases}
x, & \text{if } x \geq 0 \\
ax, & \text{if } x < 0
\end{cases}
\]

\textit{Gradient:}
\[
f'(x) =
\begin{cases}
1, & \text{if } x > 0 \\
a, & \text{if } x < 0 \\
\text{undefined}, & \text{if } x = 0
\end{cases}
\]

\textit{Curvature:}
\[
f''(x) = 0, \quad \forall x \neq 0
\]

\textit{Fourier Transform:}
\[
\mathcal{F}\{f(x)\}(k) = \frac{a-1}{k^2}
\]

\textit{L-Lipschitz:}
\[
L = \max(1, |a|)
\]

\textit{L-Smoothness:} PReLU is not \( L \)-smooth due to the gradient discontinuity at \( x = 0 \).

3. \textbf{Sine}:
\[
f(x) = \sin(\omega x)
\]

\textit{Gradient:}
\[
f'(x) = \omega \cos(\omega x)
\]

\textit{Curvature:}
\[
f''(x) = -\omega^2 \sin(\omega x)
\]

\textit{Fourier Transform:}
\[
\mathcal{F}\{f(x)\}(k) = -i \pi (\delta(\omega - k) - \delta(k + \omega))
\]

\textit{L-Lipschitz:}
\[
L = \max|f'(x)| = |\omega|
\]

\textit{L-Smoothness:}
\[
L =  \max|f''(x)| = |\omega^2|
\]

4. \textbf{ScaledSine}:
\[
f(x) = a\sin(\omega bx + c) + d
\]

\textit{Gradient:}
\[
f'(x) = a \omega b \cos(\omega bx + c)
\]

\textit{Curvature:}
\[
f''(x) = -a \omega^2 b^2 \sin(\omega bx + c)
\]

\textit{Fourier Transform:}
\begin{equation*}
\begin{split}
    \mathcal{F}\{f(x)\}(k) &= \pi(2 d \delta(k) \\
    &-i a e^{-i c}\left(-\delta(k+b \omega)+e^{2 i c} \delta(b \omega-k)\right))
\end{split}
\end{equation*}

\textit{L-Lipschitz:}
\[
L = \max|f'(x)| = |a \omega b|
\]

\textit{L-Smoothness:}
\[
L = \max|f''(x)| = |a \omega^2 b^2|
\]

5. \textbf{Gaussian}:
\[
f(x) = e^{-\frac{x^2}{2\sigma^2}}
\]

\textit{Gradient:}
\[
f'(x) = -\frac{x}{\sigma^2} e^{-\frac{x^2}{2\sigma^2}}
\]

\textit{Curvature:}
\[
f''(x) = \left(\frac{x^2}{\sigma^4} - \frac{1}{\sigma^2}\right) e^{-\frac{x^2}{2\sigma^2}}
\]

\textit{Fourier Transform:}
\[
\mathcal{F}\{f(x)\}(k) = \sqrt{2\pi \sigma^2} \cdot e^{-\frac{1}{2} \sigma^2 k^2}
\]

\textit{L-Lipschitz:}
\[
L = \max |f'(x)| = \frac{1}{\sigma}e^{-\frac{1}{2}}
\]

\textit{L-Smoothness:}
\[
L = \max |f''(x)| = \frac{2}{\sigma^2} e^{-\frac{3}{2}}
\]

6. \textbf{Laplacian}:
\[
f(x) = e^{-\frac{|x|}{a}}
\]

\textit{Gradient:}
\[
f'(x) = -\frac{x e^{-\frac{|x|}{a}}}{a |x|}
\]

\textit{Curvature:}
\[
f''(x) = \frac{e^{-\frac{|x|}{a}}}{a^2}
\]

\textit{Fourier Transform:}
\[
\mathcal{F}\{f(x)\}(k) =  \frac{2a}{a^2 k^2 + 1}, \quad \text{for } \operatorname{Re}\left(\frac{1}{a}\right) > 0
\]

7. \textbf{SuperGaussian}:
\[
f(x) = \left(e^{-\frac{x^2}{2\sigma^2}}\right)^b
\]

\textit{Gradient:}
\[
f'(x) = -\frac{b x \left( e^{-\frac{x^2}{2 \sigma^2}} \right)^b}{\sigma^2}
\]

\textit{Curvature:}
\[
f''(x) = \left( e^{-\frac{x^2}{2 \sigma^2}} \right)^b \left( \frac{b^2 x^2}{\sigma^4} - \frac{b}{\sigma^2} \right)
\]

\textit{Fourier Transform:} The calculation is too complicated.

8. \textbf{Gabor}:
\[
f(x) = e^{-ax^2} \cos(\omega x)
\]

\textit{Gradient:}
\[
f'(x) = -e^{-a x^2} \left( 2a x \cos(\omega x) + \omega \sin(\omega x) \right)
\]

\textit{Curvature:}
\begin{equation*}
    \begin{split}
        f''(x) &= e^{-a x^2}( 4 a^2 x^2 \cos(\omega x) \\
        & + 4 a x \omega \sin(\omega x) - 2 a \cos(\omega x) - \omega^2 \cos(\omega x))
    \end{split}
\end{equation*}

\textit{Fourier Transform:}
\begin{equation*}
    \begin{split}
         \mathcal{F}\{f(x)\}(k) &= \frac{\sqrt{\pi} \, e^{-\frac{(k + \omega)^2}{4a}} \left(e^{\frac{k \omega}{a}} + 1\right)}{2\sqrt{a}},\\
    & \quad \text{for } \omega \in \mathbb{R} \, \wedge \, \operatorname{Re}(a) > 0
    \end{split}
\end{equation*}

9. \textbf{Sinc}:
\[
f(x) =\frac{\sin(\pi x)}{\pi x}
\]

\textit{Gradient:}
\[
f'(x) = \frac{\pi x \cos(\pi x) - \sin(\pi x)}{\pi x^2}
\]

\textit{Curvature:}
\[
f''(x) = \frac{\left( 2 - \pi^2 x^2 \right) \sin(\pi x) - 2 \pi x \cos(\pi x)}{\pi x^3}
\]

\textit{Fourier Transform:}
\[
\mathcal{F}\{f(x)\}(k) =  \frac{1}{2} \left( \operatorname{sgn}(\pi - k) + \operatorname{sgn}(k + \pi) \right)
\]

10. \textbf{ExpSin}:
\[
f(x) = e^{\sin(ax)}
\]

\textit{Gradient:}
\[
f'(x) = a e^{\sin(a x)} \cos(a x)
\]

\textit{Curvature:}
\[
f''(x) = -a^2 e^{\sin(a x)} \left( \sin(a x) - \cos^2(a x) \right)
\]

\textit{Fourier Transform:} The calculation is too complicated.

11. \textbf{Sigmoid}:
\[
f(x) = \frac{1}{1 + e^{-x}}
\]

\textit{Gradient:}
\[
f'(x) = \frac{e^{-x}}{\left( e^{-x} + 1 \right)^2}
\]

\textit{Curvature:}
\[
f''(x) = -\frac{e^x \left( e^x - 1 \right)}{\left( e^x + 1 \right)^3}
\]

\textit{L-Lipschitz:}
\[
L = \max |f'(x)| = \frac{1}{4}
\]

\textit{L-Smoothness:}
\[
L = \max |f''(x)| = \frac{1}{6\sqrt{3}}
\]

\textit{Fourier Transform:}
\[
\mathcal{F}\{f(x)\}(k) = \pi \left( \delta(k) - i \, \mathrm{csch}(\pi k) \right)
\]

12. \textbf{Tanh}:
\[
f(x) = \frac{e^{x} - e^{-x}}{e^{x} + e^{-x}}
\]

\textit{Gradient:}
\[
f'(x) = 1-f^2(x) =\frac{4 e^{2x}}{\left( e^{2x} + 1 \right)^2}
\]

\textit{Curvature:}
\[
f''(x) = -2f(x)(1-f^2(x))=-\frac{8 e^{2x} \left( e^{2x} - 1 \right)}{\left( e^{2x} + 1 \right)^3}
\]

\textit{Fourier Transform:}
\[
\mathcal{F}\{f(x)\}(k) = -i \pi \, \mathrm{csch} \left( \frac{\pi k}{2} \right)
\]

\textit{L-Lipschitz:}
\[
L = \max |f'(x)| = 1
\]

\textit{L-Smoothness:}
\[
L = \max |f''(x)| = \frac{4}{3\sqrt{3}}
\]

13. \textbf{Quadratic}:
\[
f(x) = \frac{1}{1+(ax)^2}
\]

\textit{Gradient:}
\[
f'(x) = -\frac{2 a^2 x}{\left( a^2 x^2 + 1 \right)^2}
\]

\textit{Curvature:}
\[
f''(x) = \frac{6 a^4 x^2 - 2 a^2}{(a^2 x^2 + 1)^3}
\]

\textit{Fourier Transform:}
\[
\mathcal{F}\{f(x)\}(k) = \sqrt{\frac{\pi}{2a^2}}e^{-\frac{k}{a}}, \text{for } a >0, k>0
\]

14. \textbf{MultiQuadratic}:
\[
f(x) = \frac{1}{\sqrt{1+(ax)^2}}
\]

\textit{Gradient:}
\[
f'(x) = -\frac{a^2 x}{\left( a^2 x^2 + 1 \right)^{3/2}}
\]

\textit{Curvature:}
\[
f''(x) = \frac{a^2 \left( 2 a^2 x^2 - 1 \right)}{(a^2 x^2 + 1)^{5/2}}
\]

\textit{Fourier Transform:}
\[
\mathcal{F}\{f(x)\}(k) = \frac{2 K_0\left(\frac{|k|}{\sqrt{a^2}}\right)}{\sqrt{a^2}}
\] where $K_n(x)$ is the modified Bessel function of the second kind.

\section{Insights, One by One}
\label{sec:Insights_One_by_One}
This section serves as a supplementary explanation to Sec. \ref{sec:benchmark_leaderboard_insights}. To thoroughly investigate the characteristics of different model modules, we present a detailed analysis in this section, combining theoretical insights and experimental results to examine each observation in turn.

\subsection{Coordinate Models}
\label{sec:appendix_Coordinate_Models}

This section focuses on analyzing the characteristics of Coordinate-MLPs and Coordinate-KANs.

\textbf{1. The computational complexity of Coordinate-KANs is higher than that of Coordinate-MLPs.} 
As discussed in Sec. \ref{sec:cn_model}, unlike the linear transformations with non-trainable nonlinear activation functions employed by Coordinate-MLPs, Coordinate-KANs utilize trainable nonlinear activation functions as the ``edges" of the network, resulting in higher computational complexity. Furthermore, the computational framework of Coordinate-KANs is not fully optimized for the parallel processing capabilities of underlying GPUs, leading to suboptimal efficiency and requiring longer training times.
Table \ref{tab:time_img_reg} presents a comparison of the training time, GPU utilization, and performance between various Coordinate-MLPs and Coordinate-KANs in image regression tasks. The training time for Bspline-KAN is approximately twice that of MLPs. In contrast, the optimized RBF-KAN and Fourier-KAN achieve training times comparable to MLPs but require significantly higher GPU computational resources.
\begin{table}[!htbp]
    \centering
    \caption{The comparison of training time and GPU memory in image regression task.}
\resizebox{0.48\textwidth}{!}{
    \begin{tabular}{c|ccc|ccc}
        \hline
        & \multicolumn{3}{c|}{Coordinate-KANs} &  \multicolumn{3}{|c}{Coordinate-MLPs} \\
        \hline
        Model & Bspline & RBF &  Fourier & ReLU & Gaussian & Sine \\
        \hline
        Time$^\downarrow$(m) & 38.43 & 25.24 & 24.47 & 21.96 & 23.46 & 24.11  \\
        GPU MEM$^\downarrow$(MB) & 1552 & 638 & 818 & 496 & 540 & 520  \\
        PSNR$^\uparrow$(dB) & 21.99 & 21.62 & 33.56 & 22.99 & 36.72 & 23.47 \\
       \hline
    \end{tabular}
}
    \label{tab:time_img_reg}
\end{table}

\textbf{2. Coordinate-KANs exhibit limited task generalization capabilities.}
As shown in Table \ref{tab:benchmark_leaderboard}, the optimal model for different implicit tasks varies among the different Coordinate-KANs. This indicates that while certain KAN models may perform well on specific tasks, they lack the ability to generalize across a diverse range of implicit tasks. Furthermore, experimental results reveal that even for the same KAN model, different network architectures (e.g., depth, width) and parameters are often required to achieve convergence for different tasks. This contrasts with MLPs, which consistently converge across various tasks using the same network structure, albeit without necessarily achieving optimal performance.

\textbf{3. The parameter count of Coordinate-KANs is difficult to control.} For MLPs, since activation functions are non-trainable, all parameters are confined to the linear transformations. As a result, different Coordinate-MLPs with the same network architecture typically exhibit similar parameter counts. In contrast, for Coordinate-KANs, trainable parameters originate from the basis functions, and the parameter requirements vary across different basis functions. This leads to substantial differences in parameter counts among various Coordinate-KANs, even when the network architecture is identical. For instance, in a six-layer network with a width of 64, the parameter count of Fourier-KAN is approximately double that of RBF-KAN. Additionally, changes in input dimensionality and the embedding of positional encodings can further cause significant variations in the parameter count of Coordinate-KANs.

\textbf{4. Coordinate-KANs are particularly well-suited for learning low-frequency features.}  
As discussed in Sec. \ref{sec:Spectral_Bias_of_KAN}, KANs exhibit weaker spectral bias in the low-frequency domain compared to MLPs. This characteristic allows KANs to process signals across different frequencies more uniformly, mitigating the tendency to overfit low-frequency components, which often results in sensitivity to high-frequency noise. Such a property enables KANs to effectively remove high-frequency noise while preserving the integrity of global information. As demonstrated in Table \ref{tab:benchmark_leaderboard}, Laguerre-KAN achieves the best performance in image denoising tasks, outperforming the second-best model (FKAN+Gaussian-MLP) by 1.86 dB.

\textbf{5. As decoders for high-dimensional features, Coordinate-KANs and Coordinate-MLPs exhibit minimal differences, with their performance primarily dependent on the mapping capabilities of their respective nonlinear primitives.}
Due to the difficulty of achieving convergence with Coordinate-KANs in the naive NeRF \cite{mildenhall2020nerf} model, we evaluated the representational capabilities of Coordinate-KANs in radiance fields using the NGP \cite{muller2022ngp} model. Unlike naive NeRF, NGP stores the scene in a hash table and uses a shallow neural network (1–2 layers) to map the high-dimensional features from the hash table to color and density. This approach contrasts with implicit neural representations that take low-dimensional coordinates as input, as the primary function of the network in NGP is the mapping of high-dimensional features. As shown in Table \ref{tab:benchmark_leaderboard} (``*" indicates results tested on NGP), the overall performance of Coordinate-KANs and Coordinate-MLPs on NGP is comparable, with a difference of only 0.04 dB between the best-performing models. However, the choice of nonlinear primitives significantly affects representational capabilities. For Coordinate-MLPs, the performance gap between the best model (Sigmoid-MLP) and the worst model (Sinc-MLP) is 22.82 dB. Similarly, for Coordinate-KANs, the difference between the best model (DoG-KAN) and the worst model (Shannon-KAN) is 23.44 dB.

\textbf{6. Certain specialized Coordinate-KANs may be better suited for solving partial differential equations (PDEs).}
To evaluate the potential of the models in solving partial differential equations (PDEs), we designed a Poisson reconstruction task for images. This task involves reconstructing an image from its gradients and Laplacian, essentially requiring the network to solve the Poisson equation. As shown in Table \ref{tab:benchmark_leaderboard}, for Coordinate-MLPs, most models struggle to produce high-quality reconstructions, with the exception of those using sine-based activation functions. In contrast, certain specialized basis functions in Coordinate-KANs demonstrate significantly superior performance for this task. For example, Chebyshev2-KAN outperforms the second-best model (RFF+Tanh-MLP) by 7.19 dB in image reconstruction quality. This indicates that carefully designed basis functions may offer substantial advantages in solving PDEs and hold great promise for applications in AI4Science.

\textbf{7. Overall, Coordinate-MLPs are better suited for implicit neural representation tasks.} For the majority of such tasks, the primary objective is to effectively capture the high-frequency components of a signal while learning its low-frequency aspects, enabling the network to model more complex and detailed information. Although Coordinate-MLPs exhibit slightly higher spectral bias in the low-frequency domain compared to Coordinate-KANs, the use of positional encoding can effectively mitigate this bias, resulting in a more balanced representation of different frequency components within the signal. Additionally, Coordinate-MLPs offer advantages such as faster training and easier control over the number of parameters. Therefore, as shown in Table \ref{tab:benchmark_leaderboard}, Coordinate-MLPs are generally more suitable for implicit neural representation tasks; however, different models may be selected for specific tasks to achieve optimal performance.

\subsection{Positional Encoding}
\label{sec:appebdix_Positional_Encoding}
This section primarily provides a detailed discussion on the impact of positional encoding (\texttt{NeRF}, \texttt{RFF} and \texttt{FKAN}) on the model.

\textbf{1. Fourier-based positional encoding effectively enhances the model's capability to represent high-frequency information.}  
As discussed in Sec. \ref{sec:Details_of_Positional_Encoding}, positional encoding leverages Fourier features to map input coordinates to a higher-dimensional space, introducing high-frequency oscillations through sine and cosine functions. This process modulates the NTK to exhibit translational invariance, thereby increasing the network's sensitivity to small variations in input coordinates. Consequently, the network becomes capable of capturing finer-grained high-frequency details. As shown in Table \ref{tab:benchmark_leaderboard}, in forward problem tasks such as audio regression, image regression, and SDF regression, positional encoding generally improves the representation ability of all activation functions for audio signals. For instance, in audio representation tasks characterized by high-frequency and locally periodic patterns, most Coordinate-MLPs fail to converge without positional encoding, with the exception of sine-based activations.

\textbf{2. Positional encoding hinders the performance of sine-type activation functions in inverse problem solving.}  
As indicated in Table \ref{tab:benchmark_leaderboard}, while positional encoding significantly enhances the representational capacity of sine-type activation functions (e.g., Sine and ScaledSine) in forward problem tasks, it adversely affects their generalization ability in inverse problems. This suppression ultimately leads to a decline in representation performance. For instance, in image denoising tasks, the incorporation of positional encoding results in a degradation of approximately 5 dB in the representational performance of Sine-MLP.

\textbf{3. \texttt{NeRF} and \texttt{RFF} positional encodings may hinder the ability of Coordinate-MLPs to solve PDEs, whereas the \texttt{FKAN} positional encoding facilitates their performance in such tasks.}  
In the Poisson reconstruction task, which fundamentally involves solving the Poisson equation to recover an image from its gradients and Laplacians. \texttt{NeRF} and \texttt{RFF} positional encodings prove inadequate. These encodings lack learnable parameters, making their simple mappings challenging to generalize and unsuitable for computing gradients and Laplacians within the network. In contrast, our proposed \texttt{FKAN} positional encoding, being fully learnable, integrates seamlessly into the computational graph. It not only enhances the ability of Coordinate-MLPs to learn high-frequency components but also simplifies the computation of gradients and Laplacians. As shown in Table \ref{tab:benchmark_leaderboard}, the Poisson reconstruction task requires simultaneous attention to reconstruction performance on  image (Img.), gradients (Grad.) and Laplacians (Lap.). For instance, while \texttt{NeRF} and \texttt{RFF} positional encodings improve the PSNR on the image (Img.) of ReLU-MLPs, they lead to a decrease in performance on Grad. and Lap., indicating a reduction in their inverse problem-solving capability. In contrast, the \texttt{FKAN} positional encoding significantly enhances the inverse representational ability of ReLU-MLPs, with an improvement of approximately 11.69 dB on Grad. and 1.26 dB on Lap.

\textbf{4. \texttt{FKAN} position encoding significantly enhances the implicit representation capability of non-periodic activation function-based Coordinate-MLPs.}
For periodic activation functions (Sine and ScaledSine), \texttt{FKAN} has minimal impact on model representation due to their inherent ability to perform frequency decomposition and learn high-frequency components. However, for non-periodic activation functions (e.g., ReLU, Gaussian, Gabor, and Sigmoid), our proposed \texttt{FKAN} encoding outperforms \texttt{NeRF} and \texttt{RFF} position encodings by adaptively adjusting the bandwidth of NTK. This adjustment enables a more refined capture of different frequency components within the signal, leading to a substantial improvement in the implicit representation capability of these MLP models. As demonstrated in Table \ref{tab:benchmark_leaderboard}, incorporating \texttt{FKAN} position encoding consistently achieves state-of-the-art performance in MLP models with non-periodic activation functions. For example, in the case of the Gaussian-MLP, \texttt{FKAN} improves the model’s performance by 26.19 dB in audio regression, 0.005 (IoU) in SDF regression, 4.34 dB in image inpainting, 10.56 dB in image denoising, and 18.34 dB in CT reconstruction.

\textbf{5. For signals with relatively uniform frequency distributions (e.g., audio and images), the learnable \texttt{FKAN} exhibits superior generalization capabilities. In contrast, for signals with more complex frequency distributions, such as neural radiance fields, the fixed-frequency \texttt{NeRF} position encoding proves more effective.}
As shown in Table \ref{tab:benchmark_leaderboard}, in implicit representation tasks involving audio and images, the frequency distribution is relatively uniform (with frequencies differing only slightly), allowing position encodings like \texttt{RFF} and \texttt{FKAN}, which control the NTK bandwidth, to capture these frequency components evenly through parameters such as the Gaussian scale in \texttt{RFF} and the maximum frequency threshold in \texttt{FKAN}. This results in a significant enhancement of the model’s implicit representation capacity. However, in the case of neural radiance fields, \texttt{RFF} requires extensive hyperparameter search to identify the optimal Gaussian scale for different scenarios. Similarly, \texttt{FKAN}, with its limited number of learnable parameters, struggles to capture all frequency signals effectively. In this context, the fixed-frequency \texttt{NeRF} position encoding can more uniformly and stably learn the different frequencies. Experimental results further demonstrate that in neural radiance field tasks, the optimal models incorporate the \texttt{NeRF} position encoding within Coordinate-MLPs. Additionally, \texttt{FKAN}’s learnable parameters offer better generalization than \texttt{RFF}, thus, models embedded with \texttt{FKAN} outperform those with \texttt{RFF} in this context.

\subsection{Nonlinear Primitives}
\label{sec:appendix_Nonlinear_Primitives}
This section primarily investigates in detail the impact of various nonlinear primitives (activation functions and basis functions) on the implicit representation capability of the model.

\textbf{1. ReLU-type MLPs exhibit a pronounced spectral bias, making it challenging to learn high-frequency components of signals.} According to Theorem \ref{thm:relu_sb}, the condition number of the Hessian matrix associated with a two-layer ReLU-MLP with width \(n\) scales as \(n^4\), highlighting the strong spectral bias inherent to ReLU-MLPs. Furthermore, the Fourier transform of the ReLU activation function reveals a frequency response that decays as the inverse square of the frequency, indicating a rapid decline in its ability to represent high-frequency signals. This leads to slower model convergence. As shown in Table \ref{tab:benchmark_leaderboard}, without position encoding, ReLU-MLPs fail to capture the high-frequency and local periodic characteristics of audio signals, and their performance on other representation tasks also remains suboptimal.

\textbf{2. Sine-type MLPs effectively represent signals with uniform frequency distributions but struggle to learn complex frequency components, lacking the capability for multi-scale learning.} The Fourier transform of the Sine activation function indicates that its frequency response is concentrated at specific frequencies. Consequently, while Sine-type MLPs are well-suited for signals with evenly distributed frequencies, they are inadequate for representing signals with complex or diverse frequency distributions. As shown in Table \ref{tab:benchmark_leaderboard}, Sine-MLPs demonstrate significant advantages in representation tasks involving image and audio signals with relatively uniform frequency distributions. However, in radiance field tasks characterized by complex frequency distributions, Sine-MLPs encounter convergence issues (only 7.92dB) even when position encoding is embedded. A potential solution is to embed variable frequency bandwidths across different layers of Sine-MLPs, thereby enabling the model to acquire multi-scale frequency learning capabilities \cite{lindell2022bacon}.

\textbf{3. Gaussian-type MLPs exhibit strong responsiveness to low-frequency signals, making them particularly advantageous for low-frequency representation tasks such as image denoising.} The Fourier transform of Gaussian activation functions demonstrates an exponentially decaying frequency response, which explains their strong affinity for low-frequency components. However, their decay is slower compared to ReLU-type activation functions, granting Gaussian-type MLPs enhanced high-frequency representation capabilities relative to ReLU-MLPs. Furthermore, the L-Lipschitz and L-smoothness properties of Gaussian activation functions suggest that smaller variance parameters (\(\sigma\)) can yield larger NTK eigenvalues, accelerating model convergence. However, excessively small \(\sigma\) values increase the NTK spectral width, potentially leading to training instability. As shown in Table \ref{tab:benchmark_leaderboard}, the task of image denoising involves learning low-frequency information while suppressing high-frequency noise. By embedding \texttt{FKAN} position encoding, Gaussian-MLPs achieve optimal performance, surpassing Sine-MLPs by 3.46 dB.

\textbf{4. Stable frequency response facilitates the solution of PDEs (e.g., Poisson reconstruction).}  
The Poisson image reconstruction task is inherently the solution of the Poisson equation, where the network’s gradients and Laplacian operator are employed to recover the original image. On one hand, the solution to the Poisson equation typically manifests as a smooth function, with its frequency spectrum primarily concentrated in the low-frequency domain. Therefore, stable frequency responses can effectively capture these low-frequency components, ensuring an accurate reconstruction of the solution. On the other hand, during Poisson reconstruction, high-frequency noise or irregular boundary conditions may arise. Stable frequency responses help smooth the network’s derivatives and Laplacian calculations, thereby ensuring the stability of the model during training. Through Fourier transform analysis of activation functions, it is observed that functions such as Sine ($\delta(\omega-k)$), Tanh ($e^{-k}, e^k$), and Quadratic ($e^{-k}$) exhibit relatively stable frequency responses. This characteristic ensures that, when solving the Poisson equation, the numerical solutions experience minimal fluctuations, effectively preventing issues such as gradient explosion or vanishing. As shown in Table \ref{tab:benchmark_leaderboard}, Sine, Tanh, and Quadratic-type MLPs demonstrate significant advantages in Poisson reconstruction tasks.

\textbf{5. In Coordinate-KANs, different basis functions exhibit significant adaptability to various tasks.}
In contrast to Coordinate-MLPs, in Coordinate-KANs, the choice of basis functions results in significant differences in the network architecture, parameter count, and optimization characteristics. As a result, different basis functions demonstrate notably distinct representational capabilities across various implicit task representations. As shown in Table \ref{tab:benchmark_leaderboard}, in audio representation tasks, due to the high-frequency and local periodicity of audio signals, only periodic basis functions such as Fourier-KAN and Sine-KAN are able to converge. For Morlet-KAN, although it struggles to converge in inverse inference tasks like CT reconstruction and Poisson reconstruction, it performs well in SDF regression tasks. In contrast, Chebyshev-type KANs exhibit significant advantages in CT and Poisson reconstruction tasks. Furthermore, Laguerre-KAN achieves optimal performance in image denoising tasks, outperforming the best-performing Coordinate-MLPs (FKAN+Gaussian-MLPs) by 1.86 dB.

\section{Details of INR Tasks}
\label{sec:Details_of_INR_Tasks}
This section provides a detailed introduction to the nine implicit representations tested in the experiments, including the datasets, preprocessing procedures, and other relevant details.

\textbf{Audio Regression.}
\textit{Dataset: } We utilize audio segments provided by Siren \cite{sitzmann2020siren} and the GTZAN \cite{tzanetakis2001gtzan} dataset as the test datasets. The evaluation metrics employed are primarily the Signal-to-Noise Ratio (SNR) \cite{roux2018snr} and the Log-Spectral Distance (LSD) \cite{gray1976lsd}. Since LSD is an indirect measure for evaluation in the frequency domain, we focus primarily on the SNR metric.

\textbf{Image Regression.} 
\textit{Dataset: } We utilize the ``natural'' dataset provided by FFN \cite{tancik2020ffn} to evaluate the image regression task. This dataset contains a selection of $512\times512$ resolution images specifically created to assess the network's ability to learn high-frequency components.

\textbf{3D Shape Regression.} 
\textit{Dataset: } We utilized the Stanford 3D Scanning dataset \cite{Stanford3DScan}, in which all models were acquired using the Cyberware 3030MS optical triangulation scanner. The dataset features high-density polygonal representations, providing detailed and precise 3D shape data to support our research.

\textit{Preprocessing: } We perform point sampling on a \(512 \times 512 \times 512\) grid to generate an occupancy volume. Each voxel inside the volume is assigned a value of 1, while voxels outside the volume are assigned a value of 0, thereby constructing the ground truth Signed Distance Field.

\textbf{Image Inpainting.}
\textit{Dataset: } We utilize the images provided in the Dalmia/siren repository \cite{aman_dalmia_2020_3902941} as the experimental data for our study.

\textit{Preprocessing: } In an image, 20\% of the pixels are randomly selected as the training data, while the remaining 80\% are used as the test data.

\textbf{Image Super-Resolution.} 

\textit{Dataset: } We use the DIV2K \cite{Agustsson2017div2k} dataset, which provides high-quality, high-resolution images.

\textit{Preprocessing: } The images are downsampled by a factor of four to generate low-resolution data (training data), while the original images are used as high-resolution data (test data).

\begin{figure*}[h!]
    \centering
    \begin{minipage}{0.47\textwidth}
        \centering
        \includegraphics[width=\linewidth]{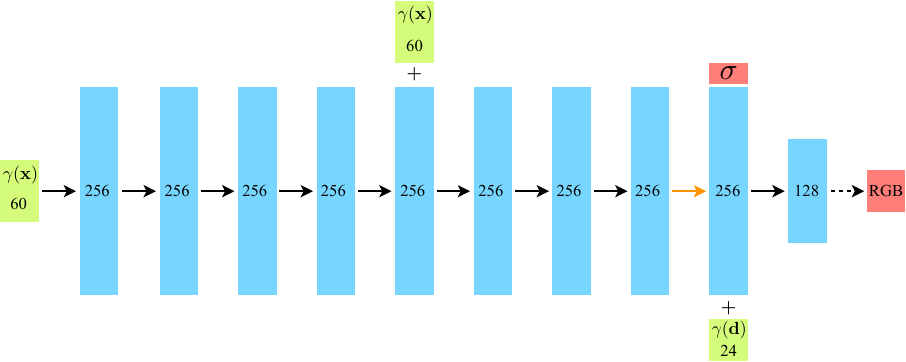}
        \caption{The network of vanilla NeRF. The image is sourced from \cite{mildenhall2020nerf}.}
    \end{minipage}\hfill
    \begin{minipage}{0.50\textwidth}
        \centering
        \includegraphics[width=\linewidth]{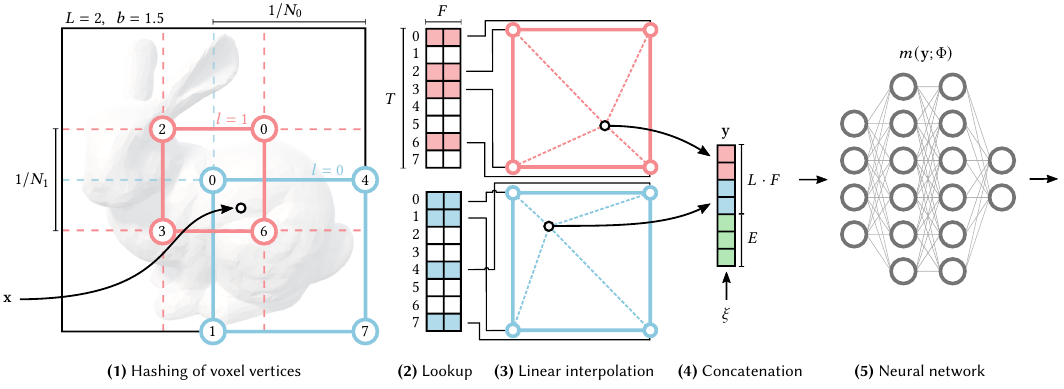}
        \caption{The network of NGP. The image is sourced from \cite{muller2022ngp}.}
    \end{minipage}
    \caption{\textbf{The role of the network in neural radiance fields.} In vanilla NeRF, the network takes low-dimensional coordinates as input, essentially performing low-dimensional regression. In contrast, the NGP architecture utilizes high-dimensional hash features as input, fundamentally functioning as a high-dimensional feature mapping.}
    \label{fig:nerf_vs_ngp}
\end{figure*}

\textbf{Image Denoising.}
\textit{Dataset: } The DIV2K dataset \cite{Agustsson2017div2k} is used to evaluate the performance of the model in the image denoising task.

\textit{Preprocessing: }  Given that the original images have a 2K resolution, we first downsample them by a factor of four. We then introduce readout noise and photon noise into the images to create noisy versions, where each pixel is subjected to an independent Poisson-distributed random variable.

\textbf{Poisson Reconstruction.}

\textit{Dataset: } Grayscale images from the skimage library \cite{van2014scikit} are used to evaluate the Poisson reconstruction of the images.

\textbf{CT Reconstruction.} 

\textit{Dataset: } We conduct experiments using the Kaggle Lung Nodule Analysis dataset \cite{9022282}.

\textit{Preprocessing: }  Only 150 measurements are used to supervise the reconstruction of the density field.

\textbf{Neural Radiance Fields.}

\textit{Dataset: } The NeRF Blender dataset \cite{mildenhall2020nerf} is used to evaluate the capability of novel view synthesis in radiance fields.

\textit{Radiance Field Model: } It is noteworthy that while the purely implicit, vanilla NeRF \cite{mildenhall2020nerf} effectively evaluates the multi-scale learning and complex representation capabilities of various models, our experiments reveal that the low-frequency characteristics of Coordinate-KANs lead to poor convergence. As a result, we utilize the NGP \cite{muller2022ngp} model to investigate the potential of KANs compared to MLPs within NGP radiance fields. As shown in Fig. \ref{fig:nerf_vs_ngp}, unlike vanilla NeRF, NGP stores the scene in a hash table without requiring additional positional encoding, with the network serving as a feature extractor that maps high-dimensional features to color and density.

\textit{Experimental Model Design: }
For Coordinate-MLPs, as illustrated in Fig. \ref{fig:nerf_vs_ngp}(a), we evaluate the representational performance of models in a vanilla NeRF under different combinations of positional encodings (\(\gamma(\mathbf{x})\)) and activation functions (\(\sigma\)). For Coordinate-KANs, due to their difficulty in converging on vanilla NeRF, the evaluation is conducted using the NGP model. Specifically, as shown in Fig. \ref{fig:nerf_vs_ngp}(b), we assess their radiance field representation capabilities by replacing different neural network modules (KANs and MLPs without positional encoding). As shown in Table \ref{tab:benchmark_leaderboard}, in the radiance field representation task, the ``*" symbol denotes results evaluated on the NGP model. Overall, as high-dimensional feature mappers, Coordinate-MLPs and Coordinate-KANs exhibit comparable performance, with their primary differences arising from the nonlinear mapping capabilities of their respective nonlinear primitives.

\section{Further Experiments}
\label{sec:Further_Experiments}
This section serves as a supplementary explanation to Sec. \ref{sec:experiments} on experiments, primarily detailing experimental setup and configurations, parameter sensitivity analysis, and qualitative experiments.

\subsection{Experimental Setup}
\label{sec:Experimental_Setup}
This section provides detailed information on the experimental setup, including hardware specifications and parameter configurations.

\textbf{Hardware:} All experiments for neural radiance fields are conducted on an Nvidia A800 GPU, while experiments for other implicit neural representation tasks are performed on two Nvidia RTX 3090 GPUs.

\textbf{Network Settings:} For neural radiance field task, the model architectures are detailed in \ref{sec:Details_of_INR_Tasks}, primarily including the vanilla NeRF model and the NGP model. For other implicit tasks, all Coordinate-MLPs are 6-layer MLP networks with a width of 256, and all Coordinate-KANs are 6-layer KAN networks with a width of 64.

\textbf{Network Optimization Settings:} All models are trained using the Adam optimizer, with the learning rate adjusted via the CosineAnnealingLR scheduler. The initial learning rate is set to \(4 \times 10^{-4}\), and the batch size is 8192.

\textbf{Positional Encoding Settings:} For \texttt{RFF} positional encoding, the Gaussian variance \(\sigma\) is set to 10, and the frequency dimension \(L\) is 32. For \texttt{NeRF} positional encoding, the frequency dimension is defined as \(\log(1 / (2 \times (2 \times 1 / B)))\), where \(B\) denotes the batch size. For \texttt{FKAN} positional encoding, the maximum frequency threshold \(\Omega\) is set to 1024.

\textbf{Nonlinear Basis/Activation Functions:} For the Gaussian and Laplacian activation functions, the Gaussian variance parameter \(\sigma\) is set to 0.1. For the Sine/ScaledSine activation functions, the frequency parameter \(\omega\) is set to 3000 for audio representation tasks and 30 for other tasks. For the Quadratic/MultiQuadratic activation functions, the parameter \(a\) is set to 1.0.

\textbf{Evaluation Metrics:}
The evaluation metrics employed for audio regression tasks include the Signal-to-Noise Ratio (SNR) and the Log-Spectral Distance (LSD). The SNR is defined as follows \cite{roux2018snr}:  
$$
\operatorname{SNR}(\hat{y}, y) = 20 \log_{10} \left( \frac{\|y\|_2^2}{\|\hat{y} - y\|_2^2} \right),
$$  
where \(y\) denotes the reference signal and \(\hat{y}\) its approximation. SNR is widely used in signal processing to evaluate the fidelity of signal approximations. The LSD measures reconstruction quality across individual frequencies and is expressed as \cite{gray1976lsd}:  
$$
\operatorname{LSD}(\hat{y}, y) = \frac{1}{L} \sum_{\ell=1}^L \sqrt{\frac{1}{K} \sum_{k=1}^K \left(X(\ell, k) - \hat{X}(\ell, k)\right)^2},
$$  
where \(X\) and \(\hat{X}\) represent the log-spectral power magnitudes of \(y\) and \(\hat{y}\), respectively, computed as \(X = \log |S|^2\), with \(S\) being the short-time Fourier transform (STFT) of the signal. The indices \(\ell\) and \(k\) correspond to frames and frequencies, respectively. In this study, frames are configured with a length of 2048. Given that LSD is an indirect measure evaluating the frequency domain, our primary focus remains on the SNR metric.

For image-related tasks, Peak Signal-to-Noise Ratio (PSNR) is employed to evaluate representation quality. For Signed Distance Function (SDF) regression tasks, Intersection over Union (IoU) is utilized to assess reconstruction accuracy.

\subsection{Parameter Sensitivity Analysis}
\label{sec:Parameter_Sensitivity_Analysis}
This section investigates the representational performance of positional encodings and activation functions under various hyperparameter configurations.

\textbf{Experimental Setup:} To explore the parameter sensitivity of positional encoding and activation functions, experiments are conducted on audio  regression task. Three audio samples are selected for evaluation: ``Bach(7s)'', ``Counting(7s)'' \cite{sitzmann2020siren}, and ``Blues(30s)''\cite{tzanetakis2001gtzan}.

\textbf{Parameter Sensitivity of Positional Encoding:}  
The parameter sensitivity experiments for positional encodings in audio representation tasks are presented in Tables \ref{tab:nerf_l}, \ref{tab:rff_l}, \ref{tab:rff_sigma}, and \ref{tab:fkan_omega}. These include the sensitivity of \texttt{NeRF} positional encoding to the frequency dimension parameter \(L\) (Table  \ref{tab:nerf_l}), \texttt{RFF} positional encoding to the frequency dimension \(L\) (Table \ref{tab:rff_l}) and Gaussian scale parameter \(\sigma\) (Table \ref{tab:rff_sigma}), and \texttt{FKAN} positional encoding to the maximum frequency threshold parameter \(\Omega\) (Table \ref{tab:fkan_omega}). The experiments indicate that (1) \texttt{NeRF} positional encoding is relatively insensitive to the frequency dimension parameter \(L\), with \(L = 16\) generally yielding optimal results. However, due to NeRF's inability to dynamically modulate the bandwidth of the NTK spectrum, its generalization capability is limited (see Table \ref{tab:benchmark_leaderboard}).  (2) For \texttt{RFF} positional encoding, two hyperparameters are involved: the frequency dimension \(L\) and the Gaussian scale parameter \(\sigma\). While \texttt{RFF} is relatively insensitive to \(\sigma\) (with \(\sqrt{\sigma} = 1000\) performing well in experiments), it is more sensitive to \(L\). The optimal \(L\) varies inconsistently across different datasets and activation functions, and the interplay between these two parameters exacerbates \texttt{RFF}'s sensitivity to hyperparameter tuning.  (3) In comparison, \texttt{FKAN} positional encoding not only allows for control over the NTK spectrum's bandwidth, enhancing model generalization, but also simplifies hyperparameter tuning with a single maximum frequency threshold parameter \(\Omega\). Furthermore, \texttt{FKAN} is relatively insensitive to \(\Omega\), with \(\Omega = 1024\) providing favorable results in experiments.

\begin{table}[!htbp]
    \centering
  \caption{The impact of the parameter $L$ in \texttt{NeRF} positional encoding (Eq. \ref{eq:pe_nerf}) on the quality of audio representation. For each row, best results are highlighted as \colorbox{colorFst}{first}, \colorbox{colorSnd}{second},  and \colorbox{colorTrd}{third}.}
\resizebox{0.75\linewidth}{!}{
  \begin{tabular}{c|c|ccccccc}
    \cline{1-9}

    & \multicolumn{2}{c}{\textbf{Parameter} $L$} & 2 &  4 & 8  & 16 & 32 & 64 \\
    \cdashline{2-9}
     Dataset & \multicolumn{2}{c}{Params} & 264K & 265K  &  268K & 272K &  280K & 296K  \\
    
    \cline{1-9}
   \multirow{8}{*}{Bach(\texttt{7s})} & \multirow{2}{*}{Gaussian} & SNR$^\uparrow$ & 16.12 & \fs 29.08  & \nd{ 25.16} & 19.86 & 21.06 & \rd{ 23.48}  \\
    & & LSD$^\downarrow$ & \nd{ 0.810} & \fs 0.737 & \rd{ 1.785} & 2.115 & 2.035 &1.880 \\ 

  \cdashline{2-9}
   &\multirow{2}{*}{ReLU} & SNR$^\uparrow$ &0.00 & 0.02 & 4.60 & \fs 24.76 & \nd{ 5.37} & \rd{ 4.89} \\
    & & LSD$^\downarrow$ & 3.997 & 3.590 & \fs 1.151 & \nd{ 1.590} & \rd{ 2.812 }& 3.000  \\ 

    \cdashline{2-9}
   &\multirow{2}{*}{Sine} & SNR$^\uparrow$ & 12.12 & 14.75 & \rd{ 34.63} & \fs 46.40 & 34.36 & \nd{ 37.62} \\
    & & LSD$^\downarrow$ & 0.914 & \rd{ 0.864} & \nd{ 0.663} & \fs 0.530 &1.328 & 1.188  \\ 

    \cdashline{2-9}
   &\multirow{2}{*}{ScaledSine} & SNR$^\uparrow$ & 14.25 & 15.70 & 33.38 & \nd{ 36.94} & \rd{ 33.46} & \fs 37.53  \\
    & & LSD$^\downarrow$ & \nd{ 0.814 }& \rd{ 0.837} & \fs 0.663 & 1.120&1.332 & 1.145  \\ 
  \hline
  \hline

   \cline{1-9}
   \multirow{8}{*}{Counting(\texttt{7s})} & \multirow{2}{*}{Gaussian} & SNR$^\uparrow$ &7.50  & 9.89 & \fs 19.31& 9.36 &\nd{  10.71} & \rd{ 10.66}  \\
    & & LSD$^\downarrow$ & \nd{ 1.760} & \fs 1.567 & \rd{ 2.589} & 3.429 & 3.323 & 3.221 \\

  \cdashline{2-9}
   &\multirow{2}{*}{ReLU} & SNR$^\uparrow$ & 0.00 & 0.01 & 0.59 &\fs  13.27 & \nd{ 6.64} &  \rd{ 6.47} \\
    & & LSD$^\downarrow$ & 3.548 & 3.256 & \nd{ 1.938} &\fs 1.326 & \rd{ 2.618} & 2.812   \\

    \cdashline{2-9}
   &\multirow{2}{*}{Sine} & SNR$^\uparrow$ & 7.04 & 10.36 & 12.73 & \fs 35.46 & \rd{ 24.16} & \nd{ 27.08} \\
    & & LSD$^\downarrow$ & \rd{ 1.723 }& 1.803 & \nd{ 1.713} & \fs 1.377 & 2.216 & 1.979 \\

    \cdashline{2-9}
   &\multirow{2}{*}{ScaledSine} & SNR$^\uparrow$ & 8.65 & 11.03  &12.55 & \fs 28.40 & \rd{ 13.06} & \nd{ 14.14}  \\
    & & LSD$^\downarrow$ & \fs 1.649 & \rd{ 1.784} & \nd{ 1.752} &1.855 & 3.145& 3.056  \\ 

    \hline
    \hline

   \cline{1-9}
   \multirow{8}{*}{Blues(\texttt{30s})} & \multirow{2}{*}{Gaussian} & SNR$^\uparrow$ & 3.12 & \nd{ 8.84} & \fs 19.82 & 7.81 & 8.50 & \rd{ 8.78}  \\
    & & LSD$^\downarrow$ &2.057 & 1.580 & \fs 0.849 & 1.594 & \rd{ 1.553} & \nd{ 1.536} \\

  \cdashline{2-9}
   &\multirow{2}{*}{ReLU} & SNR$^\uparrow$ & 0.00  & 0.01 & 0.67& \fs 14.74& \rd{ 2.58} & \nd{ 2.81}  \\
    & & LSD$^\downarrow$ & 5.140 & 4.708 & 2.843 &\fs 0.890 & \nd{ 1.730} & \rd{ 1.760 }\\

    \cdashline{2-9}
   &\multirow{2}{*}{Sine} & SNR$^\uparrow$ & 7.39 & 9.76 & \nd{ 12.96} & \fs 24.08 & 10.80& \rd{ 11.68} \\
    & & LSD$^\downarrow$ & 1.697 & 1.578  & \nd{ 1.360} & \fs 0.606 & 1.422 &  \rd{ 1.370} \\

    \cdashline{2-9}
   &\multirow{2}{*}{ScaledSine} & SNR$^\uparrow$ &6.81  & 8.22 & \nd{ 11.90} & \fs 22.86 &10.26 & \rd{ 11.09} \\
    & & LSD$^\downarrow$ & 1.751 & 1.667 & \rd{ 1.449} & \fs 0.670 & 1.454 & \nd{ 1.405} \\ 

    \hline
  \end{tabular}
}
  \label{tab:nerf_l}
\end{table}

\begin{table}[!htbp]
    \centering
  \caption{The impact of the parameter $L$ in \texttt{RFF} positional encoding (Eq. \ref{eq:pe_rff}) on the quality of audio representation.}
\resizebox{0.8\linewidth}{!}{
  \begin{tabular}{c|c|ccccccccc}
    \cline{1-11}

    & \multicolumn{2}{c}{\textbf{Parameter} $L$} & 2 &  4 & 8  & 16 & 32 & 64 & 128 & 256 \\
    \cdashline{2-11}
     Dataset & \multicolumn{2}{c}{Params} & 264K & 265K  & 267K  & 271K & 280K & 296K & 329K & 394K \\
    
    \cline{1-11}
   \multirow{8}{*}{Bach(\texttt{7s})} &\multirow{2}{*}{Gaussian} & SNR$^\uparrow$ & 18.61 & 19.41 & 19.67 & 20.21 & 20.84 & \rd{ 23.89} & \nd{ 27.06} & \fs 30.18 \\
    & & LSD$^\downarrow$ &2.199 & 2.143 & 2.125 & 2.090 & 2.046 & \rd{ 1.850} & \nd{ 1.680} & \fs 1.527 \\ 
    
    \cdashline{2-11}
    & \multirow{2}{*}{ReLU} & SNR$^\uparrow$ & 0.82 & 13.44 & 14.57 & 14.87 &  \fs 15.62 & 14.55 & \rd{ 14.94} & \nd{ 15.08} \\
    & & LSD$^\downarrow$ & 1.580 & 1.088 & 1.019 & 0.988 & 0.978 & \rd{ 0.951} &  \nd{ 0.932} & \fs 0.909 \\ 

    \cdashline{2-11}
    & \multirow{2}{*}{Sine} & SNR$^\uparrow$ & \fs 43.24 & \nd{ 41.97} & \rd{ 41.60} & 40.04 &39.02  & 38.27 & 38.37 &  38.82 \\
    & & LSD$^\downarrow$ & 0.778 & \nd{ 0.548} & \fs 0.547 & \rd{ 0.561} & 0.582 & 0.595 & 0.599 & 0.580 \\ 

     \cdashline{2-11}
    & \multirow{2}{*}{ScaledSine} & SNR$^\uparrow$ & \fs 41.82 & \nd{ 40.01} & \rd{ 39.84} & 39.02 & 38.10  & 37.17 & 36.71 & 36.94 \\
    & & LSD$^\downarrow$ & 0.877 & \fs 0.567 & \nd{ 0.568} & \rd{ 0.575} & 0.595 & 0.614 & 0.627 & 0.610 \\ 

      \hline
      \hline

   \multirow{8}{*}{Counting(\texttt{7s})} &\multirow{2}{*}{Gaussian} & SNR$^\uparrow$ & 8.73 & 9.37 & 9.69 & 10.33 & 12.14 & \fs 20.09 & \nd{ 16.78} & \rd{ 16.34}\\
    & & LSD$^\downarrow$ & 3.474 & 3.424 & 3.400 & 3.347 & 3.195 & \nd{ 2.335} & \fs 2.119  & \rd{ 2.445}\\
    
    \cdashline{2-11}
    & \multirow{2}{*}{ReLU} & SNR$^\uparrow$ & 0.74 & 4.09 & \nd{ 4.52} & \rd{ 4.48} & \fs 4.93 &  4.22 & 4.28 & 4.42\\
    & & LSD$^\downarrow$ & 1.690 & 1.489 & 1.430 & 1.412 & 1.400 &\rd{ 1.379}  & \nd{ 1.361} & \fs 1.354  \\ 

     \cdashline{2-11}
     & \multirow{2}{*}{Sine} & SNR$^\uparrow$ & \fs 28.64 & \nd{ 25.76} & \rd{ 17.62} &13.50 & 13.05  &12.97  & 13.02 & 12.92 \\
    & & LSD$^\downarrow$ & 1.396 & \fs 1.077 & \nd{ 1.184} & \rd{ 1.253} & 1.412 & 1.439 & 1.434 & 1.571 \\ 

     \cdashline{2-11}
    & \multirow{2}{*}{ScaledSine} & SNR$^\uparrow$ & \fs 27.59 & \nd{ 19.26} & \rd{ 13.34} &12.95 &12.86  & 12.83 & 12.81 & 12.80\\
    & & LSD$^\downarrow$ &\rd{ 1.332} & \fs 1.203 & \nd{ 1.269} & 1.472 & 1.558 & 1.649 & 1.648 & 1.682 \\ 

    \hline
    \hline
    \multirow{8}{*}{Blues(\texttt{30s})} &\multirow{2}{*}{Gaussian} & SNR$^\uparrow$ & 7.74 & 8.57  & 8.89 &9.66 &  11.74& \rd{ 18.06} & \nd{ 21.98} & \fs 23.16 \\
    & & LSD$^\downarrow$ & 1.593 & 1.541 & 1.522 & 1.475 & 1.348& \rd{ 0.975} & \nd{ 0.763}  & \fs 0.672 \\
    
    \cdashline{2-11}
    & \multirow{2}{*}{ReLU} & SNR$^\uparrow$ &  0.56 & 2.57 &2.88  & 2.91 & \fs 3.23 & 2.83 & \nd{ 2.92} & \rd{ 2.90}\\
    & & LSD$^\downarrow$ & 2.343& 1.964 & \nd{ 1.913} & 1.920 & \fs 1.855 & 1.950 & 1.920 & \rd{ 1.919} \\ 

     \cdashline{2-11}
    & \multirow{2}{*}{Sine} & SNR$^\uparrow$ & \fs 21.13 & \rd{ 19.91} & 19.01 & 17.68 & 17.08 & 17.29 & 18.35 & \nd{ 20.49} \\
    & & LSD$^\downarrow$ &\fs 0.742 & \nd{ 0.908} & \rd{ 0.986} & 1.078 & 1.130 & 1.103 & 1.035 & 0.874 \\ 

     \cdashline{2-11}
    & \multirow{2}{*}{ScaledSine} & SNR$^\uparrow$ & \fs 21.09 & \nd{ 19.48} & \rd{ 18.37} & 16.77 & 15.60 & 15.11 & 14.14 & 16.71 \\
    & & LSD$^\downarrow$ & \fs 0.765 & \nd{ 0.953} & \rd{ 1.047} & 1.445 & 1.215 & 1.123 & 1.172 & 1.137  \\ 
  \hline
  \end{tabular}
}
  \label{tab:rff_l}
\end{table}

\begin{table}[!ht]
    \centering
  \caption{The impact of the parameter $\sigma$ in \texttt{RFF} positional encoding (Eq. \ref{eq:pe_rff}) on the quality of audio representation.}
\resizebox{0.8\linewidth}{!}{
  \begin{tabular}{c|c|ccccccccc}
    \cline{1-11}

    Dataset & \multicolumn{2}{c}{\textbf{Parameter} $\sqrt{\sigma}$} & 1 &  20 & 40 & 60 & 80 & 100 & 1000 & 10000  \\
    
    \cline{1-11}
   \multirow{8}{*}{Bach(\texttt{7s})} &\multirow{2}{*}{Gaussian} & SNR$^\uparrow$ & 19.02 & \fs 27.72 & \nd{ 23.21} & \rd{ 21.68}  & 21.12 & 20.84 & 20.58 & 20.63  \\
    & & LSD$^\downarrow$ & \fs 0.801 & \nd{ 1.628} & \rd{ 1.899} & 1.993 & 2.028 & 2.046 & 2.064 & 2.063 \\ 
    
    \cdashline{2-11}
    & \multirow{2}{*}{ReLU} & SNR$^\uparrow$ & 0.01 & 4.42 & 9.69 & 11.91 & \rd{ 13.95} & \nd{ 15.62} & \fs 30.30 & 5.99 \\
    & & LSD$^\downarrow$ &3.968  & 1.165 & 0.988 & \nd{ 0.977} & \fs 0.970 & \rd{ 0.978} &  1.374 & 3.011 \\ 

    \cdashline{2-11}
    & \multirow{2}{*}{Sine} & SNR$^\uparrow$ & 12.00  & 28.38 & 34.43 & 36.92 & \rd{ 38.16}  & \nd{ 39.02} & \fs 45.27 & 33.23 \\
    & & LSD$^\downarrow$ &0.920 & 0.661 & 0.636 & 0.612 & \rd{ 0.592} & \nd{ 0.582} & \fs 0.517 & 1.344 \\

   \cdashline{2-11}
    & \multirow{2}{*}{ScaledSine} & SNR$^\uparrow$ & 13.71 & 24.91 & 33.03 & 36.04 & \rd{ 37.30}  & \nd{ 38.10 }& \fs 46.35 & 33.49 \\
    & & LSD$^\downarrow$ &0.859 & 0.694 & 0.631 & 0.624  & \rd{ 0.607 }& \nd{ 0.595} & \fs 0.516  & 1.330 \\ 
   
      \hline
      \hline
      
   \multirow{8}{*}{Counting(\texttt{7s})} &\multirow{2}{*}{Gaussian}  & SNR$^\uparrow$ & 8.12 & \fs  20.95 & \nd{ 16.45} & \rd{ 14.11} &  13.24 & 12.14 & 10.31 & 10.37 \\
    & & LSD$^\downarrow$ & \fs 1.651 & \nd{ 2.327} & \rd{ 2.849} & 3.041 & 3.108 & 3.195 & 3.353 & 3.349 \\ 
    
    \cdashline{2-11}
    & \multirow{2}{*}{ReLU} & SNR$^\uparrow$ & 0.00 & 0.589 & 1.40 & 2.61 &3.82 & \rd{ 4.93} & \fs 13.21 & \nd{ 8.58} \\
    & & LSD$^\downarrow$ & 3.417 & 1.964 & 1.630 & \rd{ 1.466} & \nd{ 1.402} & \fs 1.400 & 1.939  & 3.211  \\ 

    \cdashline{2-11}
    & \multirow{2}{*}{Sine} & SNR$^\uparrow$ & 6.29 & 12.63 & 12.75 & 12.85  & 12.91  &  \rd{ 13.06} & \fs 35.13 & \nd{ 22.59} \\
    & & LSD$^\downarrow$ & 1.666 & 1.706 & 1.685 & 1.640 & \rd{ 1.532}  & \nd{ 1.412} & \fs 1.113 & 2.364 \\

   \cdashline{2-11}
    & \multirow{2}{*}{ScaledSine} & SNR$^\uparrow$ & 6.23 & 12.13 & 12.56 & 12.78 & 12.84  &\rd{  12.86} & \fs 34.65  & 
    \nd{ 16.09} \\
    & & LSD$^\downarrow$ & 1.598 & 1.762 & 1.702 & 1.652 & \rd{ 1.585} & \nd{ 1.559} & \fs 1.114  & 2.890 \\ 

    \hline
    \hline
   \multirow{8}{*}{Blues(\texttt{30s})} &\multirow{2}{*}{Gaussian}  & SNR$^\uparrow$ & 4.85 & \fs 21.37 & \nd{ 17.33} & \rd{ 14.19} & 12.59  & 11.74 &8.56 & 8.57 \\
    & & LSD$^\downarrow$ & 1.876 &\fs  0.920 &\nd{ 1.026} & 1.211 & \rd{ 1.302 }& 1.348 & 1.549 & 1.549  \\ 
    
    \cdashline{2-11}
    & \multirow{2}{*}{ReLU} & SNR$^\uparrow$ & 0.00  & 0.66 & 1.39 & 2.08 & 2.68   &\rd{ 3.23}  & \fs 13.58 &  \nd{ 9.05}  \\
    & & LSD$^\downarrow$ & 4.893 & 2.826 & 2.386 & 2.136 & 1.986  &\rd{ 1.855}  & \fs 0.949 & \nd{ 1.374} \\ 

    \cdashline{2-11}
    & \multirow{2}{*}{Sine} & SNR$^\uparrow$ &4.91  &15.21  & 14.29 &15.40  & 16.22  & \nd{ 17.09} & \fs 23.53 & \rd{ 16.78}  \\
    & & LSD$^\downarrow$ &1.895 & 1.235 & 1.267 & 1.215 & 1.171 & \rd{ 1.130} & \fs 0.583 & \nd{ 1.084} \\

   \cdashline{2-11}
    & \multirow{2}{*}{ScaledSine} & SNR$^\uparrow$ & 3.866 & 10.75 & 12.08 &13.14  &  \rd{ 14.32} & \nd{ 15.60} & \fs 22.86 & 13.88  \\
    & & LSD$^\downarrow$ & 1.977 & 1.542 & 1.426 & 1.366 & 1.277 & \nd{ 1.215} & \fs 0.619 & \rd{ 1.250}  \\ 
  \hline
  \end{tabular}
}
 \label{tab:rff_sigma}
  
\end{table}

\begin{table}[!ht]
    \centering
  \caption{The impact of the parameter $\Omega$ in \texttt{FKAN} positional encoding (Eq. \ref{eq:pe_fkan_sm}) on the quality of audio representation.}
\resizebox{0.8\linewidth}{!}{
  \begin{tabular}{c|c|cccccccc}
    \cline{1-10}

    Dataset & \multicolumn{2}{c}{\textbf{Parameter} $\Omega$} & 16 &  32 & 64 & 128 & 256 & 512 & 1024  \\
    
    \cline{1-10}
   \multirow{8}{*}{Bach(\texttt{7s})} &\multirow{2}{*}{Gaussian} & SNR$^\uparrow$ & 6.38 & 11.22 & 15.03 & 20.22 & \nd 25.30 & \fs 25.88 &  \rd 25.17  \\
    & & LSD$^\downarrow$ & 1.73 & 1.66 & 1.60 & \fs 1.53 & \nd 1.55 & \rd 1.60 & 1.74 \\ 
    
    \cdashline{2-10}
    & \multirow{2}{*}{ReLU} & SNR$^\uparrow$ & 0.09 & 0.67 & 2.34 &  6.82 & \rd 11.88 & \nd 16.35 & \fs 24.73  \\
    & & LSD$^\downarrow$ & 1.88 & 1.61 & 1.46 & 1.39 & \rd 1.28 & \nd 1.22& \fs 1.17  \\ 

    \cdashline{2-10}
    & \multirow{2}{*}{Sine} & SNR$^\uparrow$ & 0.00 & 7.00 & 11.10 & 15.22 & \rd 21.30 & \nd 30.53 & \fs 36.10   \\
    & & LSD$^\downarrow$ & 3.96 & 1.41 & 1.40 & 1.26 & \rd 1.19 & \nd 0.99 & \fs 0.78  \\ 

   \cdashline{2-10}
    & \multirow{2}{*}{ScaledSine} & SNR$^\uparrow$ & 0.00 & 7.03 & 11.84 & 15.74 & \rd 22.18 & \nd 30.91 & \fs 35.70   \\
    & & LSD$^\downarrow$ & 3.88 & 1.42 & 1.44 & 1.28 & \rd 1.22 & \nd 0.99 & \fs 0.80 \\ 
   
      \hline
      \hline
      
   \multirow{8}{*}{Counting(\texttt{7s})} &\multirow{2}{*}{Gaussian}  & 
    SNR$^\uparrow$ & 0.76 & 2.43 & 6.34 & \nd 8.26 & \rd 6.42 & \fs 12.27 & 4.04  \\
    & & LSD$^\downarrow$ & 2.73 & \rd 2.16 & 2.33 & 2.25 & \nd 2.11 & \fs 1.97 & 2.43 \\ 
    
    \cdashline{2-10}
    & \multirow{2}{*}{ReLU} & SNR$^\uparrow$ & 0.00 & 0.01 & 0.39 & 0.97 & \rd 2.36 & \nd 6.16 & \fs 9.41 \\
    & & LSD$^\downarrow$ & 1.75 & 1.73 & \fs 1.69 & \nd 1.70 &  1.74 & 1.78 & \rd 1.71 \\ 

    \cdashline{2-10}
    & \multirow{2}{*}{Sine} & SNR$^\uparrow$ & 0.00 & 0.00 & 5.56 & 7.86 & \rd 8.82 & \nd 10.96  & \fs 12.68  \\
    & & LSD$^\downarrow$ & 3.22 &  3.01 & 1.94 & 1.76 &\nd  1.61 & \rd 1.68 & \fs 1.46  \\

   \cdashline{2-10}
    & \multirow{2}{*}{ScaledSine} & SNR$^\uparrow$ & 0.00 & 0.00 & 4.97 & \nd 7.93 & 6.84 &  \rd 7.65 &  \fs 12.60 \\
    & & LSD$^\downarrow$ &  3.02 & 2.94 & 1.96 & 1.75 & \rd 1.52 & \nd 1.47 & \fs 1.45 \\ 

    \hline
    \hline
   \multirow{8}{*}{Blues(\texttt{30s})} &\multirow{2}{*}{Gaussian}  & SNR$^\uparrow$ & 0.66 & 1.50 &  2.88 & \nd 6.65 & \fs 9.62 &\rd 6.45 & 5.37 \\
    & & LSD$^\downarrow$ & 1.95 & 1.59 & 1.38 & 1.23 & \fs 1.18 & \rd 1.23 & \nd 1.22   \\ 
    
    \cdashline{2-10}
    & \multirow{2}{*}{ReLU} & SNR$^\uparrow$ & 0.00 & 0.11 & 0.41 & 0.93 & \rd 1.78 & \nd 3.16 & \fs 6.71  \\
    & & LSD$^\downarrow$ & 2.97 &  2.89 & 2.42 & 2.08 & \rd 1.86 & \nd 1.66 & \fs 1.36 \\ 

    \cdashline{2-10}
    & \multirow{2}{*}{Sine} & SNR$^\uparrow$ & 0.00 & 0.00 & 1.92 & 4.11 & \rd 7.71  & \nd 9.81 & \fs 11.64 \\ 
    & & LSD$^\downarrow$ & 4.80 & 4.56 & 1.64 & 1.39 & \nd  1.32  & \rd 1.31 & \fs 1.28   \\

   \cdashline{2-10}
    & \multirow{2}{*}{ScaledSine} & SNR$^\uparrow$ & 0.00 & 0.00 & 1.86 & 4.90 & \rd 8.38 & \nd 9.94 &  \fs 11.56   \\
    & & LSD$^\downarrow$ & 4.69 & 4.54 & 1.64 & 1.32 & \rd 1.29 & \nd 1.28 & \fs 1.27 \\ 
  \hline
  \end{tabular}
}
 \label{tab:fkan_omega}
\end{table}

\begin{table}[!ht]
    \centering
  \caption{The impact of the frequency parameter \( \omega \) on Sine ($\sigma(x)=\sin(\omega x)$) and ScaledSine ($\sigma(x)=a\sin(\omega bx + c) + d$) activation functions. For each row, best results are highlighted as  \colorbox{colorFst}{first}.}
\resizebox{0.8\linewidth}{!}{
  \begin{tabular}{c|c|ccccccc}
    \hline
    Dataset & \multicolumn{2}{c}{\textbf{Parameter} $\omega$} & 3 &  30 & 300 & 3,000 & 30,000 & 300,000   \\
    
    \cline{1-9}
   \multirow{10}{*}{Bach(\texttt{7s})} &\multirow{2}{*}{ScaledSine} & SNR $\uparrow$ & -1.0 &  \fs 15.98 & 0.00 & -0.01 & -0.02 & -0.01  \\
    & & LSD $\downarrow$ &6.341 & \fs 0.778 & 2.441 & 2.442 & 2.432 & 2.466  \\ 

    \cdashline{2-9}
    & \multirow{2}{*}{\texttt{RFF}+ScaledSine} & SNR $\uparrow$ & 10.97 & \fs 38.10 & 0.05 & -0.02 & -0.01 & 0.01  \\
    & & LSD $\downarrow$ & 0.997 & \fs 0.595 & 2.465 & 2.389 & 2.491 & 2.483 \\ 
    
    \cdashline{2-9}
    & \multirow{2}{*}{\texttt{NeRF}+ScaledSine} & SNR $\uparrow$ & 0.00 & \fs 15.70 & 0.00 & -0.01 &  0.00 & -0.02  \\
    & & LSD $\downarrow$ & 5.685 & \fs 0.837 & 2.442 & 2.446 & 2.464 & 2.419    \\ 
    
    \cdashline{2-9}
    & \multirow{2}{*}{Sine} & SNR $\uparrow$ & 0.30 &  12.27 & 23.96 & 39.59 & \fs 40.36 & 0.01 \\
    & & LSD $\downarrow$ & 2.478 & 0.885 & 0.669 & \fs 0.575 &  0.683 & 2.526  \\ 

     \cdashline{2-9}
    & \multirow{2}{*}{\texttt{NeRF}+Sine} & SNR $\uparrow$ & 4.13 & \fs 42.39 & 0.80 & 0.70 & 1.12 & 0.00 \\
    & & LSD $\downarrow$ & 1.348 & \fs 0.537 & 2.537 & 2.589 & 2.523 & 2.678  \\ 

    \hline
    \hline
    
   \multirow{10}{*}{Counting(\texttt{7s})} &\multirow{2}{*}{ScaledSine} & SNR $\uparrow$ & -0.05 &  \fs 8.16 & -0.02 & -0.02 &  -0.06 & -0.01\\
    & & LSD $\downarrow$ & 5.184 & \fs 1.611 & 2.116 & 2.100 & 2.126 & 2.139 \\ 

    \cdashline{2-9}
    & \multirow{2}{*}{\texttt{RFF}+ScaledSine} & SNR $\uparrow$ & 4.04 & \fs 12.86 & -0.09 & -0.07 & -0.08 & -0.03 \\
    & & LSD $\downarrow$ & 1.730 & \fs 1.559 & 2.138 & 2.114 & 2.106 & 2.125\\ 
    
    \cdashline{2-9}
    & \multirow{2}{*}{\texttt{NeRF}+ScaledSine} & SNR $\uparrow$ &  -0.03 & \fs 11.03 &  -0.09 & -0.07 & -0.06 &  -0.01 \\
    & & LSD $\downarrow$ & 5.167 & \fs 1.784 & 2.104 & 2.126 & 2.121 & 2.114 \\ 

    \cdashline{2-9}
    & \multirow{2}{*}{Sine} & SNR $\uparrow$ & 0.26 & 7.98 & 11.79 & \fs 12.94 & 0.40 & -0.01  \\
    & & LSD $\downarrow$ & 2.153 & 1.716 & 1.710 & \fs 1.533 & 2.120 & 2.129 \\ 

     \cdashline{2-9}
    & \multirow{2}{*}{\texttt{NeRF}+Sine} & SNR $\uparrow$ & 4.81 & \fs 33.57 &  0.82 & 0.89 & 0.27 & 0.06 \\
    & & LSD $\downarrow$ & 1.849 & \fs 0.914 & 2.135 & 2.124 & 2.145 & 2.123 \\ 

  \hline
  \end{tabular}
}
  \label{tab:omega_incode_sine}
\end{table}

\begin{table}[!ht]
    \centering
\caption{The impact of the variance parameter \( \sigma \) on Gaussian ($\sigma(x) = e^{-\frac{x^2}{2\sigma^2}}$) activation function. For each row, best results are highlighted as  \colorbox{colorFst}{first}.}
  \begin{tabular}{c|ccccccc}
    \hline
     Dataset & Parameter $\sigma$ & 0.01 &  0.1 & 0.2 & 0.3 & 1.0   \\
    
   \hline
   \multirow{2}{*}{Bach(\texttt{7s})} & SNR $\uparrow$ & 0.01 & \fs 6.35 & 0.04 & 0.00 & -0.01  \\
    & LSD $\downarrow$ & 1.799 & \fs 1.130 & 3.851 & 4.120 & 6.266 \\ 

    \hline
    \multirow{2}{*}{Counting(\texttt{7s})} & SNR $\uparrow$ & 0.00 & \fs 0.74 & 0.01 & 0.00 & -0.01  \\
    & LSD $\downarrow$ & \fs 1.889 & 2.165 & 3.334 & 4.107 & 5.169  \\ 
  \hline
  \end{tabular}
  \label{tab:a_gaussian}
\end{table}

\begin{table*}[!htbp]
    \centering
    \caption{Impact of positional encoding on Coordinate-KANs for image regression tasks (PSNR$^\uparrow$).}
\resizebox{\textwidth}{!}{%
    \begin{tabular}{ccccccccccccccccccccccc}
        \cline{1-23}
        P.E. & B-Spline & Chebyshev & Chebyshev2 & Gegenbauer & Hermite & Jacobi & Laguerre & Legendre & Tayler & Bessel & Fibonacci & Lucas & Fourier & Sine & MexicanHat & Meyer & Morlet & DoG & Shannon & BSRBF & RBF & \texttt{Avg.} \\
        \cline{1-23}
        \texttt{Id.} & 21.99 & 24.52 & 27.73 & 12.20 & 28.35 &24.21 & 21.98 & 22.75 &  20.11 & 24.15 &  21.50 & 21.96 & \fs 33.56 & 21.20 & \fs 24.07 & 24.52 & 12.21 & 22.96 & 12.21 & 12.48 & 21.61 & 21.73  \\
        \texttt{NeRF} & \fs 25.38 & \fs 30.05 & \fs 32.28 & \fs 34.14 & \fs 30.61 & \fs 30.29 & \fs 24.21 & \fs 27.44 &  \fs 24.93 & \fs 28.66 & \fs 23.66 & \fs 24.49 & 24.33 & \fs 22.24 & 16.08 &\fs 27.70 & 12.21 & \fs28.60 & \fs17.05 &\fs 34.36 &  \fs24.92 & \fs 25.89  \\
        \texttt{RFF} & 18.48 &  18.41 & 21.45 & 29.08 & 20.73 &21.02 &13.25 & 15.50 & 12.21 & 15.32 & 12.71 & 16.19 & 24.92 & 12.21 & 15.64 & 15.30 & 12.21 & 16.19 & 12.21 & 17.57 & 14.28 & 16.90 \\
        \cline{1-23}
    \end{tabular}
}
    \label{tab:pe_kan_img_reg}
\end{table*}

\begin{table*}[!htbp]
    \centering
    \caption{Impact of normalization on Coordinate-MLPs in image regression tasks (PSNR$^\uparrow$).}
\resizebox{\textwidth}{!}{%
    \begin{tabular}{cccccccccccccccccc}
        \cline{1-16}
        Norm & ReLU & PReLU & Sine & ScaledSine& Gaussian & Laplacian & SuperGaussian & Gabor & Sinc & ExpSin & Sigmoid & Tanh & Quadratic & MultiQuadratic & \texttt{Avg.} \\
        \cline{1-16}
        w/o Norm & 22.99 & 22.01 & \fs23.47 & \fs44.44 & \fs36.72 & \fs38.30 & \fs36.19 & 23.26 & 24.18 & 22.36 & 18.88 & 22.32 & 26.66 & 23.04 & \fs 27.49 \\
        LayerNorm & \fs23.27 & 22.48 & 23.45 & 23.55 & 26.85 & 34.74 & 33.79 & \fs 32.29 & \fs26.46 & \fs 22.66 & \fs 21.43 & \fs22.39 & \fs32.91 & \fs 26.18 & 26.60 \\
        BatchNorm & 22.59 & \fs22.50 & 16.24 & 19.00 & 23.77 & 12.20 & 23.66 & 24.23 & 24.04 & 22.21 & 19.87 & 21.59 & 24.25 & 24.18 & 21.45 \\
        CrossNorm & 9.17 & 7.64 & 11.37 & 9.72 & 11.88 & 11.37 & 11.57 & 11.78 & 11.91 & 9.38 & 7.57 & 10.41 & 11.58 & 11.54 & 10.49    \\
        GlobalNorm & 20.07 & 17.45 & 14.13 & 15.66 & 14.11 & 16.94 & 11.08 & 14.63 & 13.51 & 12.42 & 15.77 & 15.47 & 16.24 & 18.10 & 15.40 \\
        \cline{1-16}
    \end{tabular}
}
    \label{tab:norm_img_reg}
\end{table*}

\textbf{Parameter Sensitivity of Activation Functions:}
Certain activation functions, such as Sine and Gaussian, incorporate additional parameters to regulate the frequency components learned by the network. We evaluate the impact of these parameters on the audio representation capability of activation functions through experiments conducted on audio regression tasks. Tables \ref{tab:omega_incode_sine} and \ref{tab:a_gaussian} present the parameter sensitivity results for periodic activation functions (Sine and ScaledSine) with respect to the frequency parameter \(\Omega\), and for the Gaussian activation function with respect to the variance parameter \(\sigma\), in the context of audio representation tasks. As discussed in Sec. \ref{sec:nonlinear}, activation functions with larger Lipschitz constants increase the eigenvalues of the NTK, thereby accelerating model convergence. However, while a larger L-Smooth constant broadens the spectral bandwidth and enables the model to capture more frequency components, it can also lead to training instability. These insights are valuable for guiding the selection of hyperparameters in activation functions. 
As shown in Tables \ref{tab:omega_incode_sine} and \ref{tab:a_gaussian}, the optimal frequency parameter for the Sine activation function is approximately 30, while the optimal variance parameter for the Gaussian activation function is around 0.1. The Lipschitz constants for both functions are on the order of \(10^1\), and their L-Smooth constants are on the order of \(10^2\) (see Sec. \ref{sec:Details_of_Activation_Functions}). These results indicate that for audio representation tasks, these parameter ranges effectively constrain the NTK spectrum, thereby ensuring both model convergence and training stability.

\subsection{Impact of Positional Encoding on Coordinate-KANs}
\label{sec:Impact_of_Positional_Encoding_on_Coordinate-KANs}
The parameters of Coordinate-KANs are dependent on the basis functions, and as a result, embedding position encodings leads to a substantial increase in model parameters, thereby exacerbating computational complexity. To explore this intriguing issue, we conducted experiments on image regression tasks. As shown in Table \ref{tab:pe_kan_img_reg}, we primarily evaluated the impact of \texttt{NeRF} and \texttt{RFF} position encodings on Coordinate-KANs. Position encodings, by enhancing the model's ability to learn high-frequency information, prove to be an effective strategy for Coordinate-KANs, which exhibit minimal spectral deviation in the low-frequency range. However, it is noteworthy that \texttt{NeRF} position encoding significantly improves the image representation capability of Coordinate-KANs, yielding an overall enhancement of approximately 4.16 dB. In contrast, \texttt{RFF} position encoding, due to its sensitivity to parameters, results in inferior representation performance, with a decrease of approximately 4.83 dB (although better results may be achievable through extensive hyperparameter tuning).

\begin{figure}[!htbp]
  \centering
  \resizebox{0.5\textwidth}{!}{%
  \includegraphics[width=\linewidth]{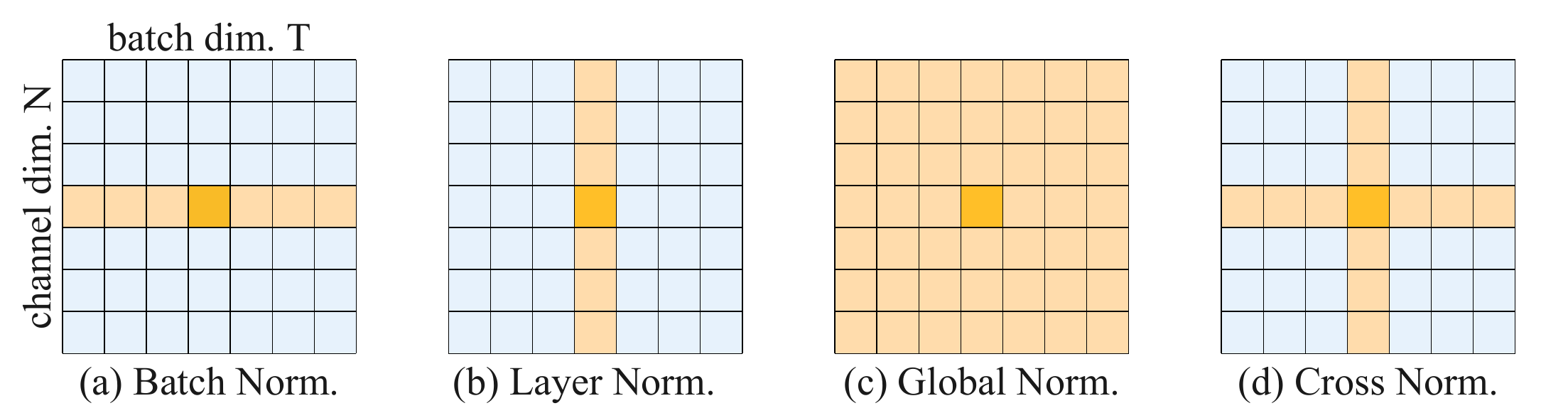}
  }
  \caption{Various normalization techniques (image source: \cite{cai2024towards}).}
  \label{fig:norms}
\end{figure}

\subsection{Impact of Normalization on Coordinate-MLPs}
\label{sec:Impact_of_Normalization_on_Coordinate-MLPs}
We observe that recent work by Norm-INR \cite{Cai2024BatchNA} has demonstrated that Batch Normalization can enhance the implicit representation capability of Coordinate-MLPs. To provide a more comprehensive and thorough analysis, we also conducted a comparative study on the impact of various normalization techniques on the implicit representation performance. As illustrated in Fig. \ref{fig:norms}, we primarily investigate the effects of four normalization techniques—Batch Normalization, Layer Normalization, Global Normalization, and Cross Normalization—on the performance of Coordinate-MLPs in image representation tasks. As shown in Table \ref{tab:norm_img_reg}, the overall impact of normalization techniques on Coordinate-MLPs is relatively limited. This is because implicit neural representations inherently require models to overfit the data, storing the information directly within the network weights. However, for certain complex activation functions (e.g., Gabor, Sinc, Sigmoid), normalization techniques can mitigate training instability, enabling more uniform capture of image signals. In contrast, Cross Normalization and Global Normalization face challenges in generalizing learned patterns from training to testing, due to significant data discrepancies between the batch size dimension and feature dimension.

\subsection{Qualitative Experiments}
\label{sec:Qualitative_Experiments}
This section presents qualitative experimental results across various tasks.

\begin{figure*}[!tbp]
  \center
  \includegraphics[width=\linewidth]{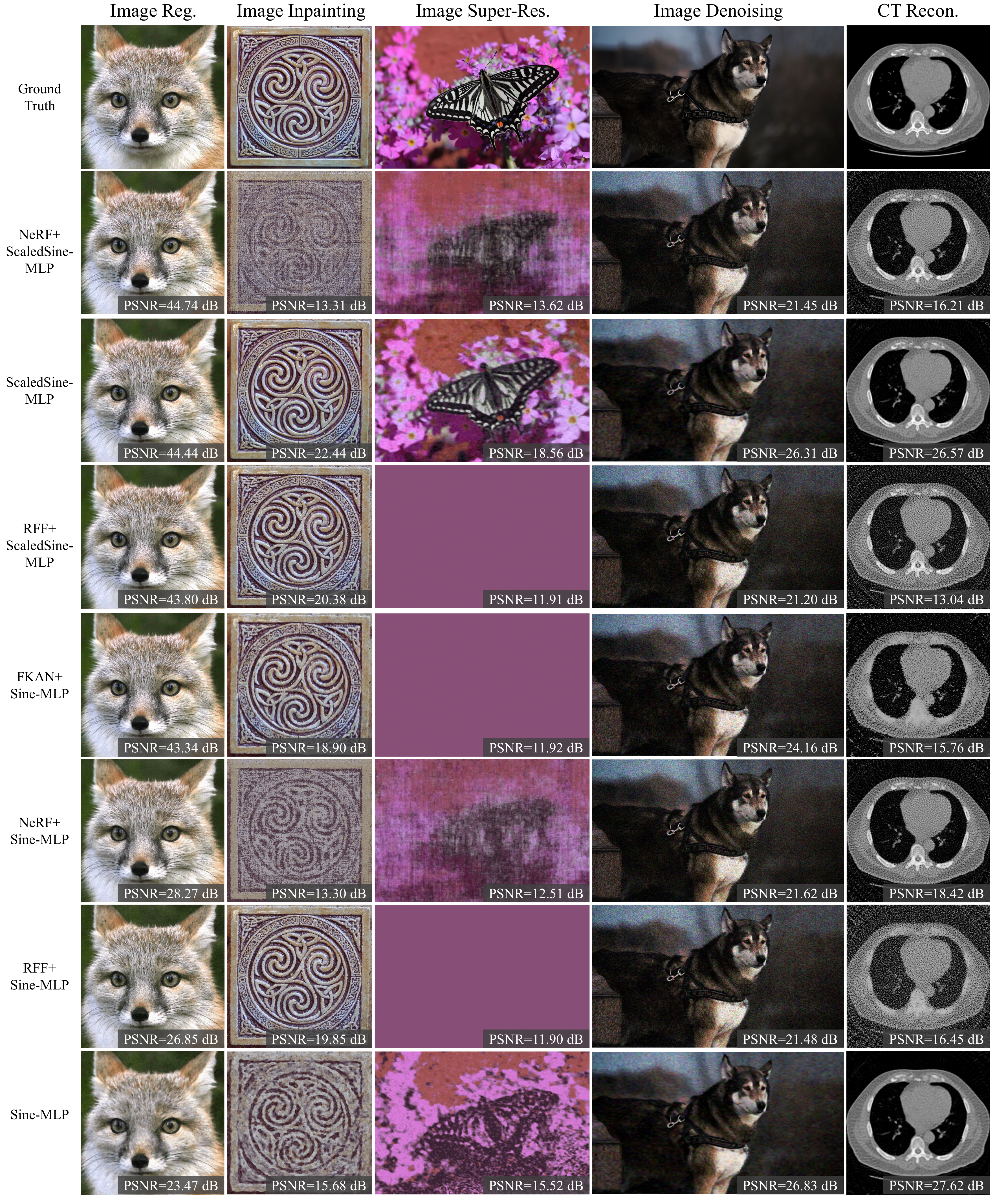}
  \caption{Qualitative experimental results related to image representation tasks.}
  \label{fig:image_psnr_sine}
\end{figure*}

\begin{figure*}[!tbp]
  \center
  \includegraphics[width=\linewidth]{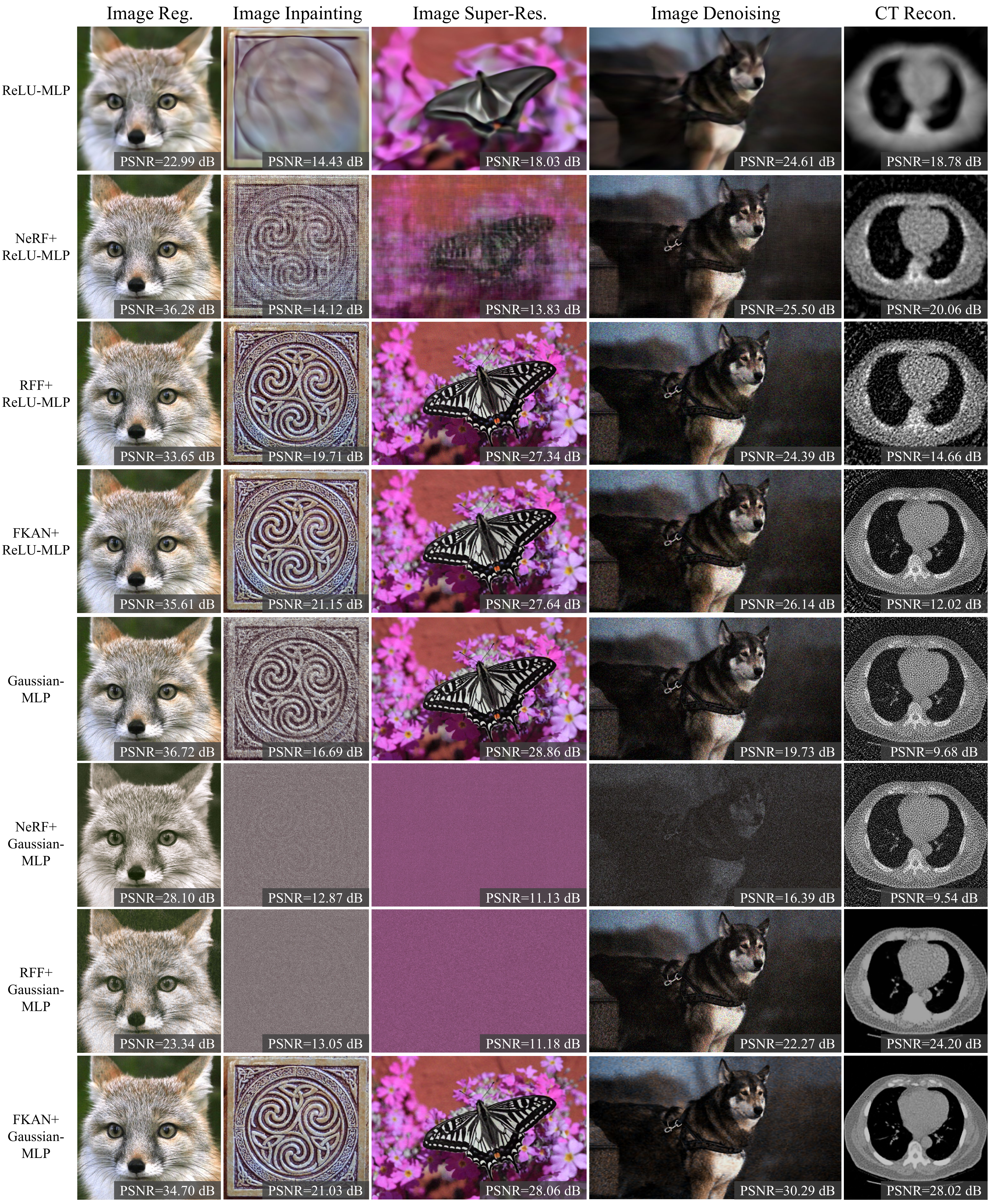}
  \caption{Qualitative experimental results related to image representation tasks.}
  \label{fig:image_psnr_relu}
\end{figure*}

\begin{figure*}[!tbp]
  \center
  \includegraphics[width=\linewidth]{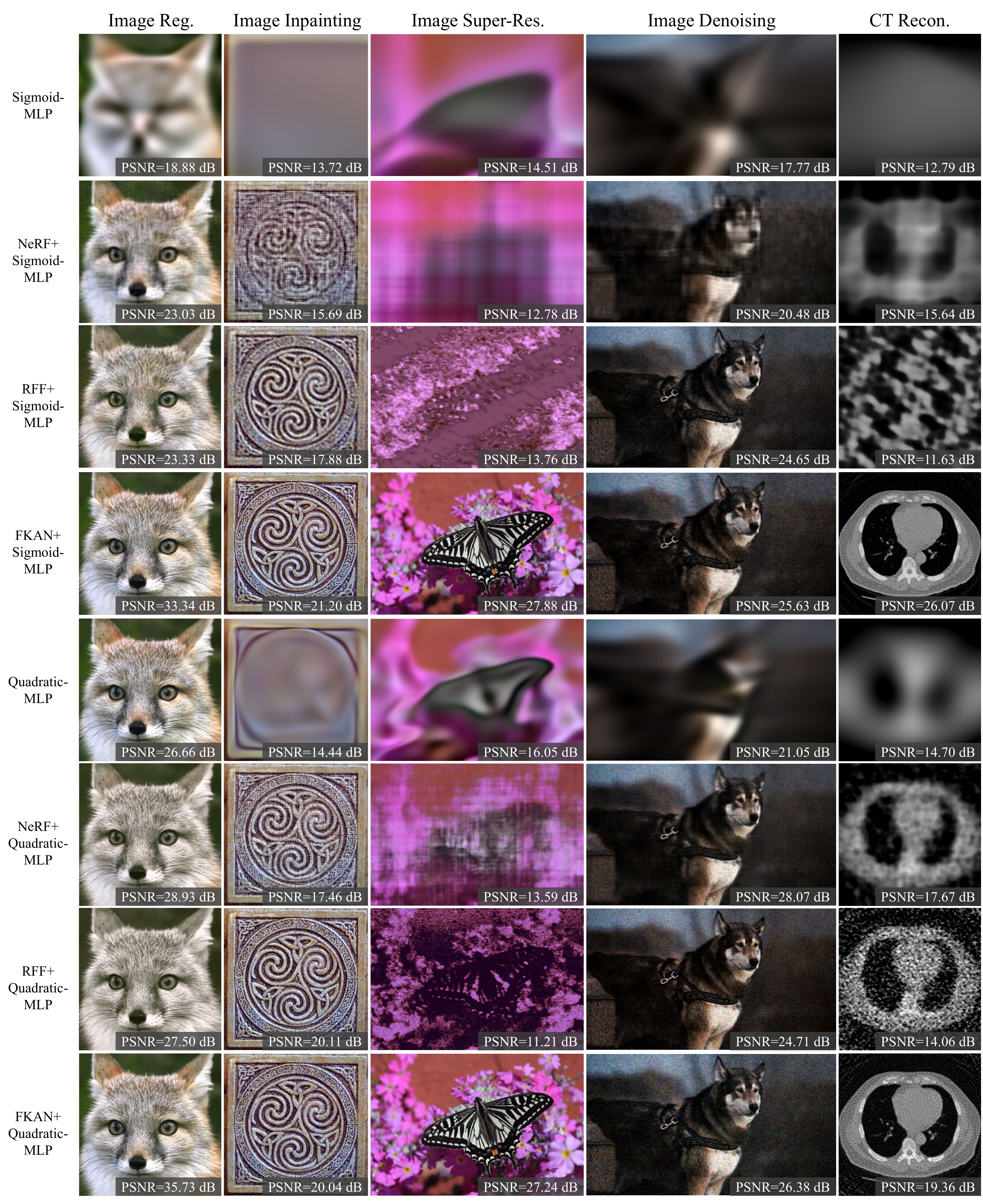}
 \caption{Qualitative experimental results related to image representation tasks.}
  \label{fig:image_psnr_sigmoid}
\end{figure*}

\begin{figure*}[!tbp]
  \center
  \includegraphics[width=\linewidth]{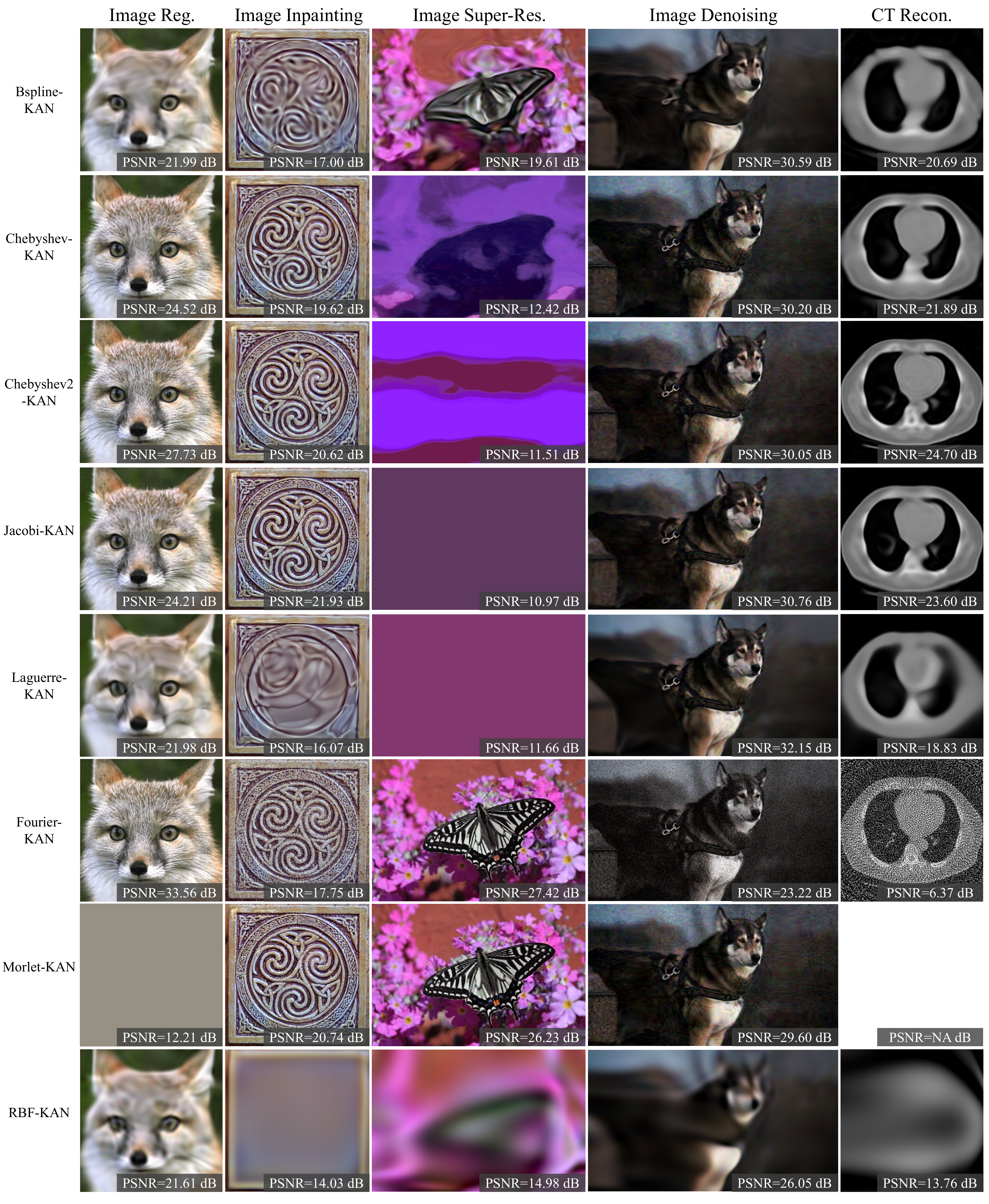}
  \caption{Qualitative experimental results related to image representation tasks.}
  \label{fig:image_psnr_kan}
\end{figure*}

Figures \ref{fig:image_psnr_sine}, \ref{fig:image_psnr_relu}, \ref{fig:image_psnr_sigmoid}, and \ref{fig:image_psnr_kan} present the qualitative experimental results for image-related tasks, including image regression, image inpainting, image super-resolution, image denoising, and CT reconstruction.  

\begin{figure*}[!tbp]
  \center
  \includegraphics[width=\linewidth]{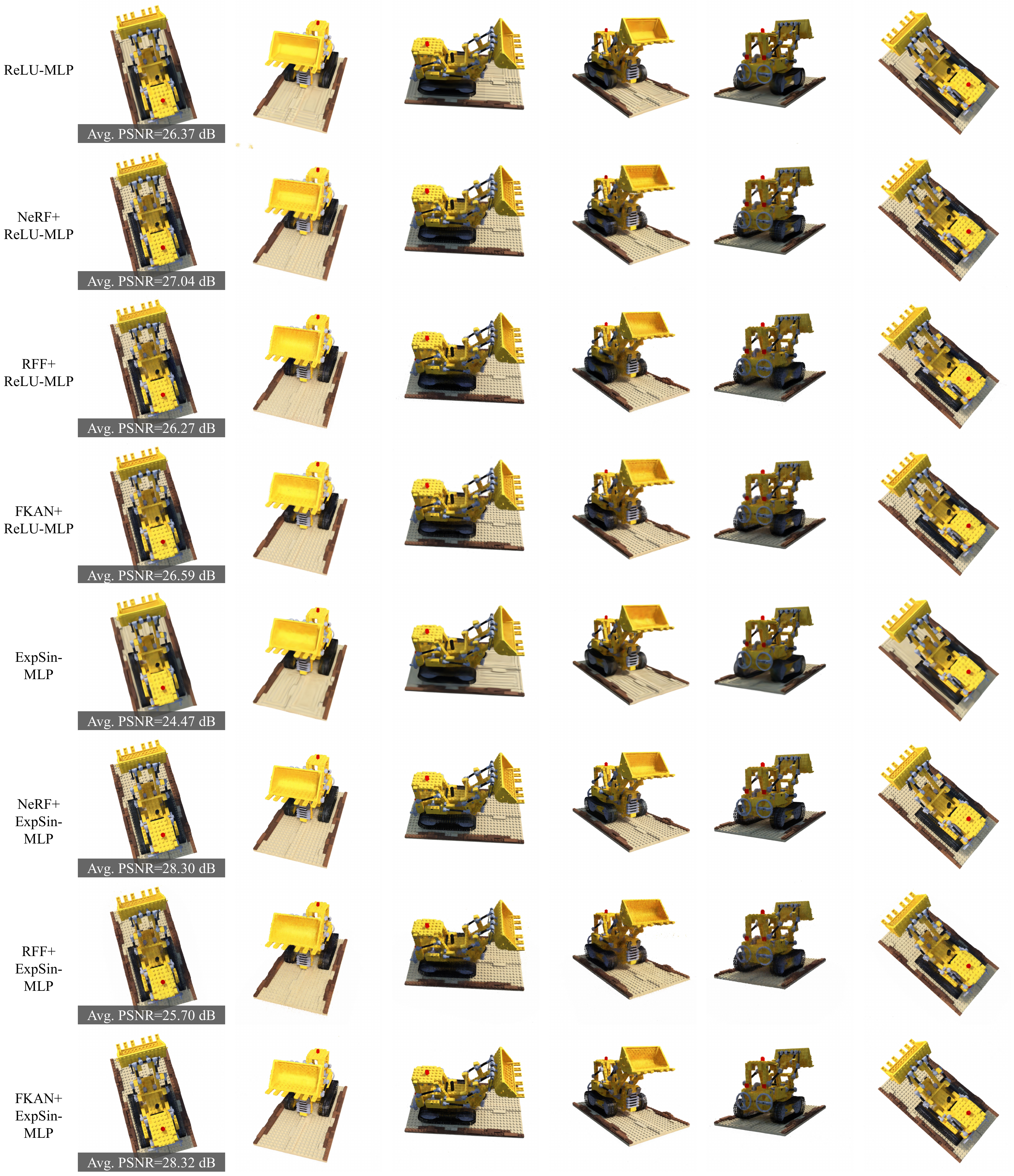}
  \caption{Qualitative results of Coordinate-MLPs on radiated fields based on vanilla NeRF \cite{mildenhall2020nerf} model.}
  \label{fig:mlp_nerf_relu}
\end{figure*}

\begin{figure*}[!tbp]
  \center
  \includegraphics[width=\linewidth]{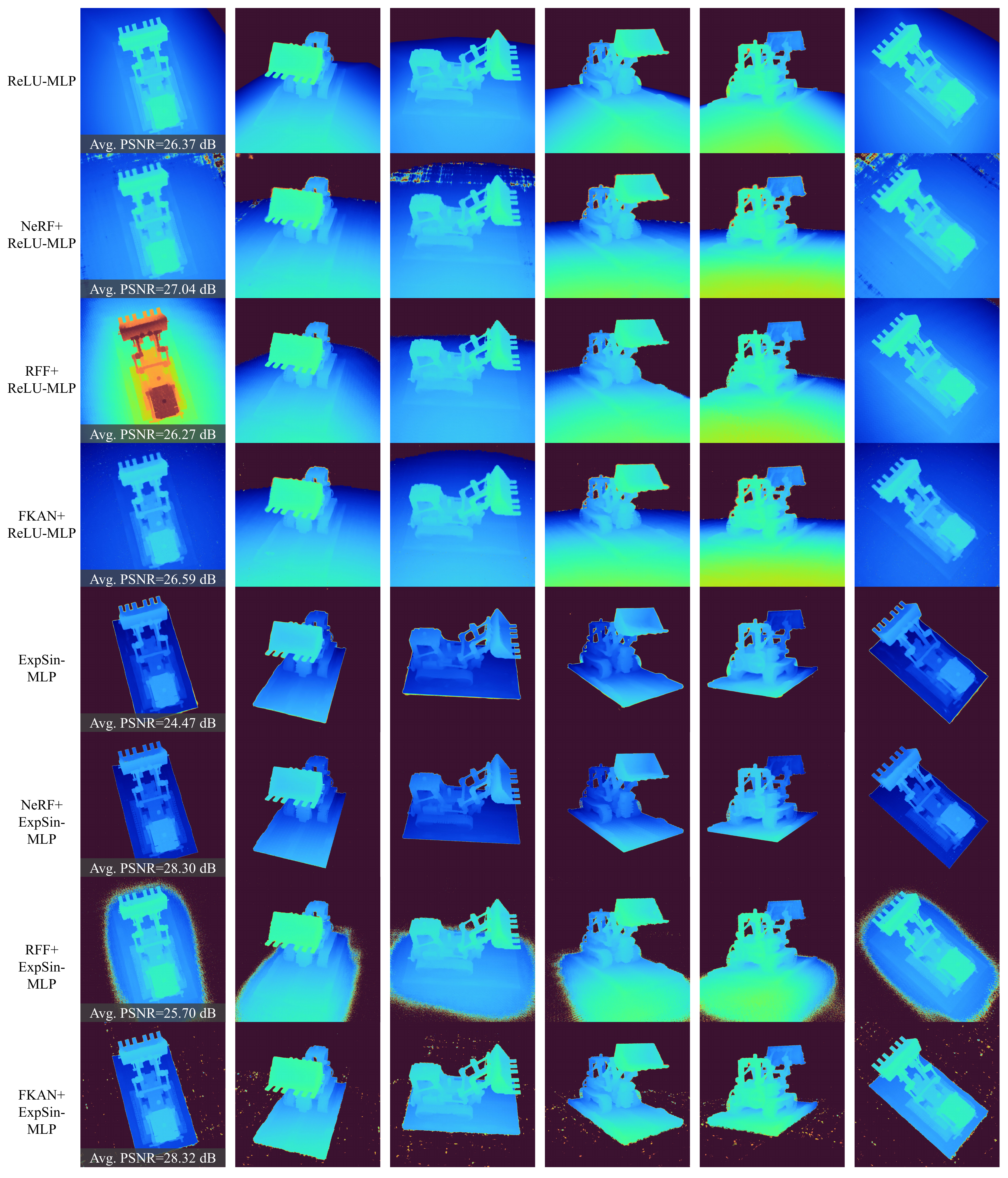}
  \caption{Qualitative results (rendered depth) of Coordinate-MLPs on radiated fields based on vanilla NeRF \cite{mildenhall2020nerf} model. Obviously, the ExpSine activation function has stronger boundary learning ability than ReLU.}
  \label{fig:mlp_nerf_relu_depth}
\end{figure*}

\begin{figure*}[!tbp]
  \center
  \includegraphics[width=\linewidth]{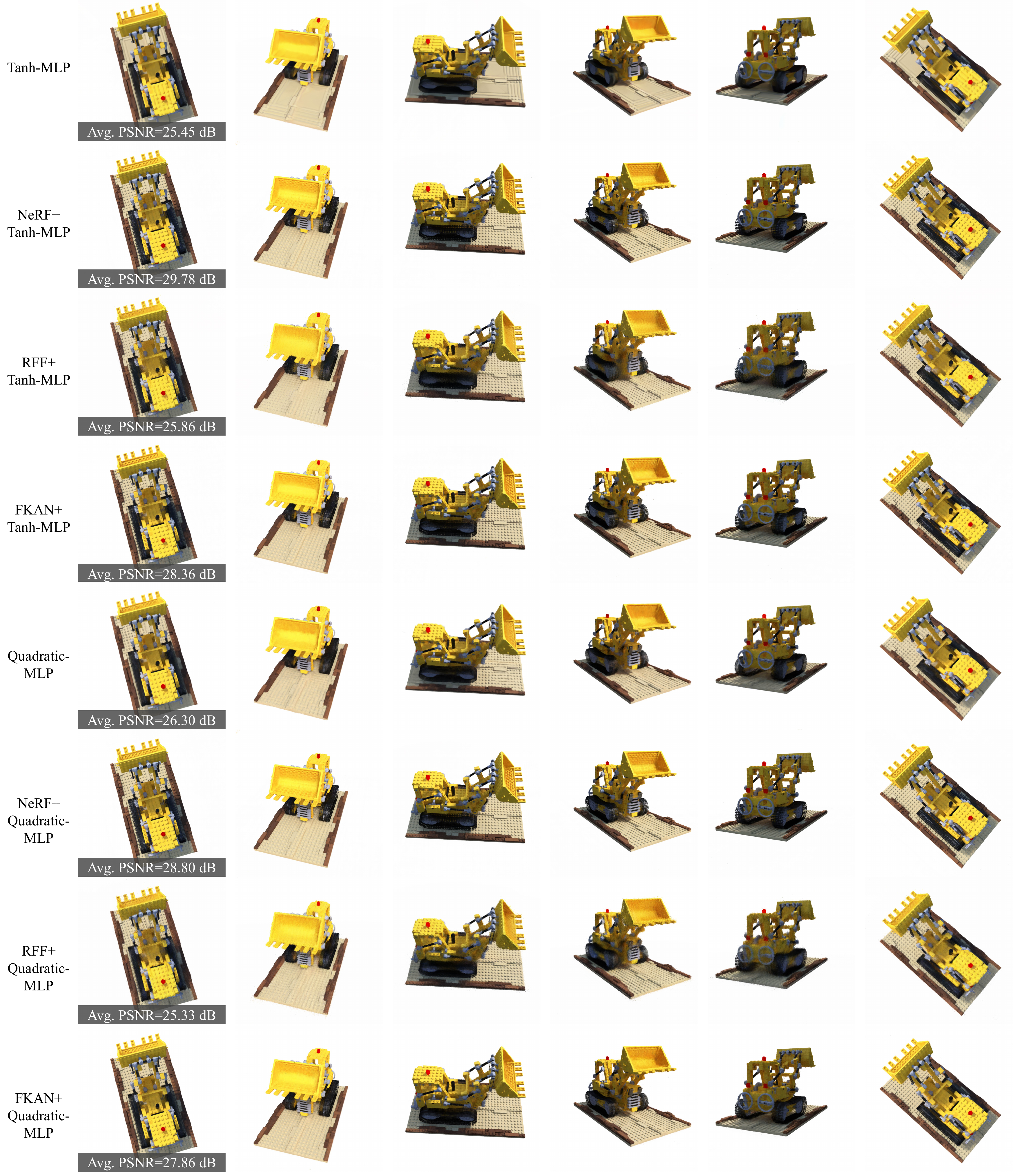}
  \caption{Qualitative results of Coordinate-MLPs on radiated fields based on vanilla NeRF \cite{mildenhall2020nerf} model. }
  \label{fig:mlp_nerf_tanh}
\end{figure*}

\begin{figure*}[!tbp]
  \center
  \includegraphics[width=\linewidth]{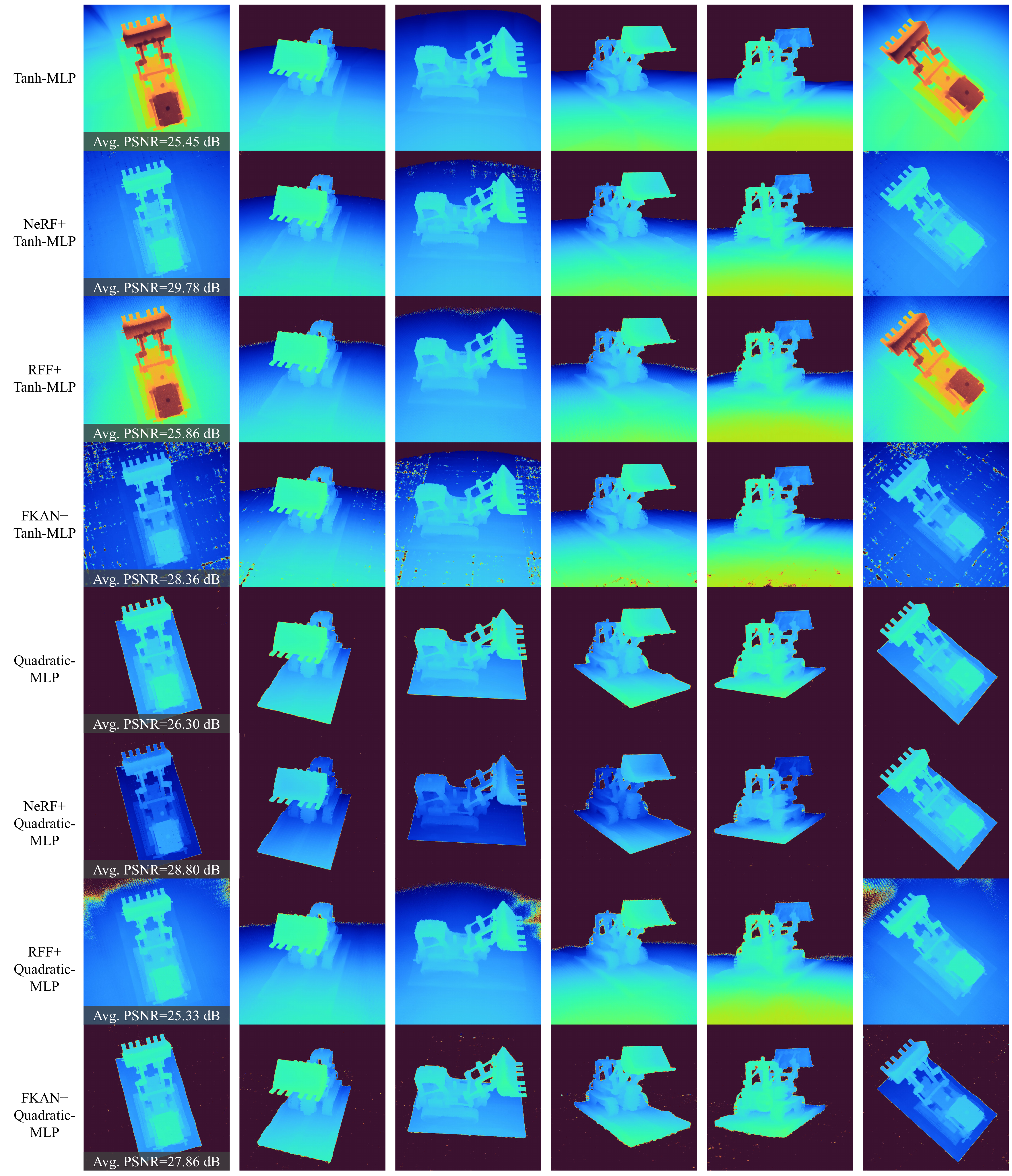}
  \caption{Qualitative results (rendered depth) of Coordinate-MLPs on radiated fields based on vanilla NeRF \cite{mildenhall2020nerf} model.The incorporation of Gaussian random noise in \texttt{RFF} effectively regulates the bandwidth of the NTK, thereby accelerating model convergence. However, this also negatively impacts the model's boundary learning capability, leading to a reduction in the geometric accuracy of the reconstruction.}
  \label{fig:mlp_nerf_tanh_depth}
\end{figure*}

Figures \ref{fig:mlp_nerf_relu}, \ref{fig:mlp_nerf_relu_depth}, \ref{fig:mlp_nerf_tanh} and \ref{fig:mlp_nerf_tanh_depth} illustrate the qualitative results of radiance field reconstruction using Coordinate-MLPs within a naive NeRF model, highlighting the effects of positional encoding and activation functions on neural radiance field reconstruction.  

\begin{figure*}[!tbp]
  \center
  \includegraphics[width=\linewidth]{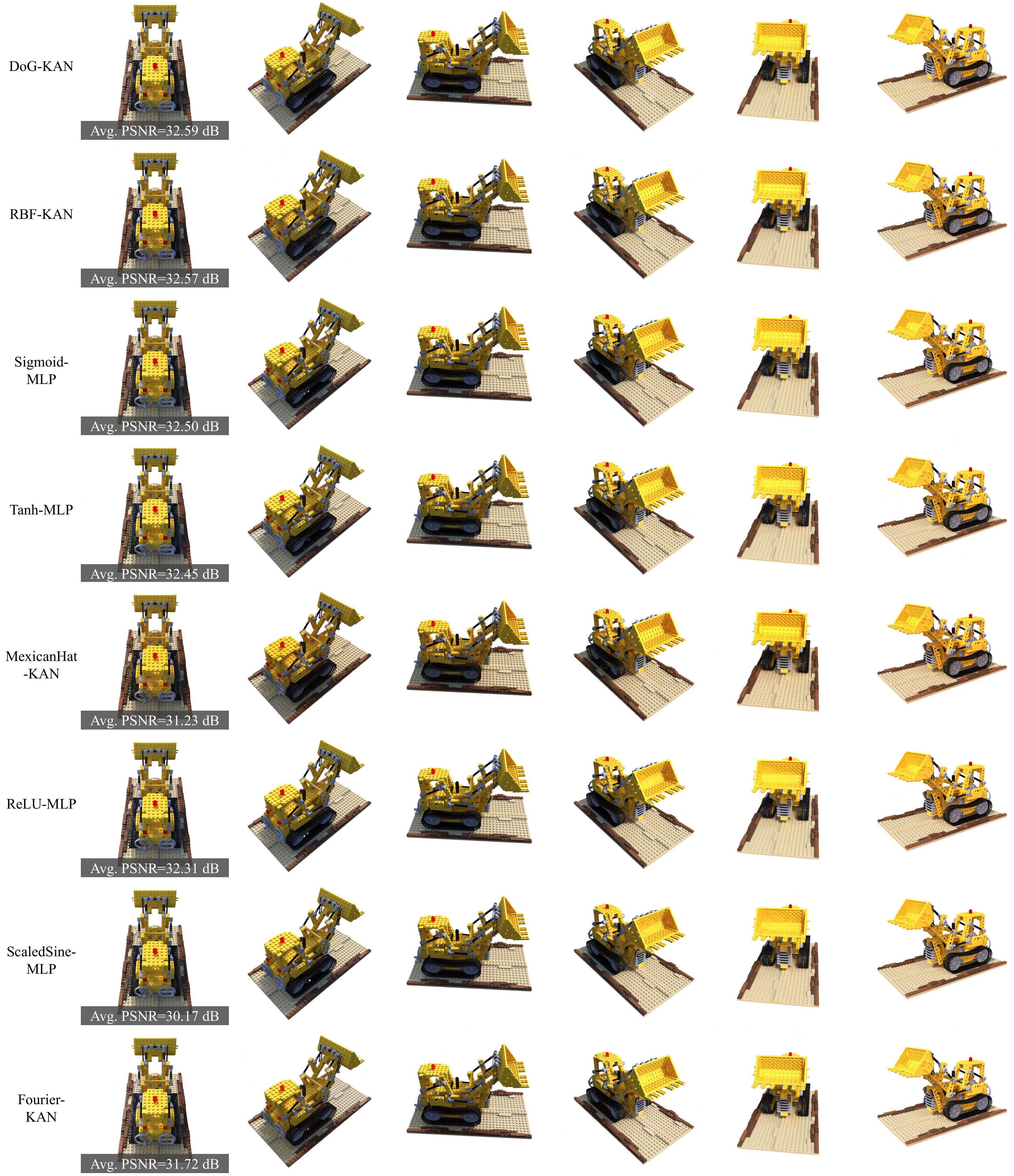}
  \caption{Qualitative results of Coordinate-MLPs and Coordiante-KANs on radiated fields based on \textbf{NGP} \cite{muller2022ngp} model. Overall, as high-dimensional feature mappers, KANs and MLPs exhibit minimal performance differences in the context of hash feature mapping.}
  \label{fig:ngp_mlp_kan}
\end{figure*}

Fig. \ref{fig:ngp_mlp_kan} compares the qualitative results of radiance field reconstruction between Coordinate-MLPs (without positional encoding) and Coordinate-KANs in the NGP model.  

\begin{figure*}[!tbp]
  \center
  \includegraphics[width=0.90\linewidth]{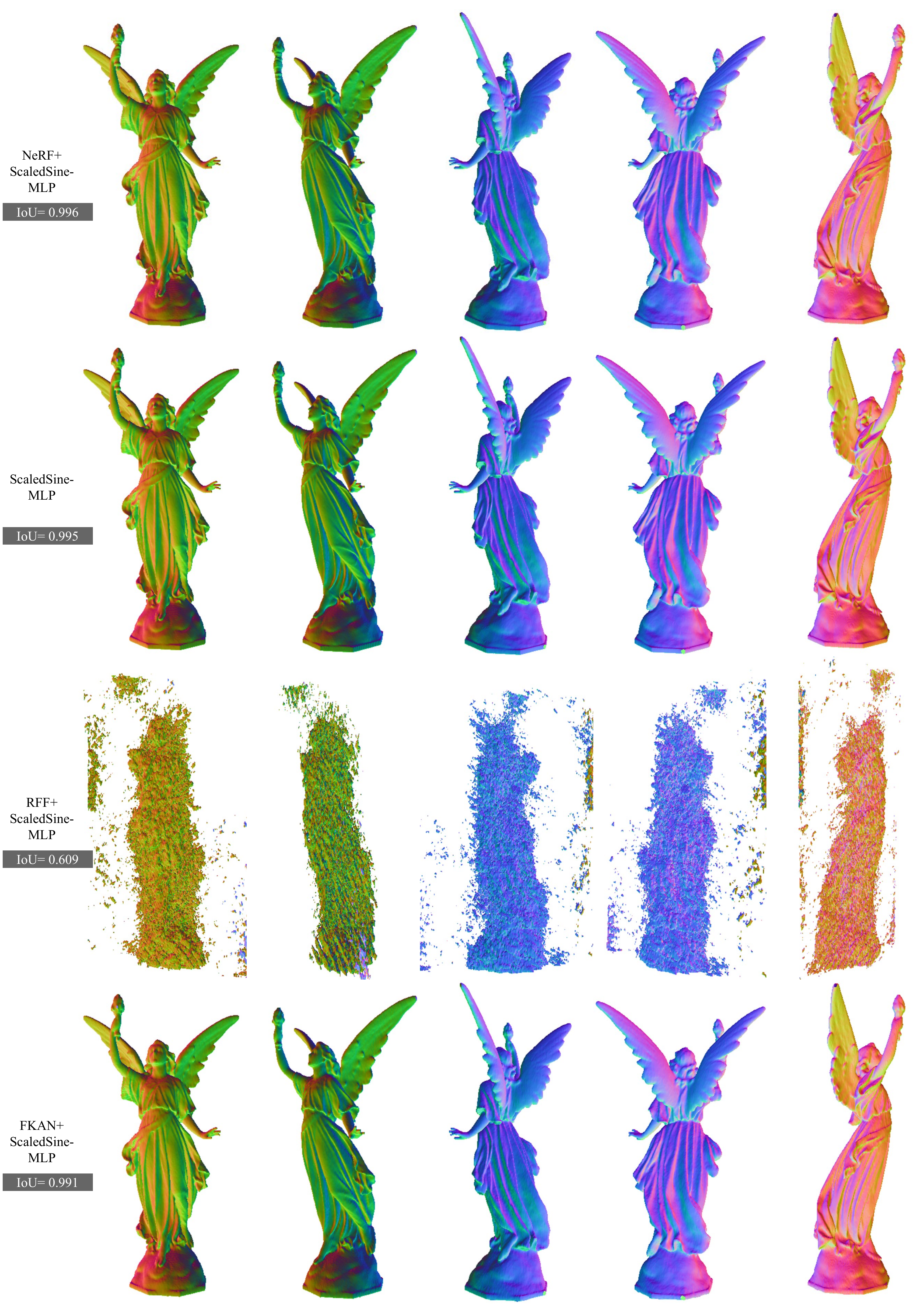}
  \caption{Qualitative results of Coordinate-MLPs on SDF regression task. }
  \label{fig:sdf_iou_sine}
\end{figure*}

\begin{figure*}[!tbp]
  \center
  \includegraphics[width=0.90\linewidth]{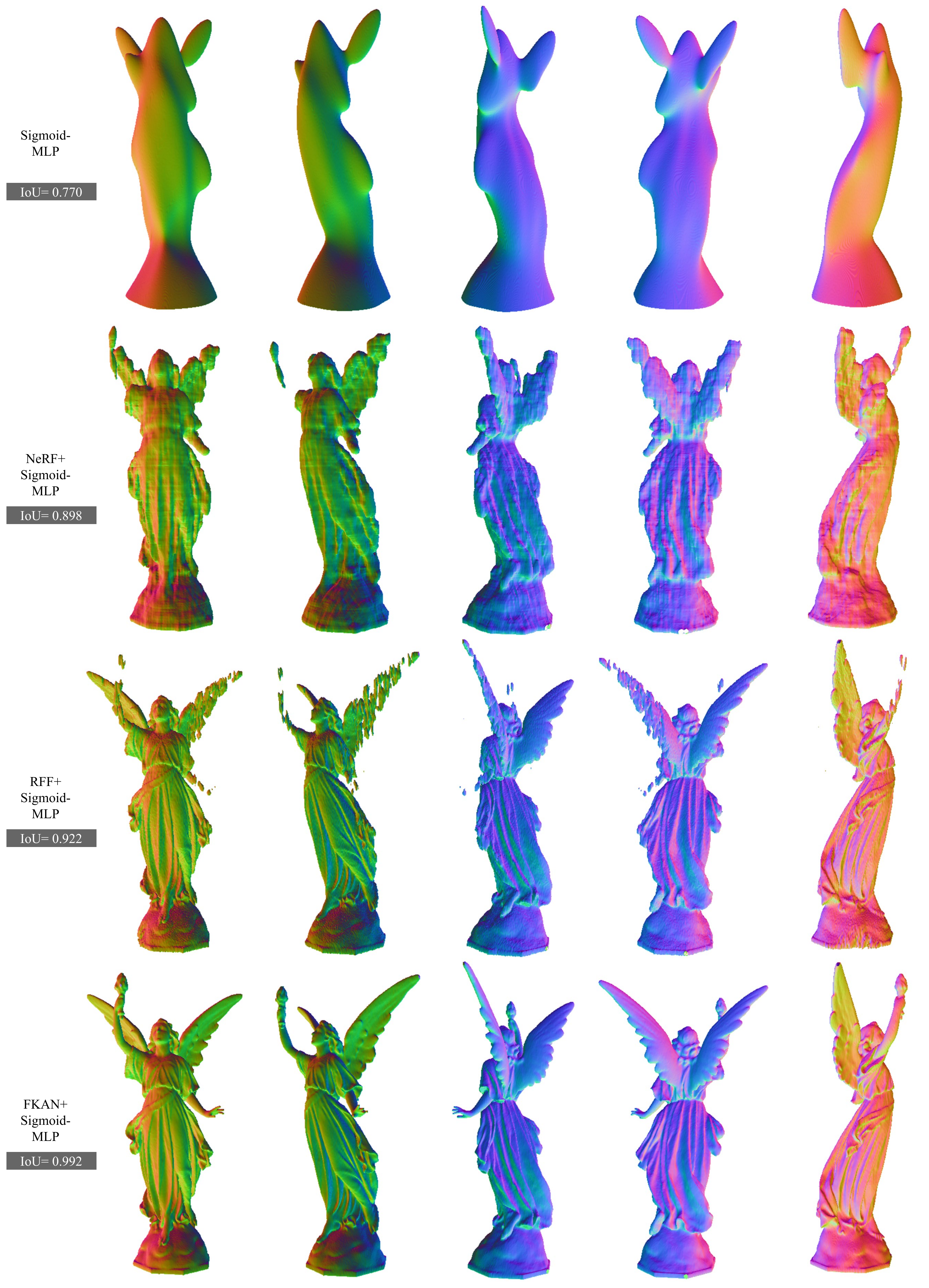}
  \caption{Qualitative results of Coordinate-MLPs on SDF regression task. }
  \label{fig:sdf_iou_sigmoid}
\end{figure*}

\begin{figure*}[!tbp]
  \center
  \includegraphics[width=0.90\linewidth]{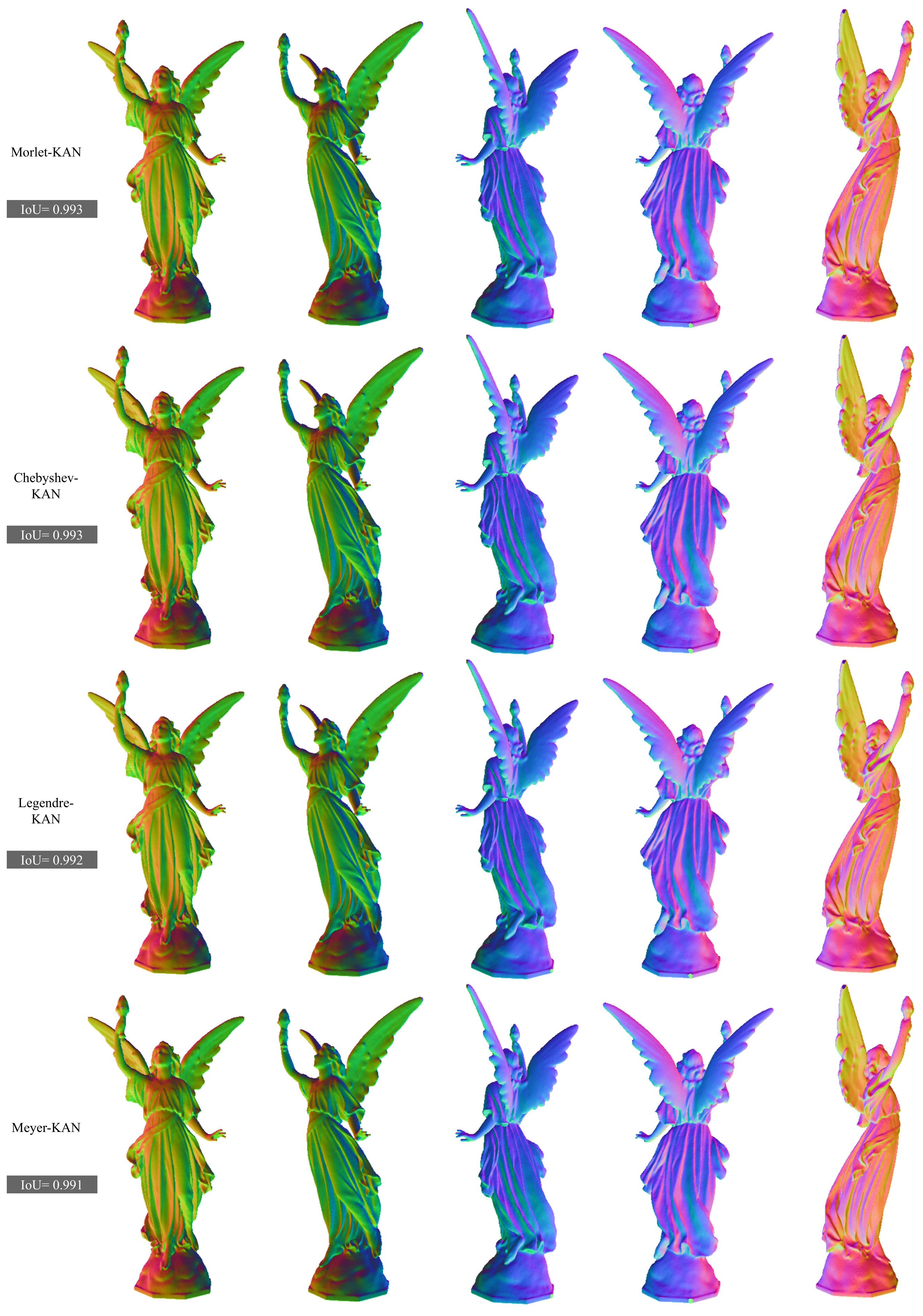}
  \caption{Qualitative results of Coordinate-KANs on SDF regression task. }
  \label{fig:sdf_iou_kan_morlet}
\end{figure*}

\begin{figure*}[!tbp]
  \center
  \includegraphics[width=0.90\linewidth]{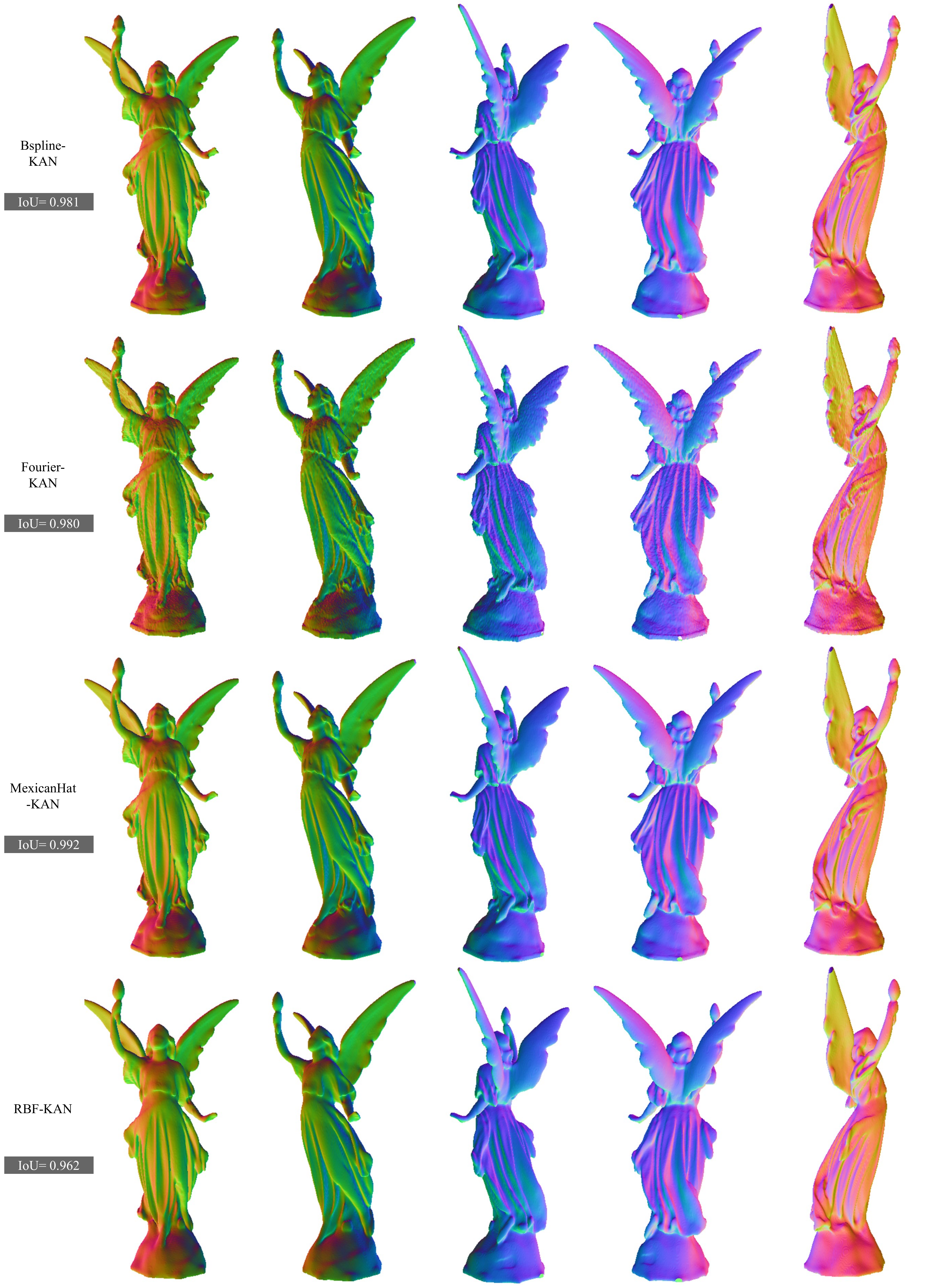}
  \caption{Qualitative results of Coordinate-KANs on SDF regression task. }
  \label{fig:sdf_iou_kan_bspline}
\end{figure*}

Figures \ref{fig:sdf_iou_sine}, \ref{fig:sdf_iou_sigmoid}, \ref{fig:sdf_iou_kan_morlet} and \ref{fig:sdf_iou_kan_bspline} showcase the qualitative results of SDF regression tasks, comparing the performance of Coordinate-MLPs and Coordinate-KANs.  

\begin{figure*}[!tbp]
  \center
  \includegraphics[width=0.86\linewidth]{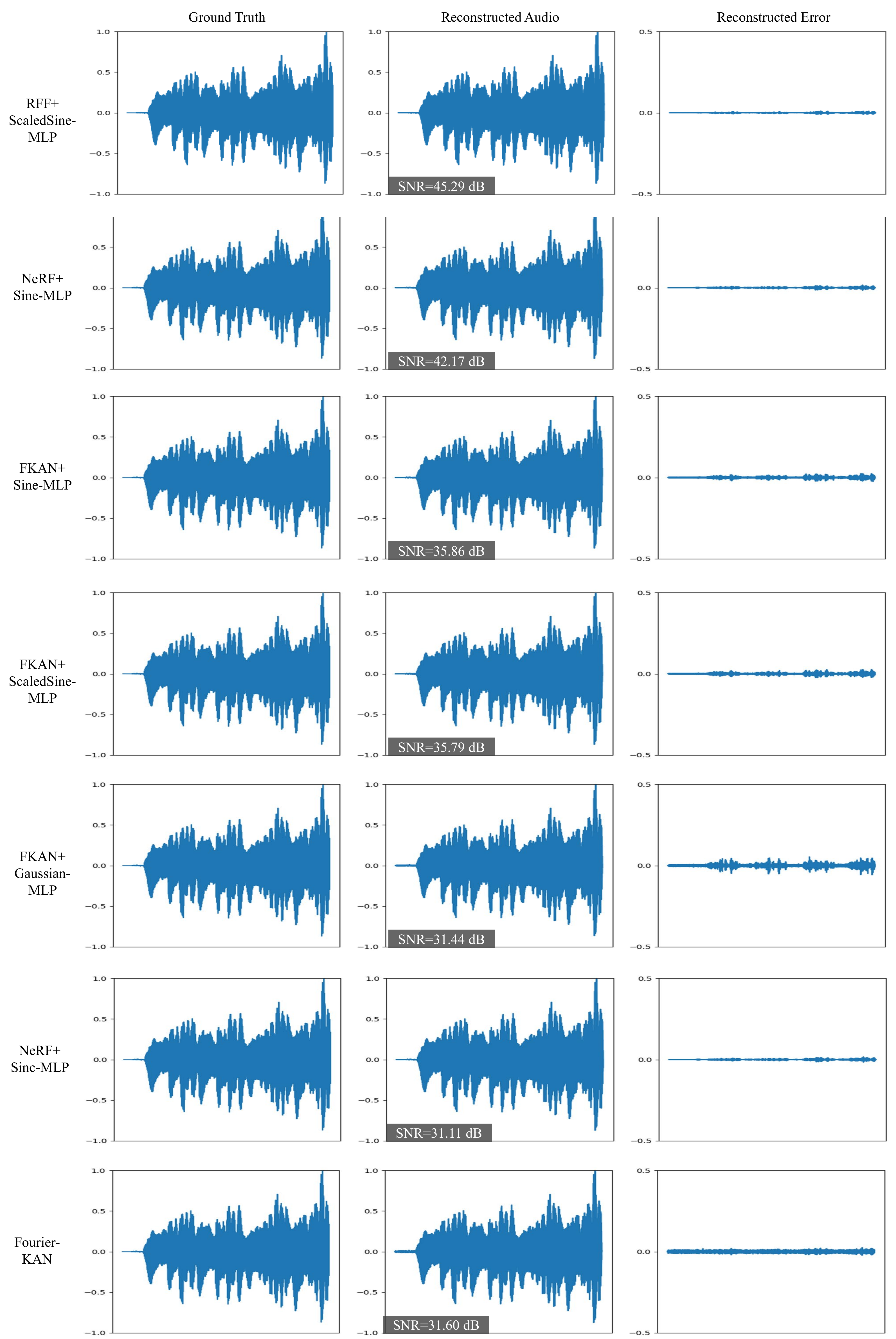}
  \caption{Qualitative results of Coordinate-KANs on Audio regression task. }
  \label{fig:audio_snr_sine}
\end{figure*}

\begin{figure*}[!tbp]
  \center
  \includegraphics[width=0.86\linewidth]{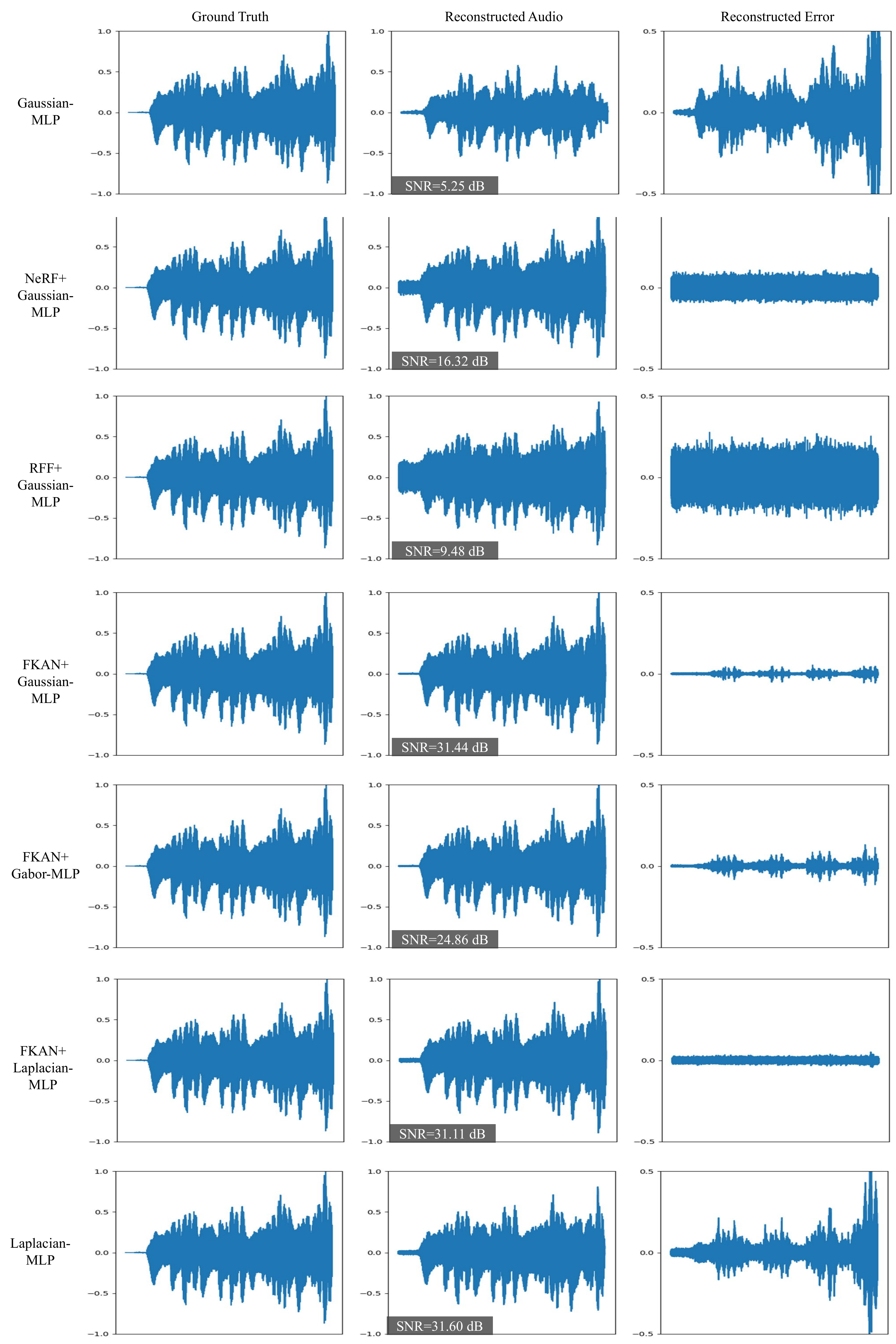}
  \caption{Qualitative results of Coordinate-KANs on Audio regression task. It is evident that the \texttt{FKAN} positional encoding significantly enhances the ability of Coordinate-MLPs to capture high-frequency and local periodicity within audio signals. }
  \label{fig:audio_snr_gaussian}
\end{figure*}

Finally, Figures \ref{fig:audio_snr_sine} and \ref{fig:audio_snr_gaussian} present the qualitative results of audio reconstruction tasks, comparing Coordinate-MLPs and Coordinate-KANs.

\end{document}